\documentclass[11pt]{article}
\usepackage[margin=1in]{geometry}
\usepackage{amsmath,amssymb,amsthm,booktabs,array,multirow}
\usepackage{graphicx}
\graphicspath{{../}{./}}
\usepackage{hyperref}
\usepackage{microtype}
\usepackage{mdframed}
\usepackage{placeins}
\newcommand{\R}{\mathbb{R}}
\newcommand{\E}{\mathbb{E}}
\newcommand{\Var}{\operatorname{Var}}

\newcommand{\vecq}{\vec q}
\newcommand{\vecx}{\vec x}

\newcommand{\vecz}{\vec z}
\newcommand{\dd}{\mathrm{d}}

\title{I-BBS: Coordinate-Free Inference of Latent Sub-Manifolds \\
Using Random Distance Matrix Theory}

\author{Igor Halperin\thanks{All
calculations, numerical analysis, and manuscript
preparation were performed by Claude Code with Opus 4.8
working as an AI assistant under the author's supervision. I
would like to thank Charles Martin for numerous discussions
and comments on the manuscript. All remaining errors are my
own. All Python code, analysis scripts, and figures
supporting this paper are available at
\url{https://github.com/ighalp/I-BBS}.}}

\date{\today}

\makeatletter
\newcommand{\labelalias}[2]{%
  \@ifundefined{r@#2}{}{%
    \expandafter\edef\csname r@#1\endcsname{\csname r@#2\endcsname}%
  }%
}
\makeatother

\begin{document}
\maketitle

\begin{center}
Email: ighalp@gmail.com
\end{center}

\begin{abstract}
Bogomolny, Bohigas and Schmit (BBS) found that the
spectrum of the pairwise distance matrix on $N$ points
sampled from a smooth $d$-dimensional manifold encodes a
signature of the underlying geometry. We develop I-BBS
(Inference-BBS), a coordinate-free method that
identifies a low-dimensional latent sub-manifold embedded
in a high-dimensional ambient distance matrix alone,
without accessing an ambient high-dimensional vector
space. It therefore applies even when that space is only
partly observable or undefined. We model
the ambient embedding by two classes of generative noise,
model-based and model-free. The noise mixes the latent
signal with off-manifold components, so the eigenvalues
reorganise collectively and the latent geometry cannot be
read off eigenvalue by eigenvalue. We recover it instead
from two integer-stable signatures that survive the noise:
the multiplicity of the top non-Perron multiplet, which
fixes $d$, and a parameter-free law for how the multiplet
positions shrink as the noise grows. On synthetic spheres $S^1$, $S^2$ and $S^3$ these
integer signatures are far more stable under noise than the
continuous spectral slope, and a blind test recovers both
the manifold and the noise model from a single distance
matrix. Applications to
neural-network representations and to the dynamic training
regime are developed in two companion papers.
\end{abstract}

\setcounter{tocdepth}{1}
\tableofcontents

\section{Introduction}
\label{sec:intro}

Many complex natural systems and human-created datasets,
such as images and text, often carry latent
low-dimensional structure beneath a high-dimensional
ambient representation. In machine learning, the
emergence of a low-dimensional manifold as a good
approximation to the high-dimensional activation space
of a neural network is itself a diagnostic of structural
changes or phase transitions during training. Whether
such a low-dimensional sub-manifold is present, and, if
so, what its intrinsic dimension and geometry are,
is therefore the central inference question across these
settings, and the question we address here from
observations alone.

Bogomolny, Bohigas, and Schmit (BBS)
\cite{bogomolny2003, bogomolny2007} showed that the spectrum
of a pairwise distance matrix (defined below) of $N$ uniform
samples from a smooth low-dimensional manifold encodes, in
the large-$N$ limit, the manifold's intrinsic dimension and
isometry-group structure. For a distance matrix $M$ of distances on a
hyper-sphere $S^{d-1}$, BBS theory
predicts a sequence of discrete multiplets at the top of
the spectrum, a delocalised power-law tail $|\Lambda_K|
\sim K^{-d/(d-1)}$ in the rank $K$ ordered by descending
magnitude, and a complementary localised branch at the
bottom whose small eigenvalues accumulate with the BBS
exponent $\beta_{\rm loc}=1/(d-1)$. Section~\ref{sec:bbs-crash} reviews the BBS
framework in the level of detail required here.

This paper develops Inference-BBS (I-BBS), an inference
framework that uses BBS theory to read the latent geometry
off the ambient distance matrix when the observed data
lies \emph{near} (rather than exactly on) a
low-dimensional sub-manifold $\mathcal M_d$ embedded in a
high-dimensional ambient space. We work on the ambient
sphere $S^{D-1}$, with $D$ of order hundreds; we use $D = 125$ and
$D = 128$ (ambient spheres $S^{124}$ and $S^{127}$) in our numerical
experiments\footnote{The unit multi-dimensional sphere
is a common ambient arena for modern high-dimensional data.
The single unit sphere $S^{D-1}$ corresponds, in
particular, to the normalized activations of modern
neural networks: normalized layers (LayerNorm, RMSNorm,
$\ell_{2}$-normalized projection heads, and similar
constructions used in transformers, convolutional, and
self-supervised models) are commonly chosen to have
unit normalization in $\R^{D}$, which is equivalent
to $S^{D-1}$. The choice $D = 128$ is motivated by the
companion papers~\cite{halperin2026OMD, halperin2026grokking},
which apply I-BBS to the distance matrix between neural
activations in the learning dynamics of neural networks.}.
\emph{I-BBS is coordinate-free: unlike most
dimension-reduction or manifold-learning methods, it
operates only on the distance matrix and not on the
underlying vector space itself, which may be partially or
fully unobservable, or even undefined.} Also, unlike many
other statistical methods, our approach can make inference
from a \textbf{single observation}, by relying on the
``magic'' of random matrix theory (RMT), see below.

We adopt the geodesic distance matrix
$M^{(n)}_{ij} := \arccos(\vec u_i\!\cdot\!\vec u_j)$
on $N$ unit vectors $\vec u_i\in S^{n-1}\subset\R^{n}$
as the primary object throughout. This is the
geodesic distance studied in BBS-2 \cite{bogomolny2007}, of
negative type unconditionally on the unit sphere
and full rank generically, so the BBS spectral
diagnostics (reviewed in Sec.~\ref{sec:bbs-crash}) are
directly available.
The signal-plus-noise picture organising the
paper is then a decomposition into three BBS
distance matrices: the deterministic latent
$M^{(d)}$ on the low-dimensional sub-manifold
$\mathcal M_d\subset S^{d-1}$ (the \emph{signal}),
the random residual $M^{(D-d)}$ on the
complementary sub-sphere $S^{D-d-1}$ (the
\emph{noise}), and the observed ambient $M^{(D)}$
on $S^{D-1}$ (the \emph{observable}). A model-based construction (see Sec.~\ref{sec:aaag}), built
from an orthogonal split of the ambient coordinates
$\R^{D} = V_{1}\oplus V_{2}$, relates the three at the
cosine-kernel level by the convex combination
$\cos M^{(D)} = (1-\varepsilon^{2})\cos M^{(d)}
+ \varepsilon^{2}\,\cos M^{(D-d)}(\bar y)$, with $\bar y$
the random residual sample on $S^{D-d-1}$. Inference asks
how the BBS spectral fingerprint of $M^{(d)}$ can be read
off $M^{(D)}$ across the range of $\varepsilon$ in which
the noise has not yet washed it out.

We hasten to clarify here what I-BBS does and
what it does not do. As a part of their analysis of random
distance matrices for manifolds embedded in some ambient
Euclidean space, the BBS theory also considers the
\emph{embedding problem}: finding a specific particle
configuration in that Euclidean space that matches a given
distance matrix for the manifold embedded in it. We work with the
high-dimensional ambient matrix $M^{(D)}$ as a noisy
observation of a latent {\it low-dimensional} matrix
$M^{(d)}$ and infer the latter directly, namely the
manifold $\mathcal M_d$ and its distance matrix $M^{(d)}$
together with their BBS spectral content, so we bypass the
embedding problem for the ambient $D$-dimensional space
altogether. We may then perfectly well add a
solution of the embedding problem for the inferred
$M^{(d)}$, recovering a configuration $\{\vec x_i\}$ on
$S^{d-1}$ that realises the latent distances, for downstream
tasks such as visualising the training process of neural
networks~\cite{halperin2026OMD, halperin2026grokking}.

We split the forward (generative) step into two classes,
borrowing the terminology from reinforcement learning. A
\emph{model-based} approach posits an explicit stochastic
embedding $\vec x_i \to \vec X_i$ of the latent unit
vectors $\vec x_i \in S^{d-1}$ into ambient unit vectors
$\vec X_i \in S^{D-1}$ and reads $M^{(D)}_{ij} =
\arccos(\vec X_i^\top \vec X_j)$ off them. A
\emph{model-free} approach specifies the conditional
distribution of $M^{(D)}_{ij}$ given $M^{(d)}_{ij}$
directly, with no $\vec X_i$ constructed.

We pursue both classes in this paper. The \emph{Residual Sphere Mixture}
(RSM) is model-based: the ambient coordinates split into a latent subspace
$V_1$ carrying the noiseless signal on $S^{d-1}$ and a residual subspace $V_2$
carrying the noise as a BBS distance matrix on $S^{D-d-1}$, but the vector space
enters only at the construction stage, with every calculation done on the
distance matrix itself (see Sec.~\ref{sec:aaag}). The \emph{Free Spectral Mixture}
(FSM) is model-free: it builds the ambient cosine kernel
$\cos M^{(D)}_{ij}$ directly as a positive-definite zonal kernel on the product
of the latent and residual sub-spheres, with a freely chosen nonnegative
spectral array $\{\beta_{pq}\}$ and a noise amplitude
$\varepsilon^{2}=1-\beta_{10}$, and no ambient vectors constructed at all
(see Sec.~\ref{sec:skm}). Both apply when the high-dimensional space is partially or
fully unobservable, or absent altogether, as for financial correlation matrices.

The model-free framing adds two things that the RSM model does not have. First,
it covers every $SO(d)\times SO(D-d)$-invariant cosine kernel, of which the
linear convex combination of RSM (see Eq.~\eqref{eq:aag-MD-cosine}) is the
two-coefficient corner. Second, it makes the angular-momentum content of the
noise readable from the observed spectrum alone, through the nonnegative array
$\{\beta_{pq}\}$ that no subspace-split model produces. Because the construction
keeps the kernel positive semidefinite by Schoenberg's theorem (the coefficients
are nonnegative), $M^{(D)}=\arccos(\cos M^{(D)})$ is a genuine spherical distance
matrix for every FSM instance. Distinct off-diagonal entries of $\cos M^{(D)}$
are not independent: they are coupled through the shared residual coordinates and
the shared spectral array.

Both classes share a single noise amplitude $\varepsilon$ controlling
$M^{(D)} - M^{(d)}$, with $\varepsilon \to 0$ collapsing the model onto the
identity. As discussed in Sec.~\ref{sec:pt-no-single-level} and
Appendix~\ref{sec:appendix-beta-sphere}, the eigenvalues of the observed ambient
distance matrix reorganise collectively, so the latent spectrum of $M^{(d)}$ is
\emph{not} read level by level. What survives the noise is the multiplet
structure, where the isotropic perturbation shifts each BBS multiplet as a whole,
so the integer multiplicities are gap-protected and give the latent-dimension
readout (see Sec.~\ref{sec:mult-invariant}).

Our inference method is able to work with a single
(multi-dimensional, matrix-valued) observation by employing
\emph{self-averaging}. The latter amounts to concentration of
spectral observables of large random matrices around their
ensemble mean with fluctuations vanishing as $N\to\infty$, so
the $N$-normalised matrix $\tfrac1N M^{(d)}$ fixes the spectral
signatures of the latent geometry to $O(N^{-1/2})$ in its
operator-normalised eigenvalues \cite{Bun2017} (the unnormalised
multiplet positions are $O(N)$ and their finite-$N$ spread is
$O(\sqrt N)$). This is the
``magic'' (or rather a kind of \emph{blessing of
dimensionality}) of RMT: the spectral functionals of one large
matrix concentrate, so a large sample size makes
inference possible, unlike regression where each coefficient
needs many observations. It also enables the \emph{blind}
setting of this paper: the multiplet multiplicities, the
shrinkage, and the residual shape are self-averaged
functionals sharp enough to read both the latent dimension and
the noise model off a single $M^{(D)}$.

Inference is then organised around the structural
signatures of the ambient matrix $M^{(D)}$:
\begin{enumerate}
\item the multiplicity $\hat h_1$ of the lowest
non-Perron multiplet, an integer-valued, gap-protected
readout of the latent dimension via the representation
theory of the latent isometry group. For a hyper-sphere
$S^{d-1}$, $\hat h_1 = d$; the inversion table for tori and
projective spaces is in Sec.~\ref{sec:multiplet-table}.
\item a parameter-free angular-momentum-level shrinkage law that
places the multiplet positions
(see Sec.~\ref{sec:attenuation-law}): each angular-momentum-$\ell$
multiplet is shrunk under noise by a factor fixed by
the Funk--Hecke spectrum of the geodesic kernel, and the
latent positions are recovered by matching the
shrinkage-corrected multiplets to the clean BBS tower.
\item the angular-momentum component that the noise populates, a
fingerprint of the generative noise class: the model-free
FSM injects an $\ell = 2$ component, through a nonnegative
spectral coefficient $\beta_{20}$, that the isotropic RSM
kernel leaves empty by parity, with the residual-spectrum
bulk providing an exploratory random-matrix consistency
check (see Sec.~\ref{sec:exp-blind}, Appendix~\ref{sec:exp-rmt}).
\end{enumerate}
The integer multiplicity is the most stable handle; the
finite-$N$-corrected power-law slope (BBS asymptote
$d/(d-1)$) is a secondary low-noise cross-check. The
stability is gap-protected: as long as the off-manifold
perturbation $\|\delta M\|_{\rm op}$ stays below half the
inter-multiplet gap $g$, Weyl's eigenvalue inequality keeps
the multiplet an isolated cluster of the same integer size
and the Davis--Kahan $\sin\theta$ theorem keeps its eigenspace
coherent (rotation bounded by $\|\delta M\|_{\rm op}/g$),
both made quantitative in Sec.~\ref{sec:multiplet-table}.

We validate the framework on synthetic hyper-sphere
latents from $S^1$, $S^2$ and $S^3$ across RSM and FSM.
The integer multiplet diagnostic recovers the correct
manifold in every realisation up to a relative noise level
$\eta = 0.5$, and beyond it for the isotropic noise model,
while a blind test identifies both the manifold and the
noise model from a single ambient matrix. The eigenvalue
positions themselves degrade smoothly, tracking the
shrinkage law.
The results in this paper rest on a combination of theoretical
results (BBS theory, the Davis--Kahan stability theorem, and
random matrix theory) with numerical analysis (Funk--Hecke-based
kernel denoising and quasi-degenerate
L\"owdin--Schrieffer--Wolff perturbation theory) and direct
simulations.
Applications to real neural-network representations and
to the dynamic training regime are developed in two
companion papers~\cite{halperin2026OMD, halperin2026grokking}.

\paragraph{Outline.}
Section~\ref{sec:bbs-crash} reviews the BBS framework.
Section~\ref{sec:construction} organises the forward
(generative) step of I-BBS into a model-free class (FSM)
and a model-based class (RSM).
Section~\ref{sec:perturbation-Dmatrix} treats the inverse
problem, and establishes the gap-protected multiplicity
invariant and the angular-momentum-level shrinkage law, and
packages the pipeline as Algorithm~1 (the I-BBS pipeline).
Section~\ref{sec:experiments} reports the inference
experiments that test it step by step.
Section~\ref{sec:discussion} discusses scope, limitations,
and extensions, and Section~\ref{sec:summary} summarises.
Appendices~\ref{sec:appendix-RSM} and~\ref{sec:appendix-FSM}
collect detailed derivations for the RSM and FSM generative
models, respectively.
Appendix~\ref{sec:appendix-beta-sphere} provides details of
RMT analysis of distance matrices.
Appendix~\ref{sec:finite-N} analyses finite-$N$ corrections
to the BBS theory.
Appendix~\ref{sec:appendix-qdpt} gives details of tests
involving the quasidegenerate perturbation theory.

\subsection{Related work}
\label{sec:related-work}

Intrinsic-dimension estimation has a long history in nonlinear dimensionality
reduction. Maximum-likelihood estimators read $d$ off nearest-neighbour
distances under a local Poisson model \cite{LevinaBickel2004}. The TwoNN
estimator \cite{Facco2017} uses the smallest neighbourhood and is stable under
density variation. Topological data analysis and persistent homology
\cite{Chazal2021, Otter2017} extract coordinate-free invariants. Spectral
methods such as Laplacian eigenmaps and diffusion maps \cite{Coifman2006}
preserve local diffusion distances. These operate on the point cloud directly,
not on the spectrum of the pairwise distance matrix.

Closer to ours is the analysis of distance and Gram matrices as random
operators in high dimension. El~Karoui \cite{ElKaroui2010} and Bordenave \cite{Bordenave2012}
analysed high-dimensional kernel matrices by Taylor expansion around the
concentrated norm, and Couillet and Liao \cite{CouilletLiao2022} treat distance-
and inner-product kernel matrices on Gaussian-mixture inputs (their Ch.~4) via a
spiked-Wishart-plus-Euclidean decomposition, with the focus on class recovery
rather than dimension inference. The EDMA framework \cite{Lele1993} handles
coordinate noise on landmark configurations by moments on centred distance
matrices. The replica analyses of M\'ezard, Parisi, and Zee \cite{Mezard1999}
and Casaburi and Vivo \cite{Casaburi2026a, Casaburi2026b} reduce the
extremal-eigenvalue problem of Euclidean random matrices to self-consistent
equations. El~Karoui \cite{ElKaroui2008} inverts the Marchenko--Pastur equation
to estimate a population spectrum from sample eigenvalues by discretising the
spectral measure into point masses. The
Manifold-Spectrometrics framework of Chen and Ma \cite{ChenMa2025} disentangles
a low-rank signal from heteroskedastic noise in sample-covariance spectra.

In kernel-learning terms, I-BBS recovers the latent BBS signal $M^{(d)}$ on
$S^{d-1}$ from the noisy ambient $M^{(D)}$ ($D\gg d$), the contamination supplied
by $M^{(D-d)}$. This is dual to kernel \emph{approximation}, where random
features \cite{RahimiRecht2007, Liu2021} compress a kernel into a tractable
surrogate: both expand the kernel in a spectral basis on the latent manifold,
but our target is kernel \emph{inference} of $M^{(d)}$ and its dimension $d$ from
the BBS spectral signatures (the $h_1=d$ multiplet, the angular-momentum-level
shrinkage of the positions, and the residual shape), reading $d$ off the
spectrum rather than fixing it.

The two model classes of Sec.~\ref{sec:construction} instantiate two conventions
for placing data near a latent manifold. The \emph{additive ambient Gaussian}
(spike-model) convention corrupts a low-dimensional signal $S$ by Gaussian noise,
$Y=S+\Sigma^{1/2}X$, with inference on $YY^\top/N$
\cite{Paul2007, DonohoGavish2014} and its heteroskedastic, two-dataset
\cite{ChenMa2025, YanChenFan2024}, and integral-operator \cite{DingMa2023}
refinements; our RSM is its sphere-native analogue, spectral-dual through shared
nonzero eigenvalues. The \emph{spectral Gram matrix} convention specifies the
noise in the cosine kernel through a Gegenbauer profile in the BBS eigenbasis
\cite{MardiaJupp2000, ElKaroui2010, Bordenave2012}; our FSM is model-free in this
sense. Methods acting directly on the point cloud or pairwise distances, such as
locally linear embedding \cite{WuWu2018}, interpoint-distance two-sample tests
\cite{Li2018}, and distance-covariance spectral analysis \cite{LiWangYao2023},
make no distributional noise assumption and are complementary. For a review of
manifold learning, see e.g.\ \cite{MeilaZhang2024}. The geometry of probability
spaces underlying the BBS continuum limit and the integral-operator analyses
\cite{DingMa2023, DingMa2025JASA} is developed in \cite{SmaleZhou2009}.

\section{BBS theory in a nutshell}
\label{sec:bbs-crash}

We first fix dimension conventions.
Throughout, $\R^{n}$ is the $n$-dimensional Euclidean space and
$S^{n-1}=\{u\in\R^{n}:\|u\|=1\}$ its unit sphere, a manifold of intrinsic
dimension $n-1$. The ambient data live on $S^{D-1}\subset\R^{D}$.
The orthogonal split $\R^{D}=\R^{d}\oplus\R^{D-d}$ places the latent signal on
$S^{d-1}\subset\R^{d}$ and the residual on $S^{D-d-1}\subset\R^{D-d}$; we call
$d$ the latent (embedding) dimension, so the latent manifold $S^{d-1}$ has
intrinsic dimension $d-1$.

Two papers underlie the analysis. BBS-1 \cite{bogomolny2003} derives the spectrum
of distance matrices on $N$ random points drawn from a base manifold $X$, with
the delocalised and localised power-law exponents set by the \emph{intrinsic}
dimension of $X$ (its worked circle $S^{1}$ is the one-dimensional case, with the
circle symmetry producing quasi-doublets). BBS-2 \cite{bogomolny2007} extends
this with the isometric-embedding problem, the negative-type (algebraic sign)
constraint, and a detailed analysis of spherical spaces $S^{d-1}\subset\R^{d}$
defined by $x_1^2+\dots+x_d^2=1$ (BBS-2 \cite{bogomolny2007} Eq.~(53)), from which
we take the multiplet structure used here. Our base manifold is the sphere
$S^{d-1}$ itself. Following the BBS-2 convention we call $d$ the embedding
dimension, so $S^{d-1}$ has intrinsic dimension $d-1$, and the BBS-1 exponents
(stated there for the intrinsic dimension) are evaluated here at intrinsic
dimension $d-1$. We work
throughout with the geodesic distance matrix
$M_{ij}=\arccos(\vec x_i\!\cdot\!\vec x_j)$ on unit-sphere samples
$\vec x_i\in S^{n-1}$ (the BBS-2 form (b) distance), which is full rank
generically and of \emph{negative type}: it has exactly one positive Perron
eigenvalue and $N-1$ non-positive ones, unconditionally, with no separate
verification needed. Two algebraic facts carry into the construction of
Sec.~\ref{sec:construction}. First, a convex combination of unit-diagonal
positive semidefinite Gram matrices is again such a Gram matrix, namely the
Gram matrix of the direct-sum embedding with the weights as block scalings, and
its entrywise $\arccos$ is therefore a valid angular-distance matrix of
negative type. Second, the squared-chord matrix is additive
under an orthogonal split $\R^{D}=V_1\oplus V_2$, $\vec X_i=\vec x_i+\vec y_i$,
\begin{equation}
\|\vec X_i - \vec X_j\|^{2}
= \|\vec x_i - \vec x_j\|^{2} + \|\vec y_i - \vec y_j\|^{2}.
\label{eq:bbs-pyth-additivity}
\end{equation}
The BBS spectral diagnostics live on the full-rank geodesic matrix $M$; the
squared-chord matrix of unit vectors is rank-deficient
($\mathrm{rank}\leq n+1$) and serves only as an algebraic intermediate.

The non-positive spectrum splits into two branches.
Throughout this paper, and following BBS-1 and BBS-2, we set the leading Perron
eigenvalue (the only positive eigenvalue of a distance matrix) aside and focus on
the remaining non-positive eigenvalues, which we sort by descending magnitude
$|\Lambda_K|$; this magnitude index is the natural one, as the BBS power laws and
multiplets live in the magnitudes of the negative-type spectrum. The spectrum
then splits into two physically distinct branches. The
\emph{delocalised} branch is the band of the roughly $\sqrt N$ largest-$|\Lambda|$
eigenvalues, whose eigenvectors are extended, plane-wave-like modes over the
manifold. The \emph{localised} branch is the small-$|\Lambda|$ remainder at
larger rank $K$, whose eigenvectors concentrate on close pairs of points. The
two meet at a crossover near $|\Lambda|\sim N^{-1/(d-1)}$.

Symmetry organises the spectrum into multiplets.
When $X$ has a non-trivial isometry group $G$, the spectrum of $M$ inherits the
irreducible-representation structure of $G$. While in the strict limit $N\to\infty$
eigenvalues are exactly degenerate within each
irreducible representation, at finite $N$ the degeneracy is lifted by an
$O(N^{-1/2})$ perturbation (see Appendix~\ref{sec:finite-N}), producing a
\emph{quasi-multiplet} of size the representation dimension at each unperturbed
eigenvalue. For $X=S^{d-1}$ the isometry group is $SO(d)$, the irreducible
representations are labelled by integer $\ell\geq 0$ with basis the
degree-$\ell$ spherical harmonics, and the multiplicity is
(BBS-2 \cite{bogomolny2007} Eq.~(58))
\begin{equation}
h(\ell, d)
= \frac{(2\ell + d - 2)\,(\ell + d - 3)!}{\ell!\,(d - 2)!}.
\label{eq:harmonics-mult}
\end{equation}
For every $d$, $h(0,d)=1$ (the constant mode); $h(\ell,2)=2$ for $\ell\geq 1$
(Fourier pairs on $S^1$), $h(\ell,3)=2\ell+1$ on $S^2$ (multiplets
$1,3,5,7,\ldots$), and $h(\ell,4)=(\ell+1)^2$ on $S^3$. These appear as the
visible plateaus at the top of the rank-ordered spectrum, the lowest giving the
integer readout of the latent dimension.

Next we present the BBS power laws on each branch.
\label{sec:bbs-power-laws}
In the continuum limit the discrete operator $(Mu)_i$ becomes an integral
operator, and solving its eigenvalue problem gives a power law on each branch.
On the delocalised branch BBS-1 solve the continuum eigenvalue problem with a
large-wavevector plane-wave ansatz. The plane-wave dispersion is
$\Lambda(q)\propto -(N/V)\,q^{-d}$ (BBS-1 Eq.~(62) at intrinsic dimension $d-1$),
and the density of plane-wave states on the $(d-1)$-dimensional sphere is
$\rho(q)\propto q^{d-2}$ (BBS-1 Eq.~(63)). Integrating $\rho(q)$ up to
$q(\Lambda)$ gives the counting function of negative eigenvalues
(BBS-1 Eq.~(64) at intrinsic dimension $d-1$)
\begin{equation}
\mathbf N(\Lambda) \approx
C_d\,\Bigl(\tfrac{N}{V}\Bigr)^{-1/d}\,(-\Lambda)^{-(d-1)/d},
\label{eq:BBS64}
\end{equation}
so the rank-ordered eigenvalues decay as
$|\Lambda_K|\propto K^{-d/(d-1)}$, a delocalised slope
$\beta_{\rm deloc}=d/(d-1)$ set by the dimension $d$ alone (the counterpart of
the localised $\beta_{\rm loc}$ below).
On the localised branch, a close pair at separation $r$ contributes
an eigenvalue $\Lambda\approx -r$ (BBS-1 Eq.~(99)); counting close pairs on the
intrinsic $(d-1)$-dimensional sphere gives the cumulative localised count
(BBS-1 Eq.~(100) at intrinsic dimension $d-1$)
\begin{equation}
\mathbf N_{\rm loc}(|\Lambda|)
= \frac{N^2\,\omega_{d-2}}{2V\,(d-1)}\,|\Lambda|^{d-1}
\;\propto\; N^2\,|\Lambda|^{d-1},
\label{eq:BBS100}
\end{equation}
i.e.\ a localised counting exponent $\beta_{\rm loc}=1/(d-1)$, with
$|\Lambda|\propto \mathbf N_{\rm loc}^{1/(d-1)}$ for the smallest eigenvalues. The
$O(1/\sqrt N)$ finite-$N$
correction of Appendix~\ref{sec:finite-N} applies to the delocalised branch; on
the localised branch the relative per-eigenvalue fluctuation is $O(1)$, so
$\beta_{\rm loc}=1/(d-1)$ stands only as a leading-order pair-counting statement.

\begin{figure}[!htbp]
\centering
\includegraphics[width=0.82\textwidth]{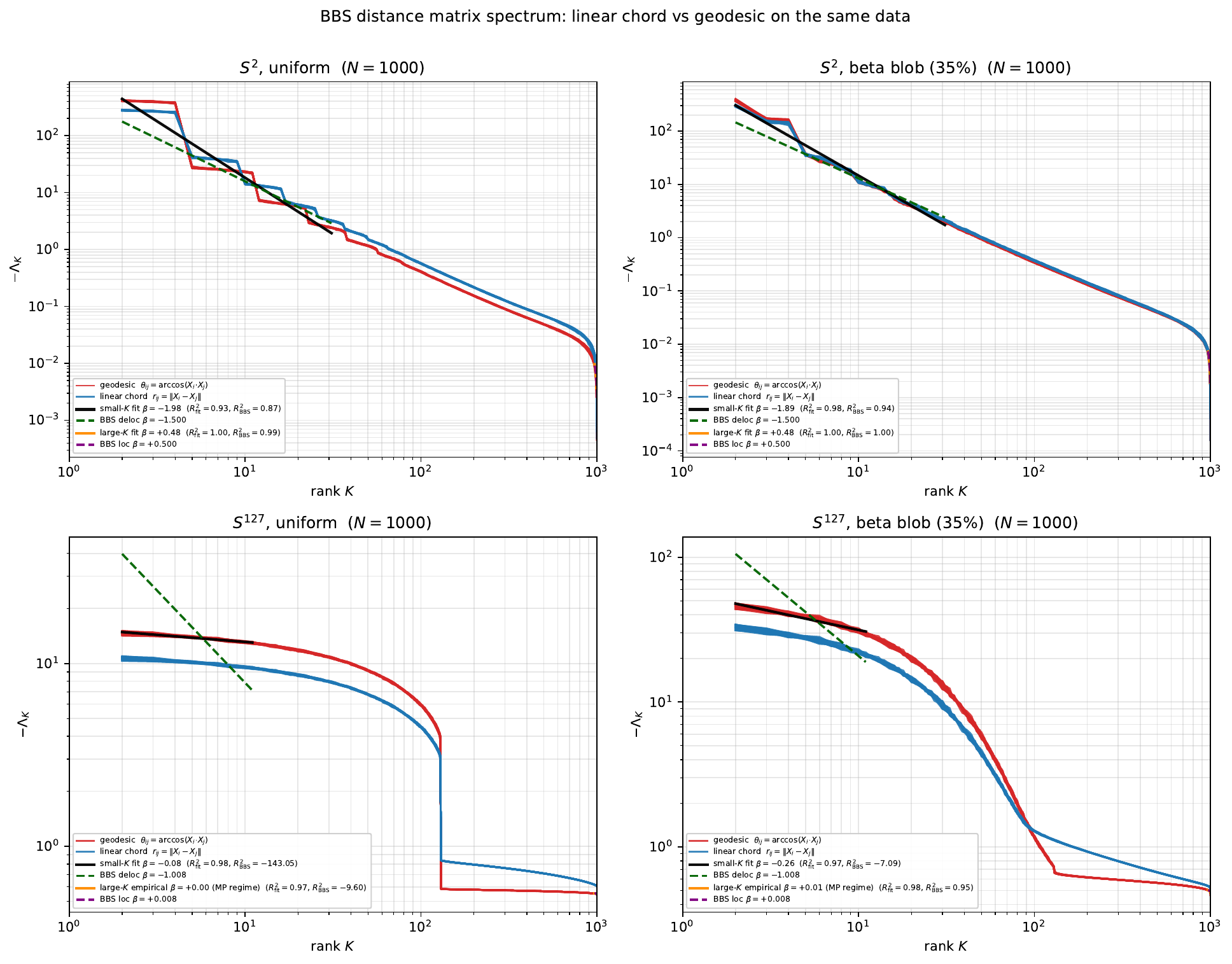}
\caption{BBS distance-matrix spectrum on $N=1000$ points on $S^2$ and
$S^{127}$, under uniform sampling (left) and a concentrated beta-product
blob (right, $35\%$ sphere coverage); 20
realisations of the geodesic (red) and linear-chord (blue) distances overlaid,
with the Perron mode at $K=1$ omitted so the spectrum is plotted from $K=2$. \emph{Top row} ($S^2$, BBS regime): the small-$K$
delocalised slope is near the BBS value $-d/(d-1)$ (up to the finite-$N$ shift of
Appendix~\ref{sec:finite-N}) and the $SO(3)$ multiplets appear as plateaus of
size $1,3,5,7,\ldots$. \emph{Bottom row} ($S^{127}$, MP regime): a dominant,
weakly decaying $\ell=1$ band of size $D+1=128$ then a sharp drop; the delocalised window
$[2,\sqrt N]$ does not extend past this first multiplet ($\sqrt N\approx31<128$),
so BBS-1 does not describe this spectrum and the BBS reference lines are
shown only to make the regime gap explicit.}
\label{fig:bbs-spectrum-both}
\end{figure}

\begin{figure}[!htbp]
\centering
\includegraphics[width=0.82\textwidth]{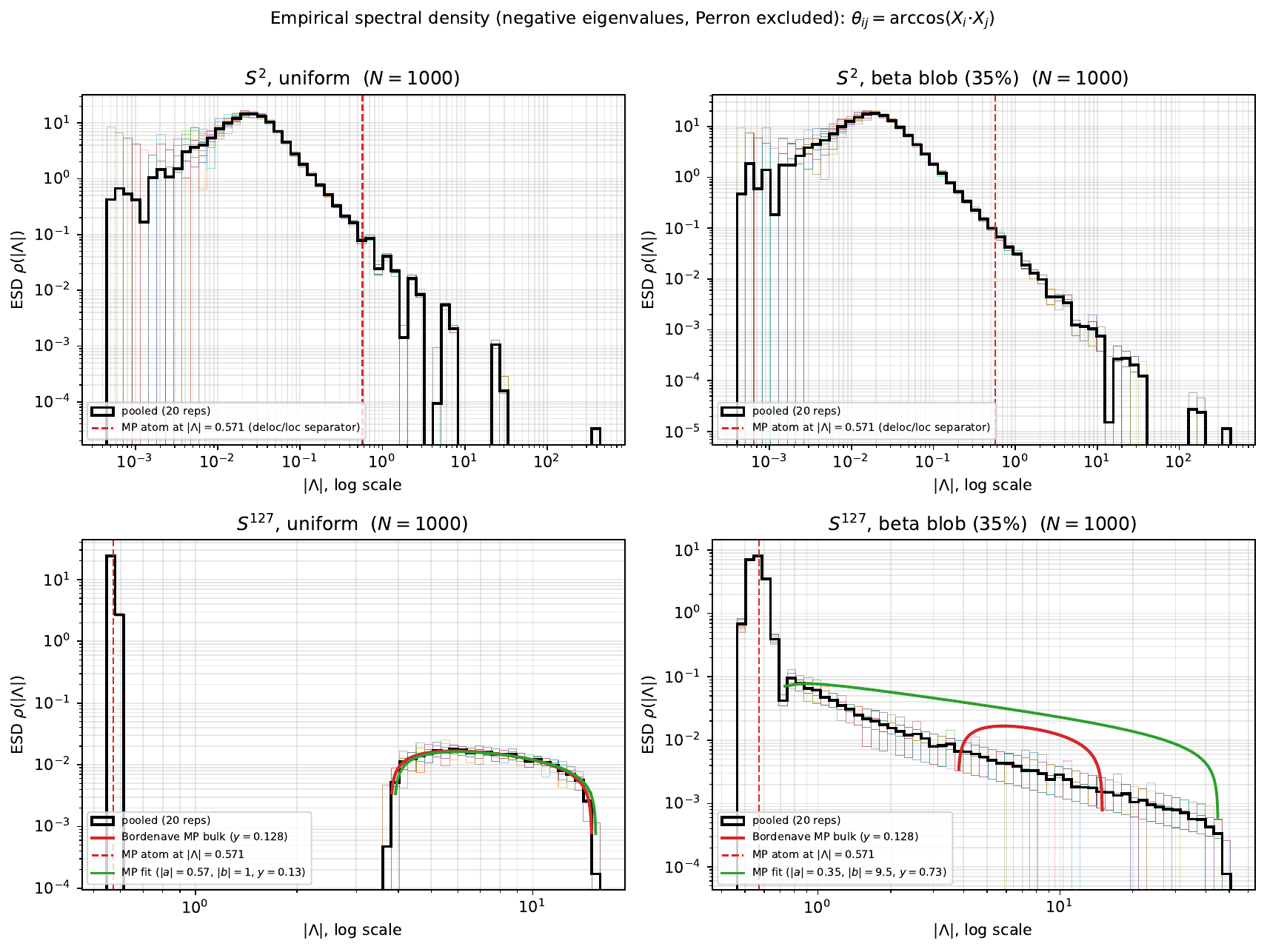}
\caption{Empirical spectral density of the negative eigenvalues of the geodesic
matrix (Perron excluded) for the data of Fig.~\ref{fig:bbs-spectrum-both}, pooled
over 20 realisations. Left column: uniform sampling; right column: the
concentrated beta-product blob. Red: the Bordenave--MP prediction
\eqref{eq:bordenave-mp-prediction}, matching the uniform $S^{127}$ (MP-regime)
bulk end-to-end and far outside the $S^2$ (BBS-regime) range; on $S^2$ the
Bordenave atom at $|\Lambda|=|1-\pi/2|$ (dashed, top row) marks the
delocalised/localised crossover. Green (both $S^{127}$ panels): an MP fit with
the aspect ratio $y$ fitted alongside the affine shift and scale; uniform
recovers $y\approx0.13$ and matches the bulk, while the concentrated blob
broadens to $y\approx0.73$ and approximately tracks the long, slowly-decaying
bulk, a deformed Marchenko--Pastur law
(Appendix~\ref{sec:appendix-beta-sphere}). The red uniform Bordenave bulk is a
reference and does not describe the concentrated case.}
\label{fig:bbs-esd-geodesic}
\end{figure}

We contrast the BBS regime with the Marchenko--Pastur limit.
The delocalised power law lives on the first $\sqrt N$ largest-magnitude
negative eigenvalues, the window $K\in[2,\sqrt N]$ (only the single positive
Perron eigenvalue at $K=1$ is excluded). This is the content of the BBS-1
validity condition (BBS-1 Eq.~(91)): the continuum approximation reproduces the
non-averaged eigenvalues only above a scaled-eigenvalue threshold that selects
the first $\sqrt N$ states. For the sphere $S^{d-1}$ this threshold reads
$|\lambda|\gtrsim N^{d/(2(d-1))}$ (BBS-1 Eq.~(91) at intrinsic dimension $d-1$).
Below it the spacing between successive continuum eigenvalues becomes comparable
to the finite-$N$ fluctuations and states mix; the quasi-multiplets cover the
same first $\sqrt N$ band. The power law is a clean, fittable decay only when this window extends
past the first ($\ell=1$) multiplet, which is $h(1,d)=d$ eigenvalues wide: this
needs $\sqrt N\gtrsim d$, i.e.\ $N\gtrsim d^{2}$. We call this the
\emph{BBS regime}: finite dimension $d$ and large sample size $N$ with
$N\gg d^{2}$. Most of the
random-matrix literature (Bordenave~\cite{Bordenave2012}, El~Karoui~%
\cite{ElKaroui2010}, and others) instead works in the
\emph{Marchenko--Pastur (MP) limit} $d,N\to\infty$ at fixed ratio $N/d$, where
the empirical spectral density of a smooth distance matrix
$A_{ij}=f(\|\vec X_i-\vec X_j\|^2)$ converges to an atom plus a Marchenko--Pastur
bulk,
\begin{equation}
\Lambda = f(0) - f(2) + 2f'(2) - 2f'(2)\,S,
\qquad S\sim\nu_{MP},
\label{eq:bordenave-mp-prediction}
\end{equation}
with $f(x)=\arccos(1-x/2)$ for the geodesic distance and $\nu_{MP}$ the MP law at
aspect ratio $y=(D+1)/N$ (so $y<1$ in the BBS regime $N\gg D$, matching the
fitted $y\approx0.13$ at $D=128$, $N=1000$ of Fig.~\ref{fig:bbs-esd-geodesic}). In this limit the two BBS branches map onto the two parts
of the law: the discrete delocalised multiplets spread out into the
Marchenko--Pastur bulk (the $-2f'(2)\,S$ term, the largest-$|\Lambda|$ band),
while the localised branch collapses onto the single atom at
$\Lambda=f(0)-f(2)+2f'(2)=1-\pi/2$ (the smallest-$|\Lambda|$ delta-peak). The
two regimes describe different corners of parameter space, and the same
construction sits in either depending on $d$ relative to $N$. We work at moderately large $N$, e.g.\ $N=1000$, which satisfies $N\gg d$
for a low-dimensional latent ($S^2$, $d=3$, well inside the BBS regime) but not
for a high-dimensional sphere ($S^{127}$, $d=128$, where $N<d^2$ puts it in the
MP regime). Figure~\ref{fig:bbs-spectrum-both} contrasts the two: on $S^2$ the
delocalised slope follows the BBS prediction $-d/(d-1)$ with the $SO(3)$
multiplet plateaus $1,3,5,\ldots$ at the top, while on $S^{127}$ the spectrum is
a single dominant, only weakly decaying band of the largest $\sim128$ eigenvalues
(the $\ell=1$ multiplet, spread out in the MP regime) ending in a sharp drop, the
delocalised window $[2,\sqrt N]$ not reaching past this first multiplet
($\sqrt N\approx31<128$).
Figure~\ref{fig:bbs-esd-geodesic} overlays the
Bordenave--MP prediction \eqref{eq:bordenave-mp-prediction} on the geodesic ESD:
it matches the $S^{127}$ bulk end-to-end while landing far outside the $S^2$
range, confirming the regime split. Ensemble fluctuations of the non-Perron
eigenvalues are at the few-percent level throughout, which is what lets us use a
single observed spectrum, rather than an ensemble average, as the inference
target.

\section{Generative models for the ambient space data}
\label{sec:construction}

We start with fixing the notation used in this section.
We work with the geodesic distance matrix
\[
M^{(n)}_{ij} \;:=\; \arccos(\vec u_i\!\cdot\!\vec u_j)
\;\in\;[0,\pi],
\]
on $N$ unit vectors $\vec u_i,\vec u_j\in S^{n-1}
\subset\R^{n}$, the BBS-2 form (b) distance of
Sec.~\ref{sec:bbs-crash}. The parenthesised
superscript denotes the dimension of the ambient
Euclidean space hosting the sphere: $M^{(D)}$ is
the ambient geodesic matrix on $S^{D-1}\subset\R^{D}$,
$M^{(d)}$ the latent geodesic matrix on
$S^{d-1}\subset\R^{d}$, and (for the
$V_{1}\oplus V_{2}$ decomposition below)
$M^{(D-d)}$ the residual geodesic matrix on
$S^{D-d-1}\subset\R^{D-d}$. Diagonals vanish,
$M^{(n)}_{ii} = 0$. The squared-chord N-matrix
$N^{(n)}_{ij} = \|\vec u_i - \vec u_j\|^{2} =
4\sin^{2}(M^{(n)}_{ij}/2)$ appears below as an
algebraic intermediate in the RSM decomposition,
where the dimensional additivity
\eqref{eq:bbs-pyth-additivity} is exploited; the
BBS spectral diagnostics themselves are read off
$M^{(n)}$. The $\ell$-th
angular-momentum multiplet on $S^{n-1}$ has dimension
$h(\ell,n)$ \eqref{eq:harmonics-mult} (BBS Eq.~(58)), giving $h(1,d) = d$ on
the latent $S^{d-1}$ and $h(1,D) = D$ on the
ambient $S^{D-1}$.
The latent unit vectors on $S^{d-1}$ are written
$\tilde x_i$ and the ambient unit vectors on
$S^{D-1}$ as $X_i$; $W_{0}\in\R^{D\times d}$ is the
noiseless isometric embedding so that
$X_i\big|_{\varepsilon=0} = W_{0}\tilde x_i$, with
$W_{0}^{\top}W_{0} = I_d$. The noise amplitude
$\varepsilon\in[0,1]$ is a single scalar; we work
at common per-particle amplitude. Throughout, $d$ denotes the
embedding dimension of the latent sphere $S^{d-1}$, whose
intrinsic manifold dimension is $d-1$, and we infer $d$
(equivalently $d-1$) through the first-multiplet multiplicity
$h(1,d)=d$. The lone exception is Table~\ref{tab:multiplet-table},
which is written in the intrinsic-dimension convention for the
beyond-sphere geometries and flags the switch in its caption.
We seek to infer $d\ll D$.

We classify embedding models into two classes by
how they specify the conditional law of $M^{(D)}$
given $M^{(d)}$, and summarise each by the matrix-
level forward (generative) drift--volatility
decomposition
\begin{equation}
M^{(D)}_{ij}
\;=\;
\mu_{ij}\bigl(M^{(d)},\varepsilon\bigr)
\;+\;
\sigma_{ij}\bigl(M^{(d)},\varepsilon\bigr)\,
G_{ij},
\qquad i\neq j,
\label{eq:layered-generic}
\end{equation}
with the noise vanishing at $\varepsilon = 0$
($\mu_{ij}\to M^{(d)}_{ij}$, $\sigma_{ij}\to 0$).
For model-specific drift $\mu_{ij}$, volatility
$\sigma_{ij}$, and centred noise $G_{ij}$, the
RSM (see Sec.~\ref{sec:aaag}) decomposition
admits a closed-form expression at the cosine-
kernel level through a linear convex-combination
identity, $\cos M^{(D)} = (1-\varepsilon^{2})
\cos M^{(d)} + \varepsilon^{2}\,\cos M^{(D-d)}(\bar y)$.%
\footnote{The residual cosine kernel is equivalently
the random Gram matrix
$\cos M^{(D-d)}_{ij}(\bar y) = G_{ij}(\bar y) :=
\bar y_i\!\cdot\!\bar y_j$ of the residual unit vectors
$\{\bar y_i\}\subset S^{D-d-1}$; we prefer the symmetric
form expressed through the residual distance matrix
$M^{(D-d)}$.}
For FSM
(see Sec.~\ref{sec:skm}) the drift $\mu_{ij}$ and volatility
$\sigma_{ij}$ are read off a positive-definite zonal kernel built
from a nonnegative spectral array $\{\beta_{pq}\}$, so the noise
is correlated across pairs sharing an index rather than
pair-diagonal, and $\cos M^{(D)}$ stays a valid Gram matrix.

\subsection{Residual Sphere Mixture (RSM)}
\label{sec:aaag}
\label{sec:anisotropic-class}

Let $X_i, X_j\in S^{D-1}\subset\R^{D}$. Decompose
the ambient space into orthogonal subspaces
$\R^{D} = V_{1}\oplus V_{2}$ ($\dim V_{1} = d$,
$\dim V_{2} = D-d$), with $V_{1}$ the latent embedding
subspace (the range of $W_{0}$) and $V_{2} = V_{1}^{\perp}$
the residual subspace, and project
$X_i = x_i + y_i$ with $x_i\in V_{1}$, $y_i\in V_{2}$, and
$\|x_i\|^{2} + \|y_i\|^{2} = 1$.
The orthogonality $V_{1}\perp V_{2}$ gives the
chord Pythagorean identity in $\R^{D}$,
\begin{equation}
\|X_i - X_j\|^{2}
\;=\;
\|x_i - x_j\|^{2}
\;+\;
\|y_i - y_j\|^{2}.
\label{eq:aag-pythagoras-chord}
\end{equation}
The squared chord between two $S^{D-1}$ points
decomposes exactly as the sum of the squared
chords of its two orthogonal projections. This is
the algebraic identity that the RSM construction
exploits; the BBS spectral matrix is the geodesic
one obtained from this via the arccosine
transformation. The projections $\{x_i, y_i\}$ have per-particle
norms summing to one but neither is unit-norm. We
rescale each projection to a unit vector on its
sub-sphere with a single common amplitude
$\varepsilon\in[0,1]$,
\begin{equation}
x_i \;=\; \sqrt{1-\varepsilon^{2}}\,W_{0}\tilde x_i,
\qquad
y_i \;=\; \varepsilon\,V_{0}\bar y_i,
\label{eq:aag-uv-parametrisation}
\end{equation}
with $\tilde x_i\in S^{d-1}$ the latent unit vector
on the $V_{1}$ sub-sphere, $\bar y_i\in S^{D-d-1}$
the residual unit vector on the $V_{2}$ sub-sphere,
and $W_{0}\in\R^{D\times d}$, $V_{0}\in\R^{D\times
(D-d)}$ fixed Stiefel frames
($W_{0}^{\top}W_{0} = I_d$, $V_{0}^{\top}V_{0} =
I_{D-d}$). The amplitude $\varepsilon$ controls the
noise level: $\varepsilon = 0$ collapses $X_i$ onto
its noiseless latent isometric image
$W_{0}\tilde x_i$; $\varepsilon = 1$ pushes $X_i$
fully onto the residual sub-sphere.

Because both factors of $x_i, x_j$ carry the common
scalar $\sqrt{1-\varepsilon^{2}}$ and both factors
of $y_i, y_j$ carry $\varepsilon$, orthonormality
of $W_{0}, V_{0}$ gives
$\|x_i - x_j\|^{2} = (1-\varepsilon^{2})\,\|\tilde x_i - \tilde x_j\|^{2}
= 2(1-\varepsilon^{2})\,(1 - \cos M^{(d)}_{ij})$ and
$\|y_i - y_j\|^{2} = \varepsilon^{2}\,\|\bar y_i - \bar y_j\|^{2}
= 2\varepsilon^{2}\,(1 - \cos M^{(D-d)}_{ij}(\bar y))$. Substituting into the chord
Pythagorean identity
\eqref{eq:aag-pythagoras-chord} and the
corresponding identity for the ambient sphere
$\|X_i - X_j\|^{2} = 2 - 2\cos M^{(D)}_{ij}$, and
cancelling the constant offset yields the RSM
generative form at the cosine-kernel level
\begin{equation}
\cos M^{(D)}_{ij}
\;=\;
(1-\varepsilon^{2})\,\cos M^{(d)}_{ij}
\;+\;
\varepsilon^{2}\,\cos M^{(D-d)}_{ij}(\bar y),
\label{eq:aag-MD-cosine}
\end{equation}
\labelalias{eq:aag-final-kernel-W}{eq:aag-MD-cosine}
\labelalias{eq:aag-final-kernel}{eq:aag-MD-cosine}
\labelalias{eq:aag-common-elev}{eq:aag-MD-cosine}
\labelalias{eq:aag-half-angle}{eq:aag-MD-cosine}
\labelalias{eq:aag-half-angle-smallangle}{eq:aag-MD-cosine}
The forward map \eqref{eq:aag-MD-cosine} is
an exact linear convex combination of two
unit-diagonal PSD matrices: the latent cosine
kernel $\cos M^{(d)}$ on the latent sub-sphere
$S^{d-1}\subset V_{1}$ (deterministic in
$\{\tilde x_i\}$) and the residual cosine kernel
$\cos M^{(D-d)}(\bar y)$ on the residual sub-sphere
$S^{D-d-1}\subset V_{2}$ (stochastic, specified
below). It is convenient to introduce the mixing angle
$\theta\in[0,\pi/2]$ defined by $\sin\theta = \varepsilon$,
so that $1-\varepsilon^{2} = \cos^{2}\theta$ and
$\varepsilon^{2} = \sin^{2}\theta$. The geodesic ambient
matrix is then the entry-wise arccosine,
\begin{equation}
M^{(D)}_{ij}
\;=\;
\arccos\!\bigl[\cos^{2}\theta\,\cos M^{(d)}_{ij}
+ \sin^{2}\theta\,\cos M^{(D-d)}_{ij}(\bar y)\bigr],
\qquad \varepsilon^{2} = \sin^{2}\theta,
\label{eq:aag-MD-geodesic}
\end{equation}
where $\arccos[\,\cdot\,]$ applies element-wise. The geodesic ambient matrix
is generically full-rank at finite $N$, of negative type by item
(c) of Sec.~\ref{sec:bbs-crash}, and the matrix on which
the BBS spectral diagnostics live. The diagonal
$M^{(D)}_{ii} = 0$ is automatic from
$\cos^{2}\theta + \sin^{2}\theta = 1$.

Expanding the mixing-angle kernel
\eqref{eq:aag-MD-geodesic} through the addition theorem for
Gegenbauer polynomials (see e.g.\ Eq.~(10.10.34) in
Erd\'elyi et al.~\cite{Erdelyi1953v2}) gives a double
Gegenbauer expansion in the latent and residual cosine
kernels (see Appendix~\ref{sec:appendix-addition}),
\begin{equation}
M^{(D)}_{ij}
\;=\;
\sum_{p,q\ge0}\Phi_{pq}(\theta)\,
C_p^{\mu}\!\bigl(\cos M^{(d)}_{ij}\bigr)\,
C_q^{\rho}\!\bigl(\cos M^{(D-d)}_{ij}(\bar y)\bigr),
\qquad
\mu=\tfrac{d-2}{2},\quad
\rho=\tfrac{D-d-2}{2},
\label{eq:aag-addition}
\end{equation}
where $\mu$ and $\rho$ are the ultraspherical
indices that set the degree of the Gegenbauer polynomials
$C^{\mu}_p$ and $C^{\rho}_q$ on the latent sphere $S^{d-1}$ and
the residual sphere $S^{D-d-1}$, respectively. Because
$\arccos(w)-\pi/2$ is odd in $w$, the coefficients
$\Phi_{pq}(\theta)$ vanish unless $p+q$ is odd, the only
even term being the mean $\Phi_{00}=\pi/2$. Averaging over
the uniform residual $\{\bar y_i\}$ projects
\eqref{eq:aag-addition} onto $q=0$, since
$\langle C_q^{\rho}(\cos M^{(D-d)}_{ij})\rangle=\delta_{q0}$,
and leaves a latent Gegenbauer series
\begin{equation}
\bigl\langle M^{(D)}_{ij}\bigr\rangle_{\bar y}
\;=\;
\frac{\pi}{2}
\;+\;
\sum_{p\ \mathrm{odd}}\Phi_{p0}(\theta)\,
C_p^{\mu}\!\bigl(\cos M^{(d)}_{ij}\bigr),
\label{eq:aag-addition-averaged}
\end{equation}
the constant $\Phi_{00}=\pi/2$ the mean angle (the Perron mode),
and each degree-$p$ term the BBS quasi-multiplet of multiplicity
$h(p,d)$ \eqref{eq:harmonics-mult}, shrunk from the noiseless
latent ($\theta=0$) by the parameter-free factor
$f_p(\theta)=\Phi_{p0}(\theta)/\Phi_{p0}(0)$, the
angular-momentum-level shrinkage law of
Sec.~\ref{sec:attenuation-law}.

Although \eqref{eq:aag-MD-cosine} is symmetric in
the two cosine-kernel terms, they play asymmetric
roles in the inference problem. The latent
configuration $\{\tilde x_i\}$ is treated as fixed
and observed through the deterministic geodesic
matrix $M^{(d)}$ on $S^{d-1}$; the residual unit
vectors $\{\bar y_i\}$ are treated as purely random
on $S^{D-d-1}$ with covariance $\Sigma_B$, and
enter through the random geodesic matrix
$M^{(D-d)}(\bar y)$ on $S^{D-d-1}$. The BBS theory
of Sec.~\ref{sec:bbs-crash} applies to both
sub-spheres: $M^{(d)}$ and $M^{(D-d)}$ are both
BBS distance matrices, the latent fixed in
$\{\tilde x_i\}$ and the residual random.
Conditional on $\{\tilde x_i\}$, the RSM forward
\eqref{eq:aag-MD-cosine} is the cosine-kernel
convex combination of these two BBS matrices, and
the geodesic ambient matrix
\eqref{eq:aag-MD-geodesic} is its entry-wise
arccosine.

For $\bar y_i$ uniform on $S^{D-d-1}$ the off-diagonal
entries of $\cos M^{(D-d)}$ are centred with variance
$1/(D-d)$ and a pair-diagonal covariance, the diagonal being
deterministic ($\cos M^{(D-d)}_{ii}=1$); these exact moments
(Marsaglia--Olkin~\cite{MarsagliaOlkin1984}) are recorded in
Appendix~\ref{sec:appendix-RSM},
Eq.~\eqref{eq:appG-moments-exact}. In the BBS-relevant regime
$N\gg D-d$, the cosine kernel $\cos M^{(D-d)}$ is
rank-deficient with rank $D-d$, and its non-zero
spectrum follows the Marchenko--Pastur law at
$y = (D-d)/N$ in the joint limit
$N, D-d\to\infty$ at fixed $y$
(see Appendix~\ref{sec:appendix-beta-saddle}).

\subsection{Free Spectral Mixture (FSM): model-free generalisation of RSM}
\label{sec:skm}
\label{sec:kernel-matrix-class}

The RSM forward \eqref{eq:aag-addition} is a double Gegenbauer
series in the two sub-sphere inner products, with coefficients
$\Phi_{pq}(\theta)$ fixed by the mixing angle
$\theta=\arcsin\varepsilon$. This form is dictated by symmetry,
not the RSM mechanism: the latent sphere $S^{d-1}$ carries an
$SO(d)$ invariance and the residual sphere $S^{D-d-1}$ an
independent $SO(D-d)$, so any ambient forward respecting both and
zonal under $SO(D-1)$ depends only on the two invariants
$\cos M^{(d)}_{ij}$ and $\cos M^{(D-d)}_{ij}$. The product
Gegenbauer functions $C_p^{(d-2)/2}\!\otimes C_q^{(D-d-2)/2}$ are
their joint Laplacian eigenbasis, so every
$SO(d)\times SO(D-d)$-invariant forward has the form
\eqref{eq:aag-addition}: the only model freedom is the coefficient
array.

The Free Spectral Mixture keeps the symmetry-mandated product
basis and frees the coefficient array, subject to the one
constraint of being a distance matrix: since
$M^{(D)}=\arccos(\cos M^{(D)})$ requires $\cos M^{(D)}$ to be a
Gram matrix, $\cos M^{(D)}\succeq 0$ with unit diagonal is
mandatory. We therefore free the spectrum at the cosine-kernel
level, where positivity is controlled, not the geodesic level
\eqref{eq:aag-addition}. The building blocks are the normalised
zonal kernels
\begin{equation}
\mathcal Z_p^{(n)}(t)
:=\frac{C_p^{(n-2)/2}(t)}{C_p^{(n-2)/2}(1)},
\qquad
\mathcal Z_p^{(n)}(1)=1,
\label{eq:fsm-zonal}
\end{equation}
the degree-$p$ Gegenbauer kernel on $S^{n-1}$. By the addition
theorem (BBS-2 \cite{bogomolny2007} Eq.~(57))
$\mathcal Z_p^{(n)}(\tilde x_i\!\cdot\!\tilde x_j)=
h(p,n)^{-1}\sum_m Y_{pm}(\tilde x_i)Y_{pm}(\tilde x_j)$ each
$\mathcal Z_p^{(n)}$ is a Gram matrix, and by Schoenberg's theorem
a zonal kernel is positive definite exactly when its Gegenbauer
coefficients are nonnegative \cite{Schoenberg1942}.

FSM is defined by a nonnegative coefficient array
$\{\beta_{pq}\}_{p,q\geq 0}$ on the product basis,
\begin{equation}
\cos M^{(D)}_{ij}
=\sum_{p,q\geq 0}\beta_{pq}\,
\mathcal Z_p^{(d)}\!\bigl(\tilde x_i\!\cdot\!\tilde x_j\bigr)\,
\mathcal Z_q^{(D-d)}\!\bigl(\bar y_i\!\cdot\!\bar y_j\bigr),
\qquad
\beta_{pq}\geq 0,
\quad
\sum_{p,q}\beta_{pq}=1,
\label{eq:fsm-psd-kernel}
\end{equation}
with $\tilde x_i\in S^{d-1}$ the latent and
$\bar y_i\in S^{D-d-1}$ the residual unit vectors drawn as in RSM.
The Schur product theorem makes each product
$\mathcal Z_p^{(d)}\odot\mathcal Z_q^{(D-d)}$ positive semidefinite
and the normalisation fixes the unit diagonal, so
$\cos M^{(D)}\succeq 0$ for every nonnegative array by construction
and $M^{(D)}=\arccos(\cos M^{(D)})$ is a genuine spherical distance
matrix (triangle inequalities and embeddability automatic). Nonnegativity is also
necessary by Schoenberg's theorem on each factor, so the FSM
simplex $\{\beta_{pq}\geq 0,\ \sum_{p,q}\beta_{pq}=1\}$ is a natural
complete positive class of $SO(d)\times SO(D-d)$-invariant kernels,
explored fully by freeing the coefficients.

RSM is one corner of this simplex. The kernel arguments are the
sub-sphere inner products
$\tilde x_i\!\cdot\!\tilde x_j=\cos M^{(d)}_{ij}$ and
$\bar y_i\!\cdot\!\bar y_j=\cos M^{(D-d)}_{ij}$, and the
degree-one zonal is the identity $\mathcal Z_1^{(n)}(t)=t$ (with
$\mathcal Z_0^{(n)}=1$). Taking $\beta_{10}=1-\varepsilon^{2}$,
$\beta_{01}=\varepsilon^{2}$ and the rest zero collapses
\eqref{eq:fsm-psd-kernel} to
$(1-\varepsilon^{2})\cos M^{(d)}_{ij}
+\varepsilon^{2}\cos M^{(D-d)}_{ij}$, the RSM identity
\eqref{eq:aag-MD-cosine}, so RSM populates only the degree-one
sectors; FSM switches on higher coefficients while keeping
positivity. Setting the noise amplitude to
$\varepsilon^{2}:=1-\beta_{10}$ (the weight off the latent
degree-one sector), FSM and isotropic RSM share the leading
top-$d$ cosine shrinkage $\E[r_K]=1-\varepsilon^{2}$ and the
$\eta_{\cos}$ calibration of Sec.~\ref{sec:attenuation-law}.

The construction is model-free in a precise sense: it commits to
the two rotational symmetries and to positivity, but to no
embedding mechanism, no per-particle map $\vec x_i\to\vec X_i$ and
no $\arccos$ expansion. A random FSM instance draws
$\{\beta_{pq}\}$ from a law on the simplex (in the experiments a
Dirichlet draw on a finite $(p,q)$ grid with the degree-one
weights set by $\varepsilon$), positive semidefinite pathwise. The
drift--volatility form \eqref{eq:layered-generic} still applies
entry by entry, but the fluctuations couple through shared indices
rather than being pair-diagonal, which is what keeps the kernel
positive semidefinite.

FSM and isotropic RSM differ in the angular-momentum component
they populate: switching on the even coefficient $\beta_{20}>0$
adds a degree-two zonal $\mathcal Z_2^{(d)}$, an $\ell=2$ component
the isotropic RSM kernel leaves empty by parity. This even-harmonic
signature drives the blind model-identification test of
Sec.~\ref{sec:exp-blind}.

\section{Inverse problem for the latent distance matrix}
\label{sec:perturbation-Dmatrix}
\label{sec:layer2-pert}

The inverse problem is to recover the latent BBS
structure of $M^{(d)}$ from a single observed ambient
matrix $M^{(D)}$, when the data lies \emph{near} a
low-$d$ sub-manifold of $S^{D-1}$ rather than exactly on
it.

\label{sec:pt-no-single-level}
The spectrum cannot be recovered eigenvalue by eigenvalue. At the cosine-kernel
level the RSM forward \eqref{eq:aag-MD-cosine} is a sum of two Gram matrices,
$\cos M^{(D)}=(1-\varepsilon^{2})\tilde X\tilde X^{\top}
+\varepsilon^{2}\bar Y\bar Y^{\top}$, but in the BBS regime its spectrum is not a
free additive convolution. With the latent dimension $d$ fixed and small the
latent term is a finite-rank deformation of the residual Gram matrix, a spiked
model governed by the Baik--Ben~Arous--P\'ech\'e and
Benaych-Georges--Nadakuditi theory of finite-rank perturbations
\cite{BaikBenArousPeche2005, BenaychGeorgesNadakuditi2011} rather than by
asymptotic freeness, which would need the two ranks to grow proportionally to
$N$ \cite{PasturVasilchuk2000, Zee1996, Mezard1999}.
In either regime the BBS structure lives in the geodesic matrix
$M^{(D)}=\arccos(\cos M^{(D)})$, a non-linear function that the Gegenbauer
addition theorem (see Appendix~\ref{sec:appendix-addition}) turns into
angular-momentum components of \emph{products} of latent and residual blocks,
so no additive deconvolution applies at
the level that carries the geometry. The ambient spectrum reorganises
collectively, fixed by a self-consistent resolvent rather than any single matrix
element \cite{Parisi2005}. A single-level perturbation theory is the wrong tool in this setting, since the
low-$K$ BBS quasi-multiplets are degenerate and require \emph{degenerate}
perturbation theory \cite{LandauLifshitz3, HoseKaldor1982}, which shifts each
multiplet as a whole (the L\"owdin--Schrieffer--Wolff block reduction
\cite{ArayaDay2025} of Sec.~\ref{sec:mult-invariant}, verified in
Appendix~\ref{sec:appendix-qdpt}).

The recovery therefore rests on two structural quantities that survive the
mixing, established below and tested numerically in
Sec.~\ref{sec:experiments}: the integer multiplicities of the low-lying
quasi-multiplets, gap-protected well into the strong-noise regime and carrying
the dimension directly (see Sec.~\ref{sec:mult-invariant}); and the multiplet
\emph{positions}, which shrink under noise by a parameter-free,
angular-momentum-dependent factor fixed by the Funk--Hecke spectrum of the
geodesic kernel (see Sec.~\ref{sec:attenuation-law}).

\subsection{The multiplet multiplicities as a gap-protected invariant}
\label{sec:mult-invariant}

The lowest non-Perron quasi-multiplets carry the isometry
group of the latent manifold through their integer
multiplicities $h(\ell,d)$ \eqref{eq:harmonics-mult}. On
$S^{d-1}$ the off-manifold noise is isotropic and does not
break $SO(d)$ on average, so it cannot split a degenerate
angular-momentum subspace at leading order. Within a multiplet the
perturbation acts as a near-degenerate block (L\"owdin /
Schrieffer--Wolff partitioning
\cite{Lowdin1962, SchriefferWolff1966}) that shifts the
multiplet as a whole while preserving its dimension. The
multiplicity is then protected by the spectral gap to the
neighbouring multiplets, and stays an integer until the
noise closes that gap. The product-kernel analysis of
Sec.~\ref{sec:product-spectrum} makes this precise: the residual
sectors that could mix into the dimension cluster sit at
$O(\varepsilon^2/(D-d))$, separated from it, so the $h(1,d)$-fold
degeneracy survives rather than being split.

Table~\ref{tab:mult-invariant} reports the recovered pattern
on the $S^{2}$ latent geometry, for the latent matrix and
for the RSM ambient matrix at the four noise levels of
Fig.~\ref{fig:aag-esd-noise-pollution}. Only the odd
degrees $\ell=1,3,5,\ldots$ contribute, since
$\arccos(t)-\pi/2$ is odd.
The first multiplet is a cleanly isolated triplet,
$h(1,3)=3$, the $S^{2}$ signature; the second is the
$\ell=3$ septet, $h(3,3)=7$. Both are reproduced exactly at
every noise level up to $\eta_{\cos}\approx 20\%$, while the
third multiplet ($\ell=5$) already overlaps the bulk: its
recovered multiplicity is $6$ rather than the ideal
$h(5,3)=11$, since the higher odd degrees $\ell=5,7,\ldots$
crowd together and merge with the bulk, an effect that
worsens as the noise grows. The gap protection is quantitative: Weyl's eigenvalue
inequality keeps each cluster isolated and the
Davis--Kahan $\sin\theta$ theorem keeps its eigenspace
coherent while $\|\delta M\|_{\rm op}<g/2$
(see Sec.~\ref{sec:multiplet-table}). The integer pattern is
thus the most stable carrier of the dimension and topology,
read off without any deconvolution.

\begin{table}[!htbp]
\centering
\caption{Multiplet sizes (and mean $|\Lambda|$) from the
$\log$-gap detector on $S^{2}$, for the latent matrix
and the RSM ambient matrix. The triplet and septet
multiplicities are preserved at all noise levels.
$N=1000$, $d=3$, $D-d=125$.}
\label{tab:mult-invariant}
\begin{tabular}{l ccc}
\hline
& $\ell=1$ & $\ell=3$ & $\ell=5$ (merging)\\
\hline
latent $M^{(d)}$ & $3\ (392.7)$ & $7\ (24.7)$ & $6\ (6.7)$\\
$\varepsilon=0.10$ & $3\ (385.7)$ & $7\ (22.6)$ & $6\ (5.5)$\\
$\varepsilon=0.22$ & $3\ (362.6)$ & $7\ (17.4)$ & $6\ (3.5)$\\
$\varepsilon=0.32$ & $3\ (334.4)$ & $7\ (12.9)$ & $6\ (2.4)$\\
$\varepsilon=0.45$ & $3\ (288.3)$ & $7\ (8.1)$ & $6\ (3.6)$\\
\hline
\end{tabular}
\end{table}

\subsection{The product-kernel spectrum and the multiplet shrinkage law}
\label{sec:attenuation-law}
\label{sec:product-spectrum}

A single observed matrix is one joint draw
$\{(\tilde x_i,\bar y_i)\}$, and as $N\to\infty$ the normalised
matrix $\tfrac1N M^{(D)}$ concentrates to the integral operator
$T_\varepsilon$ on $L^2(S^{d-1}\times S^{D-d-1})$ with kernel
\begin{equation}
K_\varepsilon=\arccos\!\bigl((1-\varepsilon^2)u+\varepsilon^2 v\bigr),
\qquad u=\tilde x\!\cdot\!\tilde x',\quad v=\bar y\!\cdot\!\bar y',
\label{eq:product-kernel}
\end{equation}

The eigenbasis is the product
$Y_{rm}(\tilde x)\,Y_{sn}(\bar y)$ of the degree-$r$ and degree-$s$
spherical harmonics on the two factors $S^{d-1}$ and $S^{D-d-1}$,
with within-degree labels $m=1,\dots,h(r,d)$ and
$n=1,\dots,h(s,D-d)$. Summed over a degree these products give the
normalised zonals $\mathcal Z^{(d)}_r$ and $\mathcal Z^{(D-d)}_s$
\eqref{eq:fsm-zonal}, where the addition theorem ties the harmonics
to the Gegenbauer polynomials.

Because $K_\varepsilon$ depends on the two points only through $u$
and $v$, it is invariant under $SO(d)$ on the latent factor and
$SO(D-d)$ on the residual factor, so $T_\varepsilon$ commutes with
$SO(d)\times SO(D-d)$ and, by Schur's lemma, acts as a single
scalar $a_{rs}(\varepsilon)$ on each degree-$(r,s)$ block,
independent of the within-degree labels $m,n$; the block is thus
$h(r,d)\,h(s,D-d)$-fold degenerate. By the Funk--Hecke theorem this
eigenvalue is the projection of the kernel $K_\varepsilon$
\eqref{eq:product-kernel} onto the product zonal
$\mathcal Z^{(d)}_r(u)\,\mathcal Z^{(D-d)}_s(v)$,
\begin{equation}
a_{rs}(\varepsilon)=\iint
\arccos\!\bigl((1-\varepsilon^2)u+\varepsilon^2 v\bigr)\,
\mathcal Z^{(d)}_r(u)\,\mathcal Z^{(D-d)}_s(v)\,
\omega_d(u)\,\omega_{D-d}(v)\,\dd u\,\dd v,
\label{eq:product-coeff}
\end{equation}
where $\mathcal Z^{(n)}_\ell\propto C_\ell^{(n-2)/2}$ is the
degree-$\ell$ zonal normalised to $\mathcal Z^{(n)}_\ell(1)=1$
\eqref{eq:fsm-zonal} and $\omega_n(t)\propto(1-t^2)^{(n-3)/2}$ is
the density of the inner product $t$ on $S^{n-1}$. For a single
factor this is the Funk--Hecke coefficient
\begin{equation}
a_\ell[\kappa]=\int_{-1}^{1}\kappa(t)\,\mathcal Z^{(d)}_\ell(t)\,
\omega_d(t)\,\dd t,
\label{eq:funk-hecke-coeff}
\end{equation}
the per-mode eigenvalue of the zonal kernel $\kappa$, shared by all
$h(\ell,d)$ modes of degree $\ell$; the kernel is positive definite
exactly when these coefficients are nonnegative (the Schoenberg
condition) \cite{SmolaOvariWilliamson2000}.

As the mean of the residual zonal vanishes above degree zero,
$\mathbb E_{\bar y}\bigl[\mathcal Z^{(D-d)}_s(v)\bigr]=\delta_{s0}$,
only the $s=0$ term survives averaging \eqref{eq:product-coeff} over
uniformly random residual coordinates $\{\bar y_i\}$ on
$S^{D-d-1}$. This produces the residual-averaged kernel
\begin{equation}
\kappa_{\rm obs}(t;\varepsilon)
=\int_{-1}^{1}
\arccos\!\bigl((1-\varepsilon^2)t+\varepsilon^2 s\bigr)\,
\omega_{D-d}(s)\,\dd s,
\label{eq:kappa-obs}
\end{equation}
zonal in $t$, with $a_{r0}(\varepsilon)=a_r[\kappa_{\rm obs}]$.
This $s=0$ column is the latent multiplet tower. Relative to the
noiseless latent kernel $\kappa_{\rm lat}(t)=\arccos t$ --- whose
even sector collapses to the Perron constant $a_0=\pi/2$
(BBS-2 \cite{bogomolny2007} Eq.~(64); $\Lambda_0\simeq N\pi/2$, set
aside) and whose non-Perron coefficients are nonzero only for odd
$\ell$, with $a_1=-\pi/8$ on $S^2$ --- each tower level is shrunk by
\begin{equation}
f_\ell(\varepsilon)
=\frac{a_\ell[\kappa_{\rm obs}(\cdot\,;\varepsilon)]}
{a_\ell[\kappa_{\rm lat}]},
\qquad
a_{r0}(\varepsilon)=f_r(\varepsilon)\,a_r[\kappa_{\rm lat}],
\label{eq:attenuation-law}
\end{equation}
the angular-momentum-level shrinkage law, fixed by
$(\varepsilon,d,D)$ with no free parameters. It is defined only on
the latent tower: $f_\ell$ is a ratio of observed to noiseless
coefficient, and only $s=0$ has a nonzero noiseless value, since
$a_{rs}(0)=a_r[\kappa_{\rm lat}]\,\delta_{s0}$. The residual sectors
$s\ge1$ vanish at $\varepsilon=0$, so they carry no latent value to
shrink from; switched on by the noise, they grow as
$O(\varepsilon^{2s})$, derived next. This is the
spiked-model eigenvalue shrinkage
\cite{Paul2007, DonohoGavish2014, Bun2017} resolved level by level
on the sphere, tied to the BBS multiplet structure by Funk--Hecke
rather than to a generic random-matrix result. ``Parameter-free''
refers to this shrinkage integral, which fits nothing to the
spectra; the pipeline as a whole carries the log-gap threshold
$\tau$, the candidate set, the finite-$N$ slope correction of
Appendix~\ref{sec:finite-N}, the rank cutoff $K_{\rm lat}$, and the
model-classification threshold. We evaluate $\kappa_{\rm obs}$ and
its coefficients by two nested Gauss--Legendre rules, an inner
average over $s$ and an outer projection onto
$\mathcal Z^{(d)}_\ell$ against $\omega_d$, exactly in
$\varepsilon$ with no series and no closed form needed.

A small-$\varepsilon$ expansion fixes how the columns scale.
Writing the argument as $u+\varepsilon^2(v-u)$,
\[
\arccos\!\bigl(u+\varepsilon^2(v-u)\bigr)
=\arccos u-\frac{\varepsilon^2(v-u)}{\sqrt{1-u^2}}+O(\varepsilon^4),
\]
the coefficient of $\varepsilon^{2k}$ is a degree-$k$ polynomial in
$v$ with no projection onto $\mathcal Z^{(D-d)}_s$ for $s>k$, so
$a_{rs}(\varepsilon)=O(\varepsilon^{2s})$ for $s\ge1$. The
$v$-independent piece $+\varepsilon^2 u/\sqrt{1-u^2}$ of the
first-order term stays in $s=0$, giving the attenuation
$\kappa_{\rm obs}=\arccos t+\varepsilon^2 t/\sqrt{1-t^2}+O(\varepsilon^4)$
and hence $f_\ell=1-O(\varepsilon^2)$ at small noise. The
$v$-linear piece $-\varepsilon^2 v/\sqrt{1-u^2}$ is the leading
residual sector,
\begin{equation}
a_{r1}(\varepsilon)=-\frac{\varepsilon^2}{D-d}\,g_r+O(\varepsilon^4),
\qquad
g_r=a^{(d)}_r\!\Bigl[\tfrac{1}{\sqrt{1-u^2}}\Bigr],
\label{eq:s1-coeff}
\end{equation}
with $1/(D-d)$ the degree-one eigenvalue of $v$ on $S^{D-d-1}$ and
$g_r$ the degree-$r$ coefficient of $1/\sqrt{1-u^2}$; since the
latter is even, $g_r$ vanishes for odd $r$, consistent with the
parity rule that $r+s$ is odd. The $s\ge1$ family thus sits at
$O(\varepsilon^2/(D-d))$, with multiplicity $h(r,d)\cdot(D-d)$,
seeding the residual bulk near zero (see
Sec.~\ref{sec:appendix-bulk}). The shrinkage law uses the full
integral \eqref{eq:kappa-obs} rather than this truncation, which is
non-uniform ($t/\sqrt{1-t^2}$ diverges at $t\to\pm1$ and the
high-$\ell$ levels leave the perturbative regime, $f_5\approx0.1$
at $\varepsilon=0.45$); the quadrature resums every order. The
quasi-degenerate L\"owdin--Schrieffer--Wolff reduction of
Appendix~\ref{sec:appendix-qdpt} reaches the same shrinkage on
$S^1,S^2,S^3$, a numerical cross-check of the resummation rather
than a proof.

The dimension cluster is the $(r,s)=(1,0)$ sector, the $h(1,d)=d$
eigenvalues at $a_{10}(\varepsilon)=f_1(\varepsilon)\,
a_1[\kappa_{\rm lat}]$ of magnitude $O(1)$. It is bounded below by
the next latent level $a_{30}=f_3\,a_3[\kappa_{\rm lat}]$ across the
BBS gap $g_{\rm BBS}=|a_1|f_1-|a_3|f_3$ that the multiplicity
readout of Sec.~\ref{sec:mult-invariant} already uses, and above by
the whole residual family at $O(\varepsilon^2/(D-d))$. Both gaps
stay open while $\varepsilon^2<\varepsilon_*^2(d,D-d)$, the smaller
of the level-crossing bound $f_1|a_1|=f_3|a_3|$ and the
residual-collision bound $f_1|a_1|=C_d\,\varepsilon^2/(D-d)$ with
$C_d=\max_r|g_r|=g_0$. For small $\varepsilon$ and large $D-d$ the
level crossing binds first: the residual family, at
$O(\varepsilon^2/(D-d))$, never reaches the $O(1)$ cluster. We
write $g_*$ for the resulting two-sided separation, positive while
$\varepsilon<\varepsilon_*(d,D-d)$.

Under uniform sampling on both factors with $d$ fixed, $D-d$
specified, $\varepsilon<\varepsilon_*(d,D-d)$, and $N\to\infty$, two
facts make this an inference statement rather than a
single-realisation observation. First, the empirical operator
converges,
$\|\tfrac1N M^{(D)}-T_\varepsilon\|_{\rm op}=O(\sqrt{\log N/N})$
\cite{SmaleZhou2009}. Second, once this deviation falls below
$g_*/2$, the Davis--Kahan theorem \cite{DavisKahan1970} fixes the
eigenvalue count in the cluster at $h(1,d)=d$ and bounds the
rotation of its empirical eigenspace by the deviation divided by
$g_*$. The recovered multiplicity is then exactly $d$ in the limit
and the cluster sits at $a_{10}(\varepsilon)$. The residual sectors
are genuine $O(\varepsilon^2/(D-d))$ eigenvalues of the product
operator, not $O(1/\sqrt N)$ fluctuations of the latent levels, so
the separation keeps them clear of the dimension cluster. This
discrete picture holds at fixed $(d,D-d)$; when $D-d$ grows
proportionally to $N$ the $s=1$ family broadens into the deformed
Marchenko--Pastur residual bulk (see Sec.~\ref{sec:appendix-bulk}),
whose edge still scales as $\varepsilon^2$ times an $O(1)$
aspect-ratio factor, so the separation persists at small
$\varepsilon$ but reads as a bulk-edge condition.

The multiplet position is read off the spectrum, blind, as the mean
$|\Lambda|$ of the $\ell$-th gap-detected cluster (see
Sec.~\ref{sec:mult-invariant}), needing no latent coordinates; when
those coordinates are available, in validation, the same quantity
is the zonal-projector estimate $\widehat\Lambda_\ell=N^{-1}
\sum_{ij} M_{ij}\mathcal Z^{(d)}_\ell(\tilde x_i\!\cdot\!\tilde x_j)$.
The shrinkage is sharply degree-dependent: at
$\eta_{\cos}\approx20\%$ the dipole retains $f_1=0.73$ while
$f_3=0.30$ and $f_5=0.10$, so noise erodes the high-$\ell$
structure first, the mechanism behind the noise-induced bias of the
delocalised rank-decay slope, which samples progressively higher
$\ell$ as $|\Lambda|$ decreases. Inverting recovers the latent
spectrum: with $d$ fixed by the multiplicity, the de-shrunk
positions $\Lambda_{\rm obs,\ell}/f_\ell(\varepsilon;d,D)$ are
required to fall back onto the clean BBS tower
$N\,a_\ell[\kappa_{\rm lat}]$ across the resolved degrees
$\ell=1,3,\dots$, and because the $f_\ell$ carry distinct
$\ell$-dependence this over-determined match fixes a single
$\varepsilon$, the lowest multiplets anchoring the fit.

Two points fix the status of this inversion. The shrinkage
\eqref{eq:attenuation-law} is a \emph{forward} spectral formula:
given $(d,D-d,\varepsilon)$ it predicts the level positions, and
the inversion reads those parameters back. It depends on the
residual dimension $D-d$ through $\omega_{D-d}$, so the ambient
dimension enters as an input that Algorithm~1 takes as $D$ (or, when
$D$ is unknown, $D-d$ joins the candidate set); ``coordinate-free''
means no ambient vectors are required, not that
$(d,D-d,\varepsilon)$ are left unspecified. With $d$ fixed, the map
$(\varepsilon,D-d)\mapsto\{f_\ell(\varepsilon;d,D-d)\}_\ell$ is, on
the candidate grid, injective across the resolved degrees, a
property we verify there rather than prove in general: $\varepsilon$
enters every level at $O(\varepsilon^2)$ while $D-d$ reshapes the
residual density and so the $\ell$-dependence of the higher levels,
and two resolved degrees overdetermine the pair, so distinct
$(\varepsilon,D-d)$ give distinct ratio vectors. When only the first multiplet is resolved this
identifiability is lost, and the method returns $d$ but not a unique
noise calibration.

\begin{figure}[!htbp]
\centering
\includegraphics[width=0.47\textwidth]{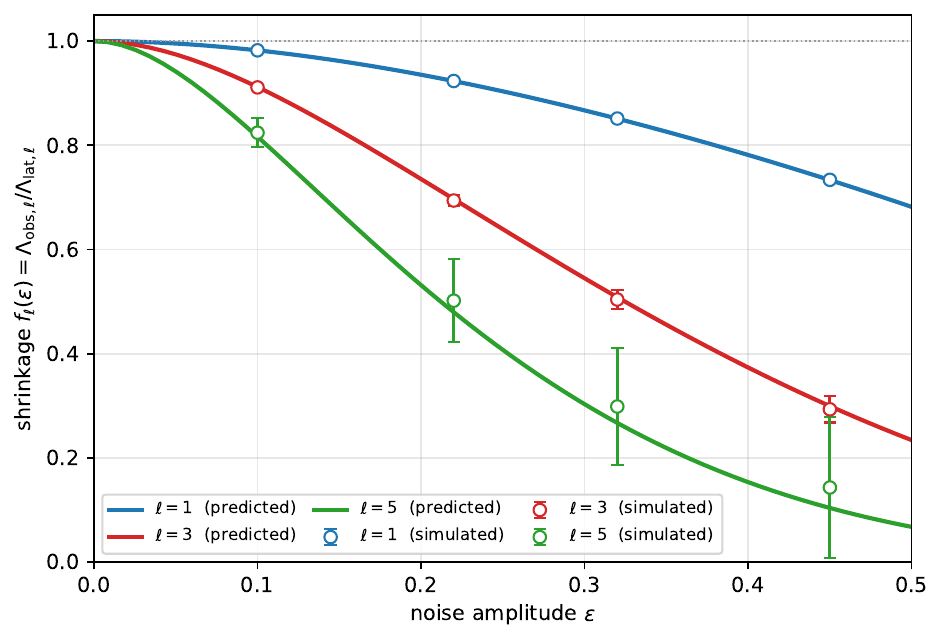}
\caption{Angular-momentum-level shrinkage $f_\ell(\varepsilon)$ for
degrees $\ell=1,3,5$: parameter-free prediction
\eqref{eq:attenuation-law} (lines) and simulated multiplet
shrinkage (markers) on the $S^{2}/S^{124}$ setup.}
\label{fig:attenuation-law}
\end{figure}

\begin{figure}[!htbp]
\centering
\includegraphics[width=0.95\textwidth]{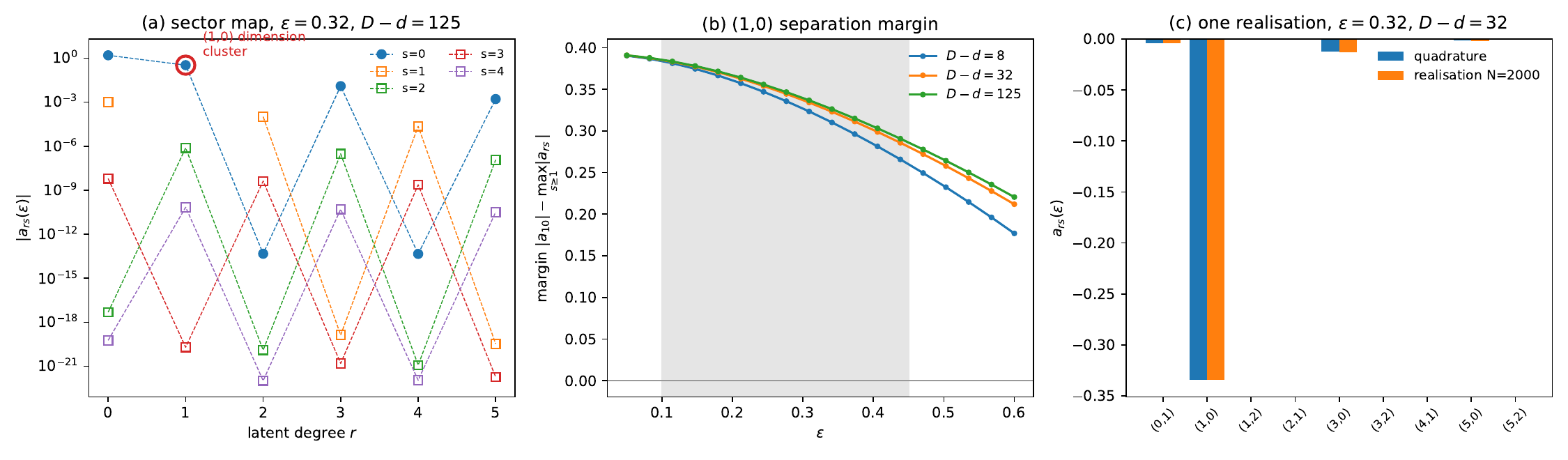}
\caption{Product-kernel spectrum of the RSM operator $T_\varepsilon$
on $S^{2}\times S^{D-d-1}$. (a) Sector map: the double Funk--Hecke
coefficients $a_{rs}(\varepsilon)$ \eqref{eq:product-coeff} by
$(r,s)$, with the $s=0$ latent tower (filled) and the
$s\ge1$ residual sectors (open) at magnitude
$O(\varepsilon^2/(D-d))$. (b) Separation margin of the $(1,0)$
dimension cluster, $|a_{10}|$ minus the largest competing
$|a_{rs}|$ with $s\ge1$, against $\varepsilon$ for several
$D-d$; the margin stays positive across the experimental noise
range. (c) Sectors resolved from a single sampled realisation
($N=2000$) against the quadrature prediction, confirming that one
matrix shows the full array, not the averaged column.}
\label{fig:product-spectrum}
\end{figure}

Table~\ref{tab:shrinkage} and Fig.~\ref{fig:attenuation-law}
compare the prediction \eqref{eq:attenuation-law} to the simulated
shrinkage on the $S^2/S^{124}$ setup, agreeing to three significant
figures for the resolved $\ell=1,3$ and within sampling scatter for
$\ell=5$. Fig.~\ref{fig:product-spectrum} evaluates the full array
$\{a_{rs}(\varepsilon)\}$ by the same quadrature extended to the
double projection \eqref{eq:product-coeff}, showing the sector map
and the $(1,0)$ separation margin versus $\varepsilon$ and $D-d$ and
confirming the cluster stays isolated across the experimental noise
range.

\begin{table}[!htbp]
\centering
\caption{Angular-momentum-level shrinkage $f_\ell(\varepsilon)$:
parameter-free prediction \eqref{eq:attenuation-law}
against the simulated ratio
$\Lambda_{\rm obs,\ell}/\Lambda_{\rm lat,\ell}$
($S^{2}/S^{124}$, $N=1000$, 12 realisations).}
\label{tab:shrinkage}
\begin{tabular}{c cc cc cc}
\hline
& \multicolumn{2}{c}{$\ell=1$}
& \multicolumn{2}{c}{$\ell=3$}
& \multicolumn{2}{c}{$\ell=5$}\\
$\varepsilon$ & pred. & sim. & pred. & sim. & pred. & sim.\\
\hline
$0.10$ & $0.982$ & $0.982$ & $0.912$ & $0.911$ & $0.816$ & $0.824$\\
$0.22$ & $0.923$ & $0.923$ & $0.697$ & $0.694$ & $0.480$ & $0.502$\\
$0.32$ & $0.851$ & $0.851$ & $0.508$ & $0.504$ & $0.267$ & $0.299$\\
$0.45$ & $0.733$ & $0.734$ & $0.299$ & $0.293$ & $0.104$ & $0.143$\\
\hline
\end{tabular}
\end{table}

\subsection{Implications for the BBS diagnostics}

The two recovery handles rest on different stabilities, not on the
single-level perturbation theory that diverges in the dense bulk. The integer
multiplicity is gap-protected by the Weyl and Davis--Kahan bounds while
$\|\delta M\|_{\rm op}$ stays below the inter-multiplet gap
(see Secs.~\ref{sec:mult-invariant} and~\ref{sec:multiplet-table}). The multiplet
positions are set by the resummed Funk--Hecke shrinkage of the noise-averaged
kernel (see Sec.~\ref{sec:attenuation-law}). The consequences for the three BBS
spectral diagnostics follow. On the \emph{delocalised branch}, the
angular-momentum-level shrinkage \eqref{eq:attenuation-law} leaves the leading
$K^{-d/(d-1)}$ decay on $K\in[2,\sqrt N]$ intact at small noise (the lowest
components shrink least). The slope bias enters through the faster erosion of the
higher-$\ell$ components and is calibrated by the finite-$N$ correction
$\Delta\beta(N,d)$ of Appendix~\ref{sec:finite-N}. For \emph{multiplet
sharpness}, the degree-dependent shrinkage draws neighbouring multiplets
together, so the multiplicity diagnostic fails once the noise closes the
inter-multiplet gap, beyond $\eta_{\cos}\approx 20\%$ on $S^{2}$
(see Sec.~\ref{sec:mult-invariant}). On the \emph{localised branch} (small
$|\Lambda|$) the relative per-eigenvalue fluctuation is $O(1)$ rather than
$O(1/\sqrt N)$ (see Appendix~\ref{sec:finite-N}), so the off-manifold noise competes
with the leading BBS prediction $\beta_{\rm loc}=1/(d-1)$ rather than being
subleading to it. The localised slope is therefore not a reliable dimension
estimator but instead fingerprints the ambient noise model, developed in
Sec.~\ref{sec:discussion} (see Fig.~\ref{fig:localized-branch}).

\subsection{Algorithm 1: I-BBS pipeline}
\label{sec:exp-recipe}

The diagnostics combine into Algorithm~1 below, framed as a
single coordinate-free pipeline for compactness. In practice
the steps are independent and usable separately depending on
the noise regime, the prior on the ambient model, and the
spectrum quality at the relevant rank windows; the numbered
ordering is one natural workflow, not a strict dependency
chain.

\begin{mdframed}[linewidth=0.8pt,
  innertopmargin=8pt, innerbottommargin=8pt,
  innerleftmargin=12pt, innerrightmargin=12pt]
\noindent\textbf{Algorithm 1: I-BBS pipeline.}
Latent sub-manifold identification from an ambient
distance matrix.

\smallskip
\noindent\textit{Input:} the ambient distance matrix
$M^{(D)}$, sample size $N$, ambient dimension $D$, log-gap
threshold $\tau$ (default $0.3$), candidate-$d$ set
$\mathcal D$. No ambient coordinates are required.\\
\textit{Output:} latent-dimension estimate $\hat d$ and
ambient noise-model verdict.

\smallskip
\begin{enumerate}
\setlength\itemsep{2pt}
\item Diagonalise $M^{(D)}$; sort eigenvalues by
descending $|\Lambda_K|$.
\item \textit{Multiplicity (primary):} apply the
$\log$-gap walk of Sec.~\ref{sec:exp-multiplets} to
the descending spectrum to read the lowest-multiplet
multiplicity $\hat h_1$, which fixes $\hat d_{h_1}$ through
the BBS relation $\hat h_1 = h(1,d)$. Gap-protected and
the primary handle.
\item \textit{Shrinkage (positions):} recover the latent
multiplet positions by inverting the parameter-free
angular-momentum-level shrinkage $f_\ell$ of
Sec.~\ref{sec:attenuation-law}; matching the shrinkage-corrected
multiplets to the BBS tower confirms $\hat d$ and gives the
latent spectrum.
\item \textit{Delocalised slope (cross-check):} fit the
rank-decay slope on $K \in [d_{\rm guess}, \sqrt N]$,
apply the finite-$N$ correction $\Delta\beta(N,
d_{\rm guess})$ from Eq.~\eqref{eq:delta-beta} and rule
\eqref{eq:d-beta-rule} to obtain $\hat d^{\rm deloc}_\beta$
(see Appendix~\ref{sec:finite-N}). The slope drifts under
noise, so this is a low-noise cross-check, not a primary
handle.
\item \textit{Noise model:} identify the ambient noise
class from the angular-momentum component the noise populates (the
$\ell=2$ component that FSM injects and the isotropic RSM
leaves empty, Sec.~\ref{sec:exp-blind}) and from the
bulk/localized-branch shape of the residual
$R = M^{(D)} - \hat M^{(d)}$ against the semicircle reference
\eqref{eq:wigner-prediction} (see Appendix~\ref{sec:exp-rmt}).
\item \textit{Report:} $\hat d = \hat d_{h_1}$, upgraded to
high confidence when the shrinkage (\textit{3}) and slope
(\textit{4}) cross-checks agree, together with the
noise-model verdict.
\end{enumerate}
\end{mdframed}

\smallskip
\noindent The multiplicity step (\textit{2}) alone fixes
$\hat d$ in every realisation of our inference experiment for
$\varepsilon \leq 0.2$ on $S^1$ and $S^2$; the shrinkage and
slope cross-checks (\textit{3}, \textit{4}) upgrade it to a
triple-agreement high-confidence estimate when they agree,
the small-$|\Lambda|$ branch being more delicate than the
large-$|\Lambda|$ one because of multiplicative finite-$N$
noise on the localised states (see Sec.~\ref{sec:discussion}). The
noise-model step (\textit{5}) then shifts the question from
``which manifold is the data on?'' to ``which ambient noise
model generated the deviation from it?''.

\section{Numerical experiments: latent sub-manifold inference}
\label{sec:experiments}

Before the inference experiments we illustrate how ambient
noise pollutes the BBS spectrum at the matrix level. We
compare the empirical spectral densities (ESDs) of the
latent $M^{(d)}$, the residual $M^{(D-d)}$, and the RSM
ambient $M^{(D)}$ \eqref{eq:aag-MD-geodesic} at
$\varepsilon\in\{0.10,0.22,0.32,0.45\}$, with $N=1000$,
latent $S^{2}$ and residual $S^{124}$. Each sub-sphere is
sampled either uniformly or from a concentrated
beta-product single-particle density: each hyperspherical
angle is drawn from a symmetric $\mathrm{Beta}(\kappa,\kappa)$
peaked at its midpoint, so increasing $\kappa$ shrinks the
particle cloud to a blob around a fixed point, an
ergodicity-breaking configuration in which the dynamics has
not explored the whole sphere. The sharpness $\kappa$ is
calibrated by simulation so the blob occupies a target
fraction of the sphere, here $40\%$, $35\%$, $30\%$, $25\%$,
and $20\%$ (Appendix~\ref{sec:appendix-beta-sphere} sets
out the beta-product model). Eigenvalues are pooled over $20$
realisations, Perron excluded. The grid calibrates to the
relative cosine-Frobenius perturbation
\begin{equation}
\eta_{\cos}(\varepsilon) :=
\frac{\|\cos M^{(D)} - \cos M^{(d)}\|_F}
     {\|\cos M^{(d)}\|_F}
= \varepsilon^{2}\,
\frac{\|G - \cos M^{(d)}\|_F}{\|\cos M^{(d)}\|_F}
\approx \varepsilon^{2},
\label{eq:eta-cos-aag}
\end{equation}
so $\eta_{\cos}\in\{1\%,5\%,10\%,20\%\}$, the perturbative
window in which the RSM mean-shift expansion of
Sec.~\ref{sec:skm} is controlled.

\begin{figure}[!htbp]
\centering
\includegraphics[width=0.82\textwidth]{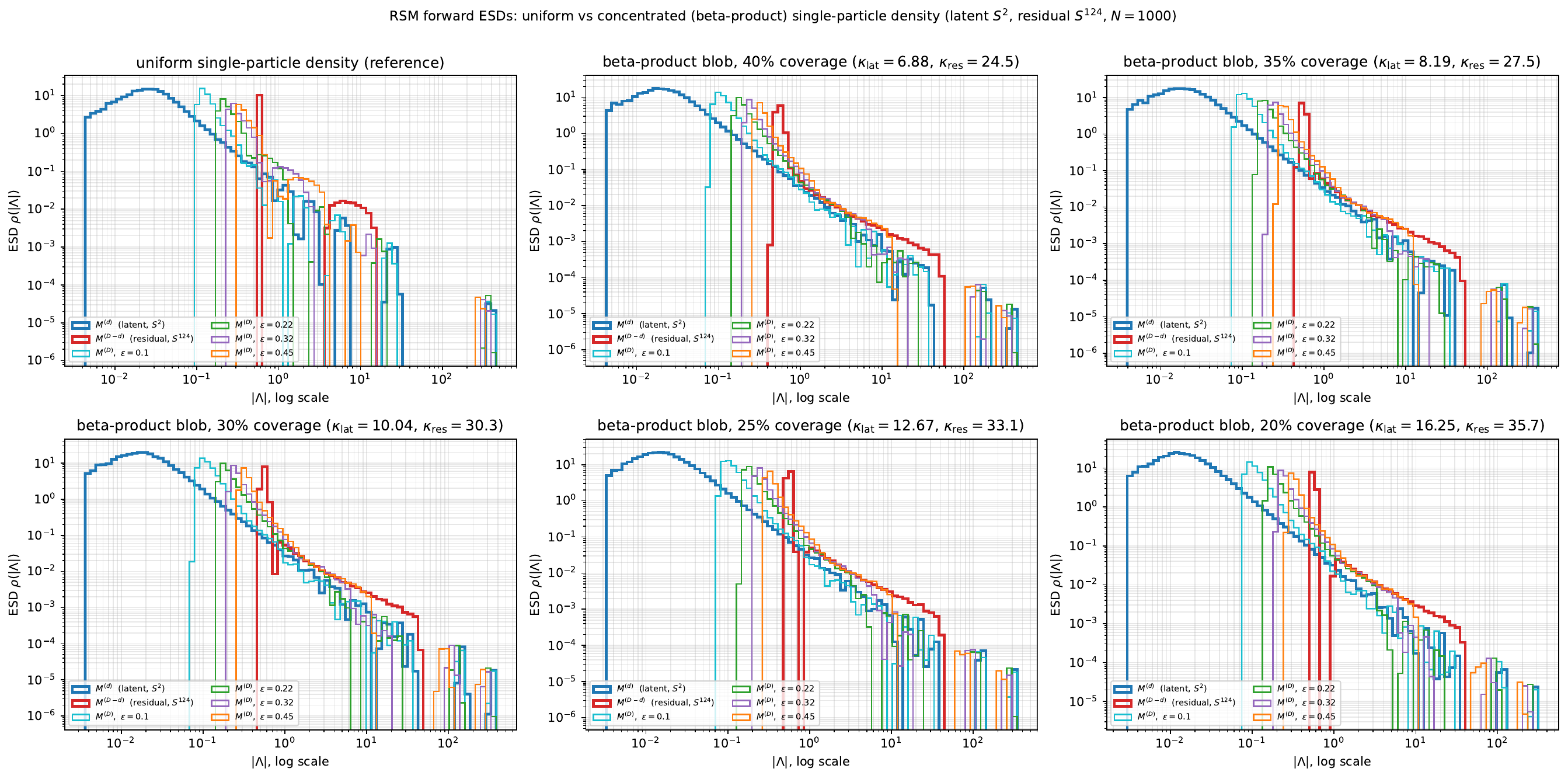}
\caption{ESDs of the latent $M^{(d)}$ (blue), residual
$M^{(D-d)}$ (red), and RSM ambient $M^{(D)}$ at
$\varepsilon\in\{0.10, 0.22, 0.32, 0.45\}$
($\eta_{\cos}\approx\{1,5,10,20\}\%$, Eq.~\eqref{eq:eta-cos-aag})
for a uniform reference (top-left) and beta-product
concentrated single-particle densities at $40\%$, $35\%$,
$30\%$, $25\%$, and $20\%$ sphere coverage (sharper beta
peaks, deeper
ergodicity breaking; the per-sphere sharpness
$\kappa_{\rm lat},\kappa_{\rm res}$ is shown in each panel).
$N = 1000$, $d = 3$, $D-d = 125$, 20 realisations pooled,
Perron excluded. As $\varepsilon$ grows the ambient ESD
interpolates from the latent BBS power law to the residual
Bordenave--MP bulk; concentrating the density reshapes the
endpoints but preserves the interpolation.}
\label{fig:aag-esd-noise-pollution}
\end{figure}

Figure~\ref{fig:aag-esd-noise-pollution} shows the result.
The latent ESD on $S^{2}$ is the BBS power law with the
$SO(3)$ multiplet spikes; the residual ESD on $S^{124}$ is
the Bordenave--MP atom-plus-bulk. As $\varepsilon$ grows the
ambient ESD interpolates between them: indistinguishable
from the latent at $\eta_{\cos}\approx 1\%$, acquiring a
Marchenko--Pastur bulk by $5$--$10\%$, and dominated by the
residual bulk by $20\%$. Concentrating the single-particle
density (the blob panels) reshapes the latent and residual
endpoints but leaves this interpolation pattern intact; the
concentration is a stress test of the uniform-sampling
assumption, treated as a misspecification in
Sec.~\ref{sec:appendix-nonuniform}. This is the free-convolution
picture made concrete,
and it frames the inference task: read the latent BBS
fingerprint off $M^{(D)}$ while the noise has not yet
washed it out.

The experiments that follow isolate the three questions that
decide whether the latent manifold can be recovered from an
observed ambient matrix. \emph{First}, how stable is the integer
multiplicity of the lowest non-Perron BBS multiplet as the
ambient noise grows, and does that stability depend on the
generative model? \emph{Second}, how do the eigenvalues that carry
that multiplet move under the same perturbation, and do they
follow the angular-momentum-level shrinkage law of
Sec.~\ref{sec:attenuation-law}? \emph{Third}, working blind from a
single observed $M^{(D)}$ and without the latent coordinates,
can we name both the manifold and the ambient noise model that
produced it?

All three experiments share one setup. We sample $N = 1000$
points uniformly on the latent spheres $S^1, S^2, S^3$ (embedding
dimension $d \in \{2, 3, 4\}$, intrinsic dimension $d-1$) and embed
them in $\R^{D}$ with $D = 128$. Two noise models drive the
ambient matrix at the cosine-kernel level: the isotropic RSM
forward \eqref{eq:aag-MD-geodesic}, with a uniform residual on
$S^{D-d-1}$, and the FSM forward of Sec.~\ref{sec:skm} with a
nonnegative array carrying weight in the $\beta_{20}$ sector,
whose kernel adds the $\ell = 2$ zonal component
$\mathcal Z_2^{(d)}(\tilde x_i\!\cdot\!\tilde x_j)$. To compare the two at equal
perturbation magnitude we drive each at a common relative
cosine-Frobenius level
$\eta = \|\cos M^{(D)} - \cos M^{(d)}\|_F /
\|\cos M^{(d)}\|_F$; for RSM this is $\eta \approx \varepsilon^2$,
while the FSM amplitude is calibrated to the same $\eta$ by
bisection. We sweep
$\eta \in \{0, 0.05, 0.10, 0.20, 0.35, 0.50, 0.65, 0.80\}$ with
20 independent realisations per cell. The upper end of this
range is well outside the perturbative window of
Sec.~\ref{sec:skm} and serves as a stress test of the discrete
diagnostics.

\subsection{Multiplicity stability of the lowest multiplet}
\label{sec:exp-multiplets}

The lowest non-Perron multiplet on $S^{d-1}$ carries the fixed
integer multiplicity $h(1, d) = d$: a doublet on $S^1$ ($d=2$), a
triplet on $S^2$ ($d=3$), a quartet on $S^3$ ($d=4$). In the rank-ordered
spectrum it appears as a plateau in $|\Lambda_K|$ bounded by two
large gaps, the Perron-to-multiplet gap at $K = 1 \to 2$ and the
multiplet-to-next gap at $K = 1 + h(1,d) \to 2 + h(1,d)$. A
detector that walks the descending $\log|\Lambda_K|$ from
$K = 2$ and stops at the first drop exceeding a fixed threshold
$\tau$ returns a multiplet size $\hat h_1 = K^\star - 1$, hence the
embedding dimension $\hat d = \hat h_1$ and intrinsic dimension
$\hat h_1 - 1$. We fix $\tau = 0.30$ throughout. The gap detects an
isolated empirical cluster; identifying it with the latent $(1,0)$
sector rests on the product-kernel separation of
Sec.~\ref{sec:product-spectrum}, not on the gap alone.
A base-ten $\log$ gap of
$\tau$ between two consecutive levels means a multiplicative ratio
$|\Lambda_{K^\star-1}|/|\Lambda_{K^\star}|\ge10^{\tau}$, hence an
additive gap $g\ge|\Lambda_{K^\star}|(10^{\tau}-1)$. At
$\tau=0.30$ the ratio is $10^{0.30}\approx2$, so the detected
boundary certifies, from the observed spectrum alone, an additive
gap $g\gtrsim|\Lambda_{K^\star}|$. The gap is observable, but
protection also needs $\|\delta M\|_{\rm op}<g/2$, which the
threshold does not by itself certify; we use $10^{\tau}\approx2$ as
the empirical proxy for the Davis--Kahan requirement of
Sec.~\ref{sec:multiplet-table}, matched to where that bound holds in
the experiments. A larger $\tau$ trades recovery reach for a
stronger separation guarantee.

\begin{figure}[!htbp]
\centering
\includegraphics[width=0.78\textwidth]{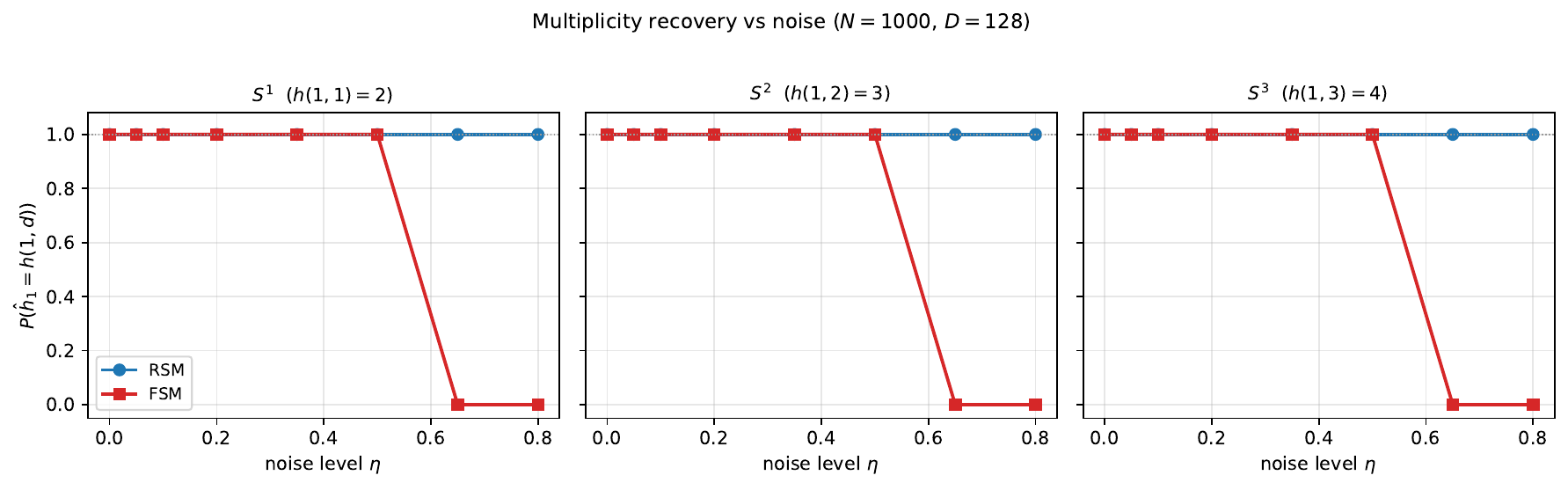}
\caption{Multiplicity recovery $P(\hat h_1 = h(1,d))$ vs
noise $\eta$ for RSM and FSM on $S^1,S^2,S^3$ ($N = 1000$,
$D = 128$, 20 realisations, detector threshold
$\tau = 0.30$). RSM recovers it up to $\eta = 0.80$; FSM up
to $\eta = 0.50$, then fails when its inter-multiplet gap
closes (Sec.~\ref{sec:exp-eigenvalues}).}
\label{fig:q1}
\end{figure}

Figure~\ref{fig:q1} reports the recovery probability
$P(\hat h_1 = h(1,d))$ over the 20 realisations. For RSM the
multiplicity is recovered in every realisation at every noise
level tested, up to $\eta = 0.80$ on all three manifolds. For
FSM it is recovered perfectly up to $\eta = 0.50$ and then
collapses to zero at $\eta \geq 0.65$, simultaneously on $S^1$,
$S^2$ and $S^3$. These rates are confirmed at higher statistics,
with pre-specified threshold and per-realisation confidence
intervals, against null and out-of-family models in
Sec.~\ref{sec:exp-validation}. The integer is therefore gap-protected across
the whole perturbative range and well beyond it. A continuous
deformation of the ambient noise cannot change a discrete
multiplicity until it closes the gap that isolates the
multiplet, and the eigenvalue analysis of
Sec.~\ref{sec:exp-eigenvalues} shows that closing the gap is
exactly what happens to FSM near $\eta \approx 0.6$. The
recovery and eigenvalue numbers for $S^2$ are collected in
Table~\ref{tab:q2}.

\subsection{Eigenvalue stability and the shrinkage law}
\label{sec:exp-eigenvalues}

The multiplicity is an integer; the eigenvalues that carry it
are not. We track the mean magnitude of the $\ell = 1$ multiplet
block ($K = 2, \dots, 1 + h(1,d)$) and normalise it by its
noiseless latent value to obtain the shrinkage factor
$|\Lambda^{(D)}_{\ell=1}| / |\Lambda^{(d)}_{\ell=1}|$, and we
record the $\log$ gap from the $\ell = 1$ block to the next
multiplet.

\begin{figure}[!htbp]
\centering
\includegraphics[width=0.80\textwidth]{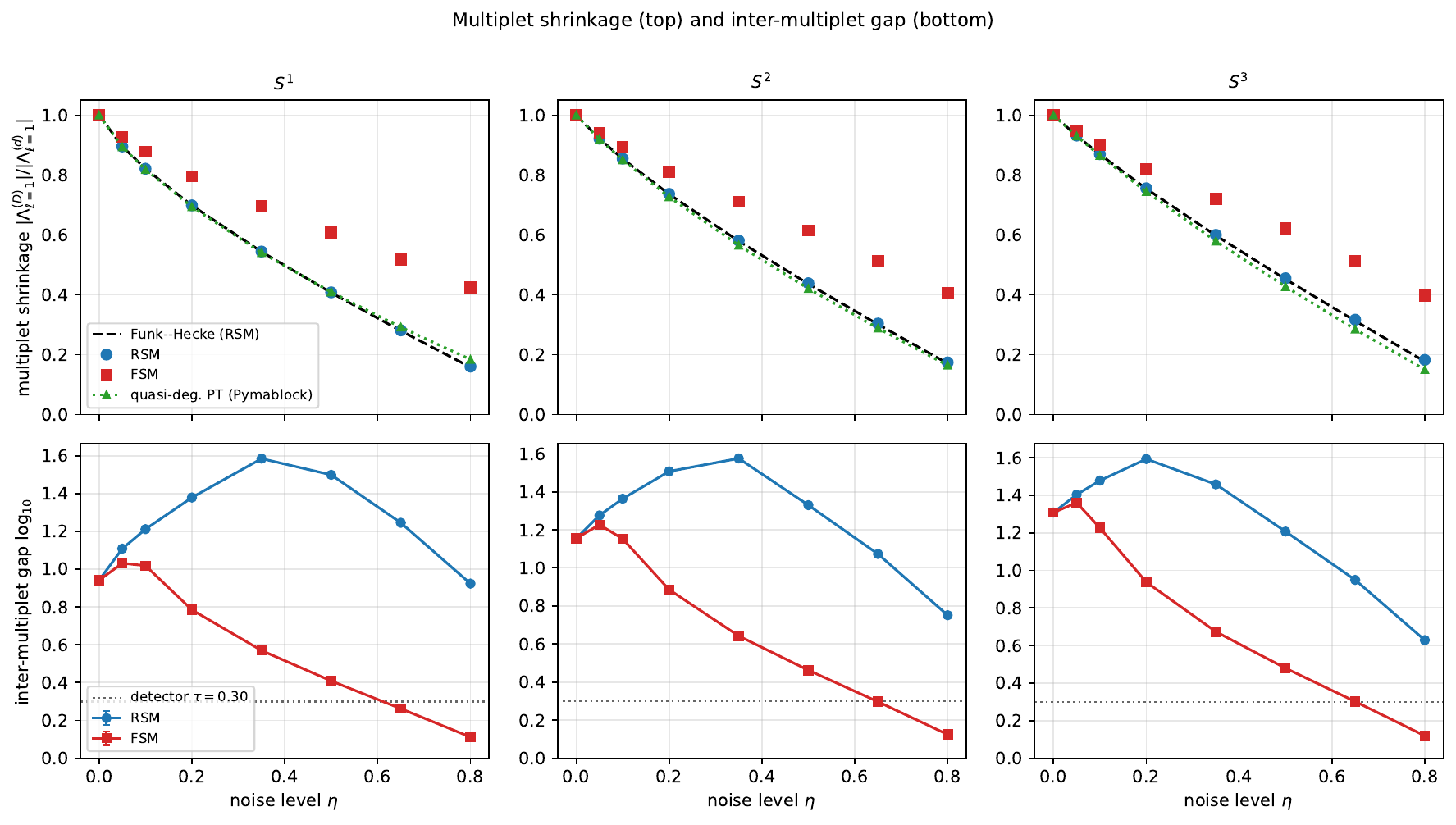}
\caption{Top: $\ell = 1$ multiplet shrinkage
$|\Lambda^{(D)}_{\ell=1}| / |\Lambda^{(d)}_{\ell=1}|$ vs
$\eta$ for RSM and FSM on $S^1,S^2,S^3$, with the Funk--Hecke
prediction $f_1$ (dashed); RSM lies on it, FSM shrinks more
slowly. The green curve is the quasi-degenerate block
reduction (Pymablock \cite{ArayaDay2025},
Appendix~\ref{sec:appendix-qdpt}) of the bounded spliced mean
drift; it tracks RSM and $f_1$ across the full range on all
three manifolds. Bottom: the
$\log$ inter-multiplet gap with the
detector threshold $\tau = 0.30$ (dotted); the RSM gap stays
wide while the FSM gap closes near $\eta \approx 0.6$, where
its recovery fails.}
\label{fig:q2}
\end{figure}

For RSM the measured shrinkage matches the parameter-free
Funk--Hecke prediction $f_1(\varepsilon)$ of
Sec.~\ref{sec:attenuation-law} to within the realisation
scatter at every noise level (Table~\ref{tab:q2},
Fig.~\ref{fig:q2} top row). The same shrinkage emerges from
the quasi-degenerate block reduction (Pymablock
\cite{ArayaDay2025}, Fig.~\ref{fig:q2} top row, green): on
$S^1, S^2, S^3$ it reproduces $f_1$ across the full range,
a numerical cross-check consistent with the recovery being the
resummed degenerate perturbation theory of
Sec.~\ref{sec:attenuation-law}. That
route needs the bounded spliced drift~\eqref{eq:aag-mu-spliced}
to avoid the fictitious $\arccos$ edge singularities, whereas
the Funk--Hecke law needs no such expansion, which is why the
pipeline reads the positions from $f_1$
(see Appendix~\ref{sec:appendix-qdpt}). FSM shrinks more slowly, because
its injected $\ell = 2$ component partly refills the kernel,
so its multiplet positions follow the FSM-specific law
rather than the RSM one. The eigenvalues thus move smoothly
and in a model-specific way, which is what the slope
diagnostics see as a drifting power law.

The two models part company in the inter-multiplet gap
(Fig.~\ref{fig:q2} bottom row). For RSM the gap widens at first,
because the higher-$\ell$ neighbours shrink faster than the
$\ell = 1$ block, and it stays above $0.6$ in $\log$ even at
$\eta = 0.80$, so the multiplet remains isolated. For FSM the
injected $\ell = 2$ multiplet sits directly above the $\ell = 1$
block and rises with $\eta$, so the gap falls monotonically and
crosses the detector threshold $\tau = 0.30$ near
$\eta \approx 0.6$. This is the mechanism behind the multiplicity
breakdown of Sec.~\ref{sec:exp-multiplets}: FSM recovery fails at
exactly the noise level where its gap closes. The continuous
eigenvalues degrade gracefully, but the integer they support is
stable until the gap that protects it is gone.

\begin{table}[!htbp]
\centering
\begin{tabular}{l|ccccccc}
$\eta$ & $0.05$ & $0.10$ & $0.20$ & $0.35$ & $0.50$ & $0.65$ & $0.80$ \\
\hline
RSM recovery & $1.00$ & $1.00$ & $1.00$ & $1.00$ & $1.00$ & $1.00$ & $1.00$ \\
RSM $|\Lambda^{(D)}_{\ell=1}|/|\Lambda^{(d)}_{\ell=1}|$
 & $0.92$ & $0.85$ & $0.74$ & $0.58$ & $0.44$ & $0.30$ & $0.17$ \\
RSM Funk--Hecke $f_1$
 & $0.92$ & $0.85$ & $0.74$ & $0.58$ & $0.44$ & $0.30$ & $0.17$ \\
RSM gap $\log$ & $1.28$ & $1.36$ & $1.51$ & $1.58$ & $1.33$ & $1.07$ & $0.75$ \\
\hline
FSM recovery & $1.00$ & $1.00$ & $1.00$ & $1.00$ & $1.00$ & $0.00$ & $0.00$ \\
FSM $|\Lambda^{(D)}_{\ell=1}|/|\Lambda^{(d)}_{\ell=1}|$
 & $0.94$ & $0.89$ & $0.81$ & $0.71$ & $0.61$ & $0.51$ & $0.41$ \\
FSM gap $\log$ & $1.23$ & $1.15$ & $0.89$ & $0.64$ & $0.46$ & $0.30$ & $0.12$ \\
\end{tabular}
\caption{Multiplicity recovery, $\ell = 1$ multiplet shrinkage,
and inter-multiplet gap on $S^2$ ($N = 1000$, $D = 128$, 20
realisations). RSM shrinkage tracks the parameter-free
Funk--Hecke prediction $f_1$ exactly; FSM shrinks more slowly
and its gap closes near $\eta \approx 0.6$, where its
multiplicity recovery fails while RSM stays intact.}
\label{tab:q2}
\end{table}

\subsection{Blind identification of the manifold and the noise model}
\label{sec:exp-blind}

The first two experiments use the latent coordinates as an
oracle. The inference question is whether the same conclusions
survive when only $M^{(D)}$ is observed. We proceed blind in two
stages, each from a single ambient matrix, repeating over 20
realisations only to estimate success rates.

Working from a single matrix is justified by
self-averaging: the spectral observables of a large
random matrix concentrate around their ensemble mean, with
relative fluctuations of order $N^{-1/2}$, so one
$N\times N$ observation already determines the eigenvalue
density and the functionals we read from it (the multiplet
multiplicity for the dimension, and the FSM-injected $\ell=2$
component for the noise model) without averaging over
realisations \cite{Bun2017}. The spectral functionals of a
single $M^{(D)}$ concentrate as $N$ grows, though its $N$
eigenvalues are not independent replicates; the realisations here
convert the per-matrix outcome into a success rate, quantified at
higher statistics with confidence intervals in
Sec.~\ref{sec:exp-validation}.

\begin{figure}[!htbp]
\centering
\includegraphics[width=0.82\textwidth]{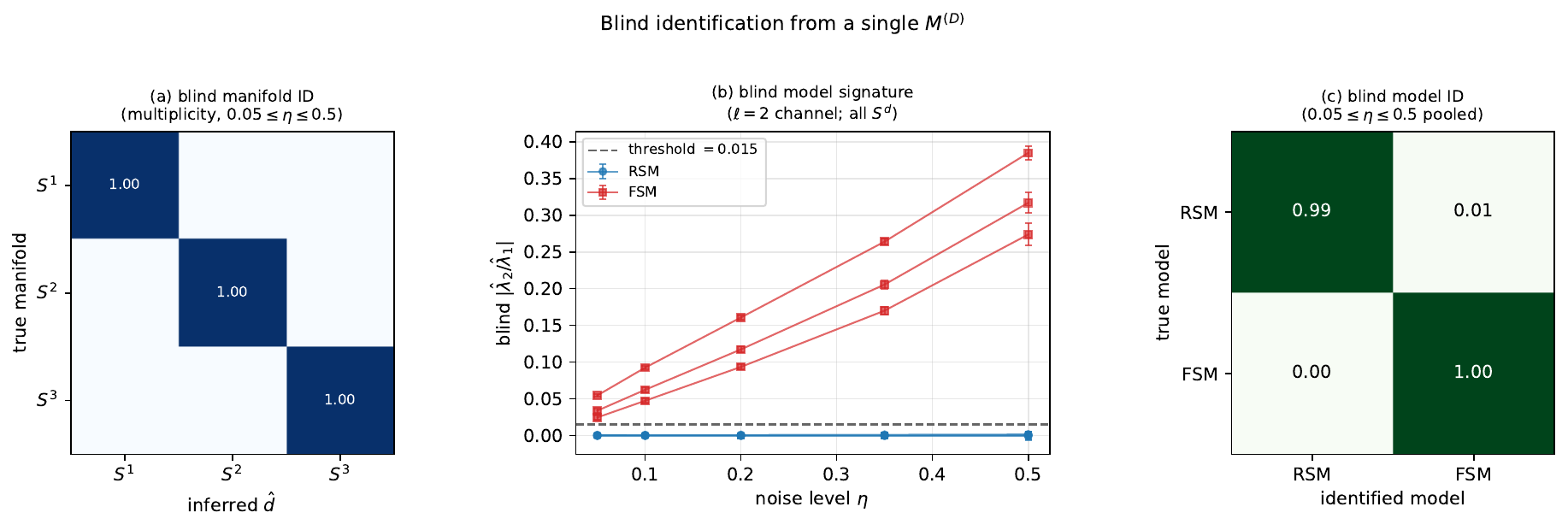}
\caption{Blind identification from a single $M^{(D)}$,
pooled over $0.05 \leq \eta \leq 0.50$ and $S^1$--$S^3$.
(a) Manifold confusion from the gap-protected multiplicity
(identity). (b) Blind $\ell = 2$ component ratio
$|\hat\lambda_2/\hat\lambda_1|$ from the top $\cos M^{(D)}$
eigenvectors: RSM at the parity floor, FSM rising with
$\eta$, separated by a geometric-mean boundary (dashed).
(c) The resulting noise-model confusion matrix.}
\label{fig:q3}
\end{figure}

The manifold is read from the gap-protected multiplicity, which
needs only the spectrum of $M^{(D)}$. Across both models and
$0.05 \leq \eta \leq 0.50$ the inferred $\hat d$ equals the true
dimension in every realisation on $S^1$, $S^2$ and $S^3$: the
manifold-confusion matrix is the identity (see Fig.~\ref{fig:q3}a).

The noise model is read from the $\ell = 2$ component, which is the
FSM signature. The geodesic kernel $\arccos(t)$ has only odd-$\ell$
content, since $\arccos(t) - \pi/2$ is odd, so RSM leaves the
$\ell = 2$ multiplet empty up to a parity-suppressed
$O(1/\sqrt N)$ Monte-Carlo floor, whereas FSM populates it
through its nonnegative $\beta_{20}$ coefficient. To measure it blind we estimate the
latent directions from the top $\hat d$ eigenvectors of
$\cos M^{(D)}$, which span the degree-one block: for centred
uniform spherical data the latent linear-harmonic Gram component
$\tilde x_i\!\cdot\!\tilde x_j$ has rank $h(1,d)=\hat d$ and
carries the leading non-Perron eigenvalues, the $\ell=0$ constant
mode being suppressed by the centring. We form the implied
unit-diagonal Gram matrix and read the component ratio
$|\hat\lambda_2 / \hat\lambda_1|$ from the zonal projector
estimator. The estimated basis fixes orientation only up to a
sign, so the discriminator uses the magnitude. The two classes
separate cleanly (see Fig.~\ref{fig:q3}b): RSM stays at the floor,
$|\hat\lambda_2/\hat\lambda_1| \leq 0.002$, at all $\eta$, while
FSM rises from $0.034$ at $\eta = 0.05$ to $0.32$ at
$\eta = 0.50$. A geometric-mean boundary at $0.015$ identifies
the model correctly in $99\%$ of RSM realisations and $100\%$ of
FSM realisations, pooled over $S^1$--$S^3$ and
$0.05 \leq \eta \leq 0.50$ (see Fig.~\ref{fig:q3}c). Model
identification is meaningful only where the manifold itself is
identified, so it is reported on the regime where the
multiplicity recovers. Beyond $\eta \approx 0.6$ the failed
dimension estimate makes the blind component ratio uninformative.

The residual-spectrum consistency check, which tests whether the
off-manifold deviation behaves as random-matrix noise once the
recovered geometry is subtracted, is reported in
Appendix~\ref{sec:exp-rmt}.

\subsection{Statistical validation: null models, out-of-family noise, and decision rates}
\label{sec:exp-validation}

The experiments above report recovery on in-family models. To
test the detector as a decision rule we fix its only free choice,
the log-gap threshold $\tau=0.30$, in advance, run it over $200$
independent realisations per cell, and report the per-realisation
decision rate with a Wilson $95\%$ confidence interval rather than
a pooled histogram. The setup is $N=600$ on a latent $S^2$
($d=3$), candidate set $d\in\{2,\dots,8\}$, ambient $D=128$. We
add two ingredients the in-family study lacks: an out-of-family
noise model, with the residual directions drawn from a
heavy-tailed Student-$t$ law ($\nu=3$) outside both the RSM and
FSM families, and three null models that carry no low-dimensional
spherical latent factor, where a faithful detector should not
report a small integer multiplet. The nulls are uniform points on an ambient
sphere $S^{20}$ (whose first multiplet $h_1=21$ lies outside the
candidate set), an anisotropic Gaussian with coordinate amplitudes
$\propto k^{-3/4}$ across the $D{=}128$ axes (a continuous
covariance spectrum with no integer plateau), and a random
symmetric distance matrix with off-diagonal entries
$\mathrm{Uniform}[0,1)$ (pure noise).

\begin{figure}[!htbp]
\centering
\includegraphics[width=0.95\textwidth]{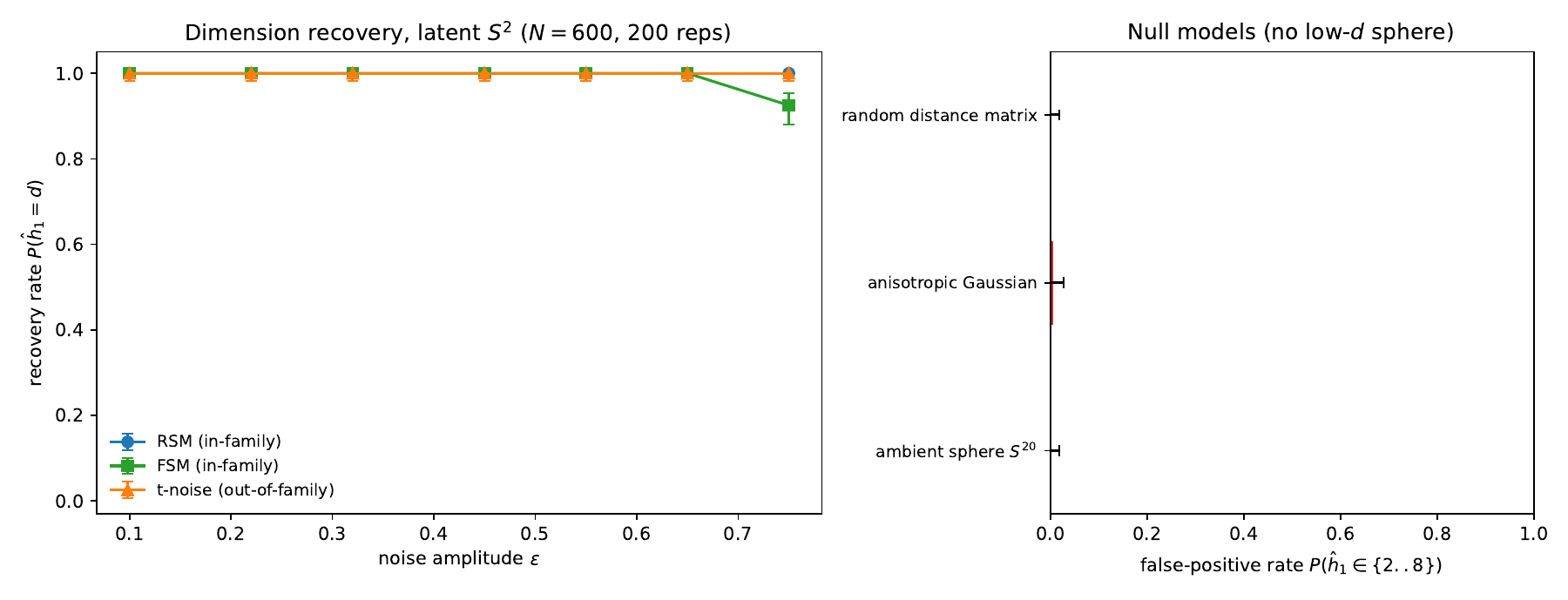}
\caption{Statistical validation of the multiplicity detector with
$\tau=0.30$ fixed in advance, $N=600$, $200$ realisations per
cell, error bars Wilson $95\%$ intervals. Left: dimension-recovery
rate $P(\hat h_1=d)$ versus noise for the in-family RSM and FSM
models and the out-of-family Student-$t$ residual model; RSM and
$t$-noise hold to $\varepsilon=0.75$ while FSM degrades near
$\varepsilon=0.75$. Right: false-positive rate
$P(\hat h_1\in\{2,\dots,8\})$ on three null models without a
low-dimensional spherical latent, all consistent with zero.}
\label{fig:validation}
\end{figure}

Table~\ref{tab:validation} and Fig.~\ref{fig:validation} report
the result. The in-family RSM recovers $h_1=3$ in every
realisation up to $\varepsilon=0.75$, the FSM up to
$\varepsilon=0.65$ and in $92.5\%$ of realisations at
$\varepsilon=0.75$ (the FSM losing its gap earlier, as in
Sec.~\ref{sec:exp-multiplets}), and the out-of-family $t$-noise
model is recovered throughout, since the detector reads only the
top multiplet and is insensitive to the residual law. The three
null models give a false-positive rate consistent with zero
($\leq 0.5\%$), so the detector does not hallucinate a small
sphere on data that lacks one. The classification of Sec.~\ref{sec:exp-blind}
is evaluated on held-out realisations distinct from those used to
place its boundary, separating threshold selection from
evaluation.

\begin{table}[!htbp]
\centering
\caption{Per-realisation decision rates with Wilson $95\%$
intervals, $N=600$, latent $S^2$, $200$ realisations per cell,
threshold $\tau=0.30$ fixed in advance. In-family and
out-of-family rows give the recovery rate $P(\hat h_1=3)$; null
rows give the false-positive rate $P(\hat h_1\in\{2,\dots,8\})$.}
\label{tab:validation}
\begin{tabular}{l ccc}
\hline
& $\varepsilon=0.45$ & $\varepsilon=0.65$ & $\varepsilon=0.75$ \\
\hline
RSM (in-family) & $1.00\,[.98,1]$ & $1.00\,[.98,1]$ & $1.00\,[.98,1]$ \\
FSM (in-family) & $1.00\,[.98,1]$ & $1.00\,[.98,1]$ & $0.93\,[.88,.95]$ \\
$t$-noise (out-of-family) & $1.00\,[.98,1]$ & $1.00\,[.98,1]$ & $1.00\,[.98,1]$ \\
\hline
& \multicolumn{3}{c}{false-positive rate} \\
\hline
null: ambient $S^{20}$ & \multicolumn{3}{c}{$0.00\,[.00,.02]$} \\
null: anisotropic Gaussian & \multicolumn{3}{c}{$0.005\,[.001,.03]$} \\
null: random matrix & \multicolumn{3}{c}{$0.00\,[.00,.02]$} \\
\hline
\end{tabular}
\end{table}

The full study underlying every figure fixes its fit windows and
normalisations in advance, collected in
Table~\ref{tab:reproducibility} for reproducibility.

\begin{table}[!htbp]
\centering
\caption{Reproducibility settings for the diagnostics. All
thresholds and windows are fixed before evaluation.}
\label{tab:reproducibility}
\begin{tabular}{l l}
\hline
quantity & setting \\
\hline
log-gap threshold $\tau$ & $0.30$ (base ten), fixed \\
candidate dimensions & $d\in\{2,\dots,8\}$ \\
corrected slope-fit window (per candidate $d$) & $K\in[d_{\rm guess},\sqrt N]$ \\
residual rank cutoff $K_{\rm lat}$ & $\lfloor\sqrt N\rfloor$ \\
eigenvalue normalisation & operator units $\Lambda/N$ \\
Perron mode & excluded ($\ell=0$) throughout \\
shrinkage quadrature & nested Gauss--Legendre, no $\varepsilon$ series \\
realisations per validation cell & $200$ (Wilson $95\%$ CI) \\
\hline
\end{tabular}
\end{table}

\section{Discussion}
\label{sec:discussion}

The central contribution of this paper is a coordinate-free
inference method, packaged as Algorithm~1, that reads the
latent geometry off the ambient distance matrix alone,
without constructing the ambient vectors. Because the ambient
spectrum is a finite-rank spiked deformation that reorganises
collectively, it is not recovered eigenvalue by
eigenvalue. Recovery rests instead on the gap-protected
integer multiplicity and the resummed angular-momentum
shrinkage, not on the
divergent single-level perturbation theory. The diagnostics
are complementary and degrade at different rates with ambient
noise, and we discuss below the empirical lessons and the
limitations of the framework.

\begin{figure}[!htbp]
\centering
\includegraphics[width=0.80\textwidth]{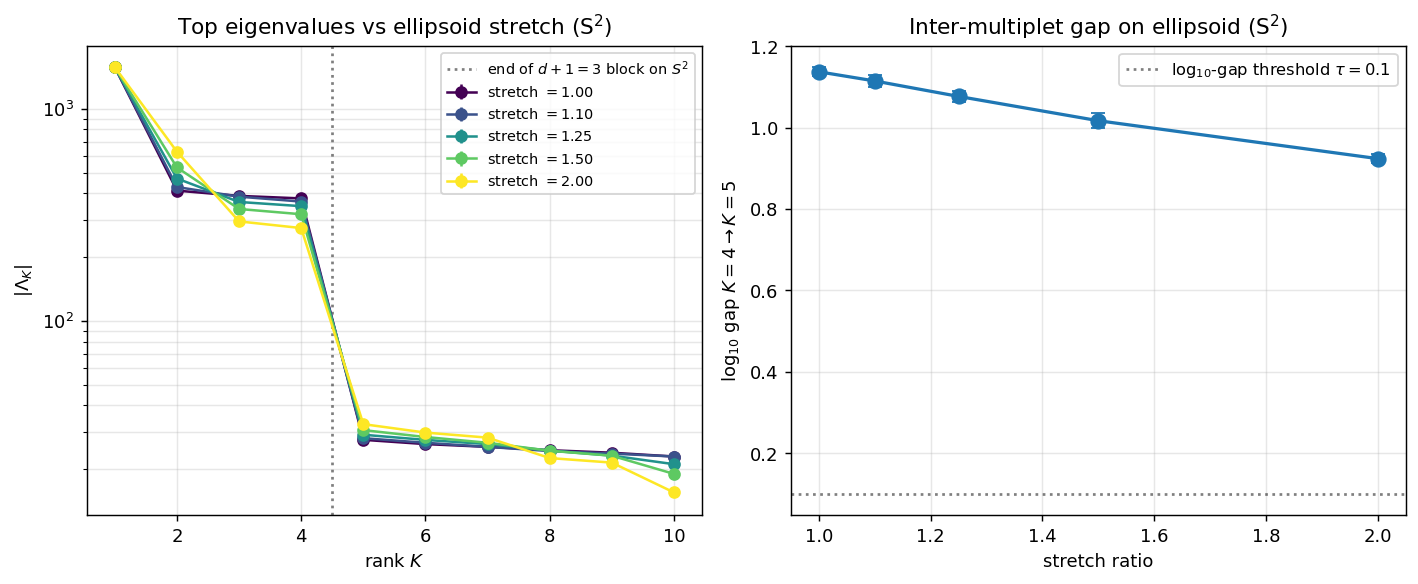}
\caption{Anisotropic-sampling negative control on $S^2$,
$N = 1000$. Latent samples are drawn uniformly on
$S^2$ then stretched along $z$ by factor $r$ before
forming the unit-direction distance matrix; taking unit
directions keeps the spherical metric and instead
concentrates the sampling into an axial density, so the split
below is the $\mathcal T_q$ symmetry breaking of
Appendix~\ref{sec:appendix-nonuniform}, not an intrinsic
ellipsoid geometry. Left: top-$10$
$|\Lambda_K|$ vs rank $K$ for each $r$. The lowest
non-Perron triplet at $K \in \{2, 3, 4\}$ (hyper-sphere
$d + 1 = 3$ irrep) splits into a singlet at $K = 2$ plus a
doublet at $K \in \{3, 4\}$ for $r \gtrsim 1.25$. Right:
$\log$ gap between $K = 4$ and $K = 5$ vs $r$. The
gap separating the (now-split) first block from the
$\ell = 2$ quintet remains open, but the internal split
within the original triplet drives the multiplet
diagnostic to misreport $\hat h_1 = 1$.}
\label{fig:ellipsoid}
\end{figure}

The clearest empirical result is that the latent
dimension is a discrete, gap-protected signature in the
spectrum of $M^{(D)}$ (see Sec.~\ref{sec:mult-invariant}): the
multiplicity $h(1, d) = d$ of the lowest non-Perron
BBS multiplet is recovered correctly in every realisation
up to a relative noise level $\eta = 0.5$, and up to
$\eta = 0.8$ for the isotropic noise model, while the
continuous slope diagnostics degrade smoothly with noise.
The stability mechanism is
analogous to topological invariants in gapped quantum
systems, where Chern numbers, representation multiplicities,
and ground-state degeneracies are stable under any continuous
perturbation that does not close the relevant gap. Here the relevant
gap is the inter-multiplet $\log$ gap, which on $S^2$
at $N = 1000$ is $\sim 1.2$ and stays open until the noise
finally closes it near $\eta \approx 0.6$, where the
multiplicity recovery fails. The analogy is mechanistic, not
topological: $h(1, d) = d$ is the dimension of the
first nontrivial $SO(d)$ irreducible representation of the
hyper-sphere $S^{d-1}$, and that degeneracy is protected only
while the sampling respects $SO(d)$. A nonuniform sampling
density $q$ breaks the symmetry and splits the multiplet into the
irreducible representations of the subgroup fixing $q$, an effect
already present in the limiting operator $\mathcal T_q$ and
quantified in Appendix~\ref{sec:appendix-nonuniform}. It is a
misspecification of the uniform-sampling assumption, not an
ambient-noise effect, and it scales as $O(N\alpha)$ in the density
anisotropy $\alpha$ rather than as an $O(\sqrt N)$ fluctuation.
Figure~\ref{fig:ellipsoid} reports a direct test on $S^2$.
Drawing points uniformly, stretching them along $z$ by a factor
$r$, and renormalising to unit directions does not change the
spherical metric; it reweights the sampling into an axial
density on $S^2$. The hyper-sphere triplet ($r = 1$, detected
sequence $[3, 5]$) then splits at $r \gtrsim 1.25$ into
singlet + doublet (detected $[1, 2, 5]$), the decomposition under
the residual $SO(2) \times \mathbb{Z}_2$ symmetry predicted by
$\mathcal T_q$. The single-multiplet hyper-sphere inversion fails
past the tolerance radius (see Appendix~\ref{sec:appendix-nonuniform}),
but the full multiplet sequence still identifies the latent
geometry class through the table of
Sec.~\ref{sec:multiplet-table}.

A useful split emerges from the inference experiment:
the \emph{latent geometry} (the dimension $d$ and the
multiplet sequence) is encoded at the \emph{top} of the
descending-$|\Lambda|$ rank-ordered spectrum (low $K$),
while the \emph{embedding model} (the ambient noise
model) is encoded in the angular-momentum components that the
geometry leaves empty. The two questions are answered by
different parts of the same data, and Algorithm~1
organises this split into a single workflow. The
dimension follows from the gap-protected multiplicity,
and the noise model follows from the $\ell = 2$ component:
the isotropic RSM kernel leaves it at the
parity-suppressed floor, while FSM populates it through
its $P_2$ term, a distinction to which the dimension
signal is blind (see Sec.~\ref{sec:exp-blind}). Both are read
blind from a single $M^{(D)}$, without oracle access to
$M^{(d)}$. An independent consistency check on the
off-manifold residual, confirming that the deviation from
the recovered geometry is random-matrix noise, is given in
Appendix~\ref{sec:exp-rmt}
(see Eq.~\eqref{eq:operational-residual},
Fig.~\ref{fig:residual-rmt-merged}).

\begin{figure}[!htbp]
\centering
\includegraphics[width=0.80\textwidth]{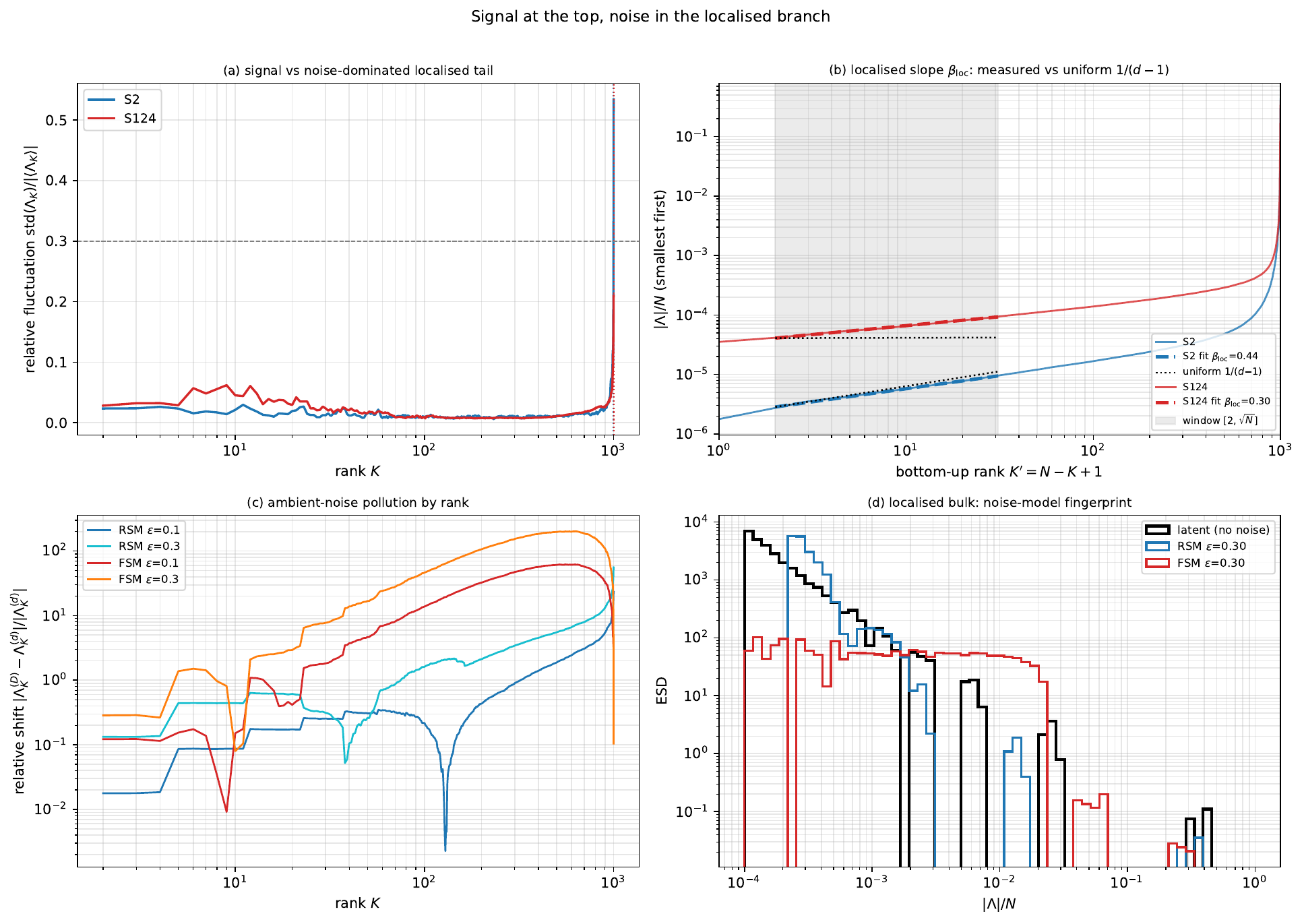}
\caption{Signal and noise across the spectrum ($N=1000$,
Perron excluded). Panels (a,b) use non-uniform
(Beta-concentrated) latent configurations at zero ambient
noise; (c,d) are on $S^2$ with the two ambient-noise models,
isotropic RSM and anisotropic ($\ell=2$-injecting) FSM. (a)
Relative eigenvalue fluctuation by rank: $\sim10^{-2}$ on the
delocalised branch (low $K$), rising to $\sim0.5$ on the
localised branch (high $K$). (b) Localised slope
$\beta_{\rm loc}$ (dashed) in bottom-up rank $K'=N-K+1$
versus the uniform-sphere $1/(d-1)$ (dotted) over the window
$K'\in[2,\sqrt N]$ (shaded); see text. (c) Per-rank relative
shift under ambient noise: $O(10^{-2})$ delocalised,
$O(1$--$10^2)$ localised. (d) Localised-bulk ESD: RSM decays
more steeply, FSM flattens toward a Wigner-like plateau.}
\label{fig:localized-branch}
\end{figure}

Our analysis suggests that the signal sits at the top of the spectrum while noise resides at the bottom.
The generalised sample-covariance structure
$M=\Phi A\Phi^\top$ (see Appendix~\ref{sec:appendix-beta-sphere})
organises this split through a Baik--Ben~Arous--P\'ech\'e
(BBP) transition of the finite-rank spiked deformation of the
residual Marchenko--Pastur law
\cite{BaikBenArousPeche2005, BenaychGeorgesNadakuditi2011}. The
control parameter is the magnitude of each (shrunk) multiplet
eigenvalue $N\,a_\ell[\kappa_{\rm obs}]$ measured against the
residual bulk edge, the deformation having fixed rank $h(\ell,d)$
per multiplet in the BBS regime. Each non-Perron multiplet is a separate spike, so
the transition governs the handful of largest such outliers,
not a single top eigenvalue, and the one positive Perron
eigenvalue (the trivial $\ell=0$ constant mode) is excluded
throughout. A multiplet above the bulk edge is supercritical
and appears as a stable, self-averaging outlier (relative
fluctuation $\sim N^{-1/2}$): this is the delocalised branch,
which carries the dimension through the multiplet
multiplicities. As the noise grows (or $\ell$ increases) the
shrinkage of Sec.~\ref{sec:attenuation-law} drives the
multiplet below the edge, where it turns subcritical and
merges into the bulk: this is the localised branch, the
small-$|\Lambda|$ region where the sampling fluctuation is of
order the eigenvalues themselves
(see Fig.~\ref{fig:localized-branch}a). The two ends of the
spectrum therefore play opposite roles, and the localised
branch is a poor geometry estimator but a sharp noise probe.
Its power-law slope already fails as a route to $d$ at zero
noise. Because $\beta_{\rm loc}=1/(d-1)$ is a uniform-sphere
result, a non-uniform latent distribution distorts it: the
Beta-concentrated configurations of panels (a,b) give
$\beta_{\rm loc}=0.44$ on $S^2$ (versus $0.5$) and $0.30$ on
the strongly concentrated $S^{124}$ (versus $0.008$), so the
distribution, not the nominal dimension, sets the slope
(see Fig.~\ref{fig:localized-branch}b). Ambient noise compounds
this, polluting the localised branch by two to three orders
of magnitude more than the delocalised branch in relative
shift (see Fig.~\ref{fig:localized-branch}c). The same
sensitivity makes the localised branch the sharpest
fingerprint of the noise mechanism: the isotropic RSM and
the anisotropic, $\ell=2$-injecting FSM fill the
small-$|\Lambda|$ region with visibly different shapes, FSM
flattening it toward a Wigner-like plateau
(see Fig.~\ref{fig:localized-branch}d). This is the residual
random-matrix discrimination of Appendix~\ref{sec:exp-rmt},
read directly off the localised branch.

We next turn to the finite-$N$-corrected exponent and its limitations.
The corrected-slope estimator
$\hat\beta^{\rm corr}_{\rm deloc} = \hat\beta_{\rm deloc} -
\Delta\beta(N, d_{\rm guess})$ subtracts the deterministic
finite-window offset of Appendix~\ref{sec:finite-N} to anchor
the rank-decay slope to the BBS asymptote at zero noise. That
offset is not a noise effect: the BBS power law is the
$\ell\to\infty$ asymptote, and the fit window $[2,\sqrt N]$
samples pre-asymptotic multiplets, so $\Delta\beta(N,d)$ is the
operator's finite-window curvature, fixed by the Funk--Hecke
coefficients (see Sec.~\ref{sec:finite-N-window}). It is
non-negligible ($\Delta\beta \approx 0.5$ at $N = 1000$ across
$d$, large enough to invalidate direct $d$-inversion from the
raw exponent) and decays with a shallow effective exponent
$\approx0.2$--$0.3$ for $d = 2, 3, 4$ (see Fig.~\ref{fig:beta-del-N}),
the shallow rate expected rather than anomalous. Under noise the
slope acquires a further, $\varepsilon$-dependent bias from the
angular-momentum-level shrinkage law
(see Sec.~\ref{sec:attenuation-law}), which erodes the higher-$\ell$
(smaller-$|\Lambda|$) components fastest, so the slope is a
derived, secondary handle.
The localised branch is treated at leading order
$\beta_{\rm loc} = 1/(d-1)$ only: the $v^2 \sim N$ obstruction
(see Sec.~\ref{sec:bbs-power-laws}) prevents a controlled
$1/\sqrt N$ expansion, and the empirical $\beta_{\rm loc}$
holds to $10\text{--}30\%$ finite-$N$ noise at $d \geq 2$.

\emph{Which BBS power law is the better dimension diagnostic at finite $N$?}
At $N = 1000$, $D = 128$, $\varepsilon = 0.05$ the semicircle-reference edge
$2\sqrt{N v}$ of the residual is $\sim 1\text{--}5$. The
latent landmark $|\Lambda^{(d)}_{1+h(1,d)}|$ is of
order $10^{2}\text{--}10^{3}$, parametrically outside the
noise. The small-$|\Lambda|$ landmark
$|\Lambda^{(d)}_{K=100}| \sim 10^{-1}\text{--}10^{0}$
sits \emph{inside} the noise. The delocalised power law is
therefore shielded from the noise floor and the localised
one is not: the small-$|\Lambda|$ slope estimate degrades
with $\varepsilon$ at the rate the noise grows. \textbf{The
large-$|\Lambda|$ delocalised power law is the more
reliable continuous diagnostic.} The small-$|\Lambda|$
exponent is a weak low-$\varepsilon$ consistency check at
best. The integer multiplet diagnostic, sitting yet
further above the noise floor at the very top of the
spectrum, is the most stable of the three.

The entry-wise perturbation theory is kept
only as the mean-drift correction for model identification. For isotropic RSM the
small parameter is $\delta_{\rm iso} = \varepsilon\sqrt D$
(see Sec.~\ref{sec:aaag}), so most of the isotropic RSM runs in
Sec.~\ref{sec:experiments} are stress tests beyond strict
PT validity rather than perturbative checks. The
finite-$N$ derivation in Appendix~\ref{sec:finite-N}
additionally uses a local plane-wave ansatz on the
Euclidean tangent space, ignoring intrinsic curvature; for
flat latents (e.g.\ $T^2$ Fourier representations) the
curvature corrections vanish identically and the CLT
correction derived here is the dominant subleading effect.

\subsection{Extending the framework beyond hyper-spheres}
\label{sec:multiplet-table}

The empirical experiments above are confined to hyper-sphere
latents, but the I-BBS framework is independent of that
choice. We collect here the general statements that allow
the diagnostic structure of Algorithm~1 to be applied to a
broader class of latents without further simulation.

The multiplet structure follows from the latent isometry group,
rigorously for the two-point homogeneous spaces. For a compact
rank-one symmetric space $\mathcal M_d$ (the hyper-spheres and the
real, complex, and quaternionic projective spaces, together with
the Cayley plane) the geodesic distance depends on a single
invariant, the kernel is zonal, and the Funk--Hecke and addition
theorems diagonalise $M^{(d)}$ exactly into multiplets whose
multiplicities are the dimensions of the harmonic blocks
\cite{Helgason1984, Azevedo2017}. Table~\ref{tab:multiplet-table}
records these entries. For a general compact homogeneous space
$G/H$ the situation is more delicate: $L^2(G/H)$ decomposes with
multiplicities governed by the $H$-fixed vectors in each
irreducible representation of $G$, a $G$-invariant kernel acts by
scalars on the resulting isotypic blocks, and the Laplace ordering
of those blocks need not coincide with the spectral ordering of
the distance matrix. The first-multiplet entries beyond the
rank-one symmetric spaces are therefore indicative candidates to
be confirmed case by case, not consequences of a single universal
rule.
\begin{table}[!htbp]
\centering
\renewcommand{\arraystretch}{1.4}
\begin{tabular}{|p{3.4cm}|p{3.6cm}|p{3.2cm}|p{3.2cm}|}
\hline
latent $\mathcal M_d$ & isometry group $G$ &
first-multiplet multiplicity $h_1$ & inversion
$d \leftarrow h_1$ \\
\hline
hyper-sphere $S^{d}$ & $SO(d+1)$ & $d + 1$ &
$d = h_1 - 1$ \\
\hline
flat torus $T^d = (S^1)^d$ &
$T^d \rtimes \mathrm{Aut}(\Lambda)$ &
$2d$ (equal-side lattice; generally $2$ per
shortest-vector pair) & $d = h_1/2$ (equal-side) \\
\hline
real projective $\mathbb{RP}^d$ &
$PO(d+1)$ &
$d(d+3)/2$ ($\ell = 2$ irrep, parity selects out
$\ell = 1$) & nonlinear in $h_1$ \\
\hline
\end{tabular}
\renewcommand{\arraystretch}{1.0}
\caption{Multiplet-multiplicity table for representative
latent geometries. The hyper-sphere and projective-space rows are
the rigorous rank-one symmetric-space entries; the flat-torus row
is lattice-dependent, with $h_1=2d$ only for the equal-side
lattice and generally $2$ per shortest-vector pair, and is
included as an indicative candidate. The inversion from $h_1$ to
$d$ is unique within each row but not across rows: detection of
$h_1$ produces a short list of compatible candidates whose
remaining ambiguity is resolved by the rank-decay slope
and the residual-RMT fingerprint of Algorithm~1. In this
table $d$ is the intrinsic manifold dimension; the
hyper-sphere $S^{d}$ is written $S^{d-1}$ in the
embedding-$d$ convention of the rest of the paper, where its
lowest multiplet has multiplicity $h_1=d$.}
\label{tab:multiplet-table}
\end{table}

The multiplicity is gap-protected by a Davis--Kahan bound.
Let $M^{0}$ be the unperturbed BBS matrix on the latent
$\mathcal M_d$, with an isolated cluster of eigenvalues
$\{\Lambda^{(0)}_{K_1}, \ldots, \Lambda^{(0)}_{K_{h_1}}\}$
separated from the rest of the spectrum by a gap $g > 0$
(the inter-multiplet gap). Suppose the ambient matrix is
$M^{(D)} = M^{0} + \delta M$ with $\|\delta M\|_{\rm op}
< g/2$. By Weyl's eigenvalue inequality (the addition problem of
\cite{PasturVasilchuk2000}) each eigenvalue moves by at most
$\|\delta M\|_{\rm op}<g/2$, so the eigenvalues in any
$g/2$-neighbourhood of the multiplet remain an isolated
cluster of exactly $h_1$ elements; the Davis--Kahan
$\sin\theta$ theorem \cite{DavisKahan1970} further bounds the
rotation of the corresponding $h_1$-dimensional eigenspace by
$\|\delta M\|_{\rm op}/g$, so the multiplet stays a coherent
block. The multiplicity is preserved; the same operator-norm
and gap-controlled Davis--Kahan estimate underlies the
closely related signal-plus-noise analysis of
\cite{ChenMa2025}.
The experiments calibrate the noise by the relative
cosine-Frobenius level $\eta$ (see Sec.~\ref{sec:experiments}),
but the norm entering the bound is the operator norm, which we
compute directly from $\delta M = M^{(D)}-M^{(d)}$: on $S^2$ it
is a fixed fraction of the Frobenius perturbation,
$\|\delta M\|_{\rm op}\approx0.55$--$0.6\,\|\delta M\|_F$, and
with the latent gap $g$ read from $M^{0}$ the condition
$\|\delta M\|_{\rm op}<g/2$ holds throughout the recovery
regime (to $\eta\approx0.4$ at $N=1000$). The multiplet is
observed to survive beyond this point as well, the
Davis--Kahan criterion being sufficient rather than tight.
This converts the empirical observation that
``the multiplet survives under ambient noise'' on
hyper-spheres into a quantitative bound on any latent
$\mathcal M_d$ with a resolved inter-multiplet gap. In the
hyper-sphere case the first inter-multiplet gap is
$g = N\,(|a_1|f_1-|a_3|f_3)+o(N)$, of order $N$, since the
lowest multiplets sit at the fixed continuum eigenvalues
$N a_\ell$ with $O(1)$ spacing
(see Appendix~\ref{sec:appendix-qdpt}); the $N^{1-2/d}$ scaling of
the BBS finite-$N$ analysis (see Appendix~\ref{sec:finite-N})
describes the crowded level spacing high in the tower, not the
gap that protects the first multiplet. For other latents the
relevant $g$ is computed from the spectral gap of $M^{0}$ in the
BBS limit.

The noise models and the residual fingerprint are not specific to spheres.
The two generative model classes (RSM and FSM) and the
residual-RMT diagnostic make no use of the latent being a
sphere: RSM is a convex combination of two sub-sphere Gram kernels
at the cosine level, whose isotropic limit resembles an ambient
additive perturbation without our claiming an exact
ambient-Gaussian or heat-kernel-Brownian equivalence, and FSM
specifies a positive-definite Gram-kernel construction in terms of
the latent $M^{(d)}$. Each class's first-order kernel mean and
variance is computed from $M^{(d)}$ and the ambient
geometry. Substituting a non-spherical $M^{(d)}$ leaves
the perturbative structure of Sec.~\ref{sec:layer2-pert}
intact, with the only manifold-specific input being the
unperturbed BBS spectrum on $\mathcal M_d$. The FSM
construction extends naturally to other kernels (heat
kernel on the sphere, hyperbolic-space kernels,
flat-torus kernels, etc.) by replacing the hyper-sphere
Gegenbauer basis with the appropriate spectral basis on
the latent manifold, with the BBS spectral signatures
replaced by the corresponding multiplet structure of the
latent isometry group. For the rank-one compact symmetric
spaces (the two-point homogeneous spaces: the hyper-spheres
together with the real, complex, and quaternionic projective
spaces of Table~\ref{tab:multiplet-table} and the Cayley
plane) the Funk--Hecke theorem, the Mercer expansion, and the
addition theorem carry over verbatim with the Gegenbauer
polynomials replaced by Jacobi polynomials
$R^{\alpha,\beta}_\ell$, whose parameters encode the space and
whose eigenvalue and multiplicity growth $\delta(\ell,d)\sim
\ell^{d-1}$ reproduce the same multiplet counting
\cite{Azevedo2017}.

We close by noting the empirical scope of this paper.
Quantitative validation in Sec.~\ref{sec:experiments} is
restricted to hyper-sphere latents $S^1$, $S^2$ and $S^3$.
Departures from uniform sampling are treated as a
misspecification, with the multiplet splitting and a tolerance
radius derived in Appendix~\ref{sec:appendix-nonuniform} and
illustrated by the ellipsoid experiment of
Sec.~\ref{sec:discussion} (see Fig.~\ref{fig:ellipsoid}).
Direct empirical tests of the framework on non-spherical
latents, on real neural-network representations, and on the
oracle-free residual reconstruction protocol of
Appendix~\ref{sec:exp-rmt} appear in the two companion
papers~\cite{halperin2026OMD, halperin2026grokking}.

\section{Summary}
\label{sec:summary}
\label{sec:summary-findings}

I-BBS infers a latent sub-manifold from the ambient
distance matrix alone, without constructing the ambient
vectors. Being coordinate-free, it applies even where a
vector space is undefined or only partly observable. It analyses the full observable distance matrix $M^{(D)}$
in a blind setting, using different parts of its spectrum to
learn the latent low-dimensional manifold from the
largest-$|\Lambda|$ eigenvalues and the noise model from the
lowest-$|\Lambda|$ ones; the two readouts together form the
basis of Algorithm~1.

The latent geometry sits at the top of the rank-ordered
spectrum. The multiplicity of the lowest non-Perron
multiplet is the integer-valued, gap-protected signature of
the dimension; for hyper-spheres $h(1,d)=d$, and the
inversion for tori, projective spaces, and products is
tabulated in Sec.~\ref{sec:multiplet-table}. It is recovered
correctly on $S^1,S^2,S^3$ up to $\eta=0.5$ under both noise
models, and up to $\eta=0.8$ for the isotropic one. The
multiplet positions are then placed by the parameter-free
angular-momentum-level shrinkage law
(see Sec.~\ref{sec:attenuation-law}), exact in the noise
amplitude, with the finite-$N$-corrected large-$K$ slope
$d/(d-1)$ as a low-noise cross-check. The embedding model,
in turn, is read from the bottom of the spectrum: the
$\ell=2$ component that FSM injects and the isotropic RSM
leaves empty identifies the noise class blind from a single
$M^{(D)}$ (see Sec.~\ref{sec:exp-blind}), corroborated by the
residual-bulk shape against the exploratory semicircle reference
(see Appendix~\ref{sec:exp-rmt}).

Two companion papers apply the same framework to real
machine-learning tasks, as a diagnostic for structural
changes in the trained model or input
data~\cite{halperin2026OMD}, and for the residual-stream
representations of modular-addition
transformers~\cite{halperin2026grokking}.

\appendix
\renewcommand{\theequation}{\thesection.\arabic{equation}}
\numberwithin{equation}{section}

\section{Detailed derivations for the RSM model}
\label{sec:appendix-RSM}

This appendix collects all derivations for the RSM model
of Sec.~\ref{sec:aaag}: (i) the residual-Gram noise moments
and pair-graph covariance; (ii) the derivation
of the RSM forward triple
\eqref{eq:aag-mu-spliced}--\eqref{eq:aag-sigma-spliced} and
the per-pair applicability constraints that bound the
$\cot,\csc$ divergence to the BBS top-$K$ window; and (iii) the double
Gegenbauer expansion of the geodesic kernel, the
angular-momentum-level shrinkage law, and the It\^o drift with its
noise-induced potential.

\subsection{Forward triple and noise moments}
\label{sec:appendix-RSM-forward}

The RSM noise is carried by the residual Gram matrix
$G_{ij}=\bar y_i\!\cdot\!\bar y_j$ of $N$ i.i.d.\ uniform
samples $\bar y_i\in S^{p-1}$ ($p:=D-d$), with $G_{ii}=1$
and $(1+G_{ij})/2\sim\mathrm{Beta}((p{-}1)/2,(p{-}1)/2)$ for
$i\neq j$ (Marsaglia--Olkin~\cite{MarsagliaOlkin1984}). From
$\E[\bar y_i\bar y_i^{\top}]=I_p/p$ and independence of
$\bar y_i,\bar y_j$ for $i\neq j$, the moments entering the
RSM forward are exact (not asymptotic),
\begin{equation}
\E[G_{ij}]=0,\quad
\mathrm{Var}[G_{ij}]=\tfrac{1}{p},\quad
\mathrm{Cov}[G_{ij},G_{kl}]=\tfrac{1}{p}\,
\delta_{\{i,j\},\{k,l\}}\ \ (i\neq j,\ k\neq l),
\label{eq:appG-moments-exact}
\end{equation}
so the pair-graph covariance is pair-diagonal, with zero
shared-vertex contribution
$\E[(\bar y_i^{\top}\bar y_j)(\bar y_j^{\top}\bar y_l)]=0$
($i\neq j\neq l$); the off-diagonal noise carries the
pair-graph structure of the $N$-particle configuration
($N{-}1$ entries share each index), shared-vertex corrections
entering only sub-leading through the $\arccos$ non-linearity
and quartic Wick contractions, reaching $O(1)$ only for a
low-rank residual covariance. These moments feed the forward
triple \eqref{eq:aag-mu-spliced}--\eqref{eq:aag-sigma-spliced};
the bulk of $G$ is the Marchenko--Pastur law, the uniform
($C=I$) case of the deformed-MP saddle of
Appendix~\ref{sec:appendix-beta-saddle}.

Take Eq.~\eqref{eq:aag-final-kernel-W} of
Sec.~\ref{sec:aaag} as the starting point. Separating the
noiseless piece from the $\varepsilon^{2}$ perturbation, the
cosine-kernel argument has the form
$\cos M^{(d)}_{ij}+y_{ij}$ with
$y_{ij}:=\varepsilon^{2}\bigl(\cos M^{(D-d)}_{ij}(\bar y)-\cos M^{(d)}_{ij}\bigr)$.
Inverting
through the entry-wise $\arccos$, $M^{(D)}_{ij}=
\arccos(\cos M^{(d)}_{ij}+y_{ij})$, with $|y_{ij}|$
small relative to the distance of $\cos M^{(d)}_{ij}$ from the
branch points $\pm1$ (the PT control parameter, satisfied for
$M^{(d)}_{ij}$ bounded away from $0$ and $\pi$), Taylor-expand the
$\arccos$ around the unperturbed argument
$\cos M^{(d)}_{ij}$ in $M^{(d)}_{ij}$ itself via
\begin{equation}
\arccos\!\bigl(\cos M^{(d)}_{ij}+y_{ij}\bigr)
\;=\;
M^{(d)}_{ij}
\;-\;
\frac{y_{ij}}{\sin M^{(d)}_{ij}}
\;-\;
\frac{\cos M^{(d)}_{ij}}
{2\sin^{3}M^{(d)}_{ij}}\,y_{ij}^{2}
\;+\;
O(y_{ij}^{3}),
\label{eq:arccos-Taylor-Mij}
\end{equation}
and substitute $y_{ij}$ from its definition above.
The $y_{ij}^{2}$ term is $O(\varepsilon^{4})$ and
drops at the order kept here. The linear term gives
\begin{equation}
M^{(D)}_{ij}
\;=\;
M^{(d)}_{ij}
\;+\;
\frac{\varepsilon_i^{2} + \varepsilon_j^{2}}{2}\,
\cot M^{(d)}_{ij}
\;-\;
\frac{\varepsilon_i\varepsilon_j}
{\sin M^{(d)}_{ij}}\,
G_{ij}
\;+\;
O(\varepsilon^{4}),
\label{eq:aag-Mij-expansion}
\end{equation}
valid throughout $M^{(d)}_{ij}\in(0,\pi)\setminus
\{\pi/2\}$. The symmetric
$(\varepsilon_i^{2}+\varepsilon_j^{2})/2$ piece is
deterministic and reduces to
$\varepsilon^{2}\cot M^{(d)}_{ij}$ at common
amplitude $\varepsilon_i=\varepsilon$; the
cross-term carries the stochastic noise through
$G_{ij}$.

At $i=j$ the kernel
\eqref{eq:aag-final-kernel-W} has argument $1$ ($G_{ii}=1$),
so $M^{(D)}_{ii}=0$ exactly; the Taylor formula
\eqref{eq:aag-Mij-expansion} is $0/0$ there (limit $0$) and
applies off-diagonal only.
Near the boundaries the half-angle identity
\eqref{eq:aag-half-angle} of Sec.~\ref{sec:aaag}, at
leading order in $\varepsilon$ and small angles, gives
\begin{equation}
M^{(D)}_{ij}
\;\simeq\;
\sqrt{(M^{(d)}_{ij})^{2}
+(\varepsilon_i^{2}+\varepsilon_j^{2})
-2\varepsilon_i\varepsilon_j\,G_{ij}\,}
\label{eq:aag-Mij-tail-nonlinear}
\end{equation}
(right tail: $m_{ij}=\pi-M^{(d)}_{ij}$,
$G_{ij}\to-G_{ij}$). Linearising in $G_{ij}$ gives the
$\mu^{\rm B},\sigma^{\rm B}$ branches of
\eqref{eq:aag-mu-spliced}--\eqref{eq:aag-sigma-spliced};
like the bulk form these are off-diagonal only, the diagonal
taken from \eqref{eq:aag-final-kernel-W} directly.

\begin{figure}[!htbp]
\centering
\includegraphics[width=0.82\textwidth]{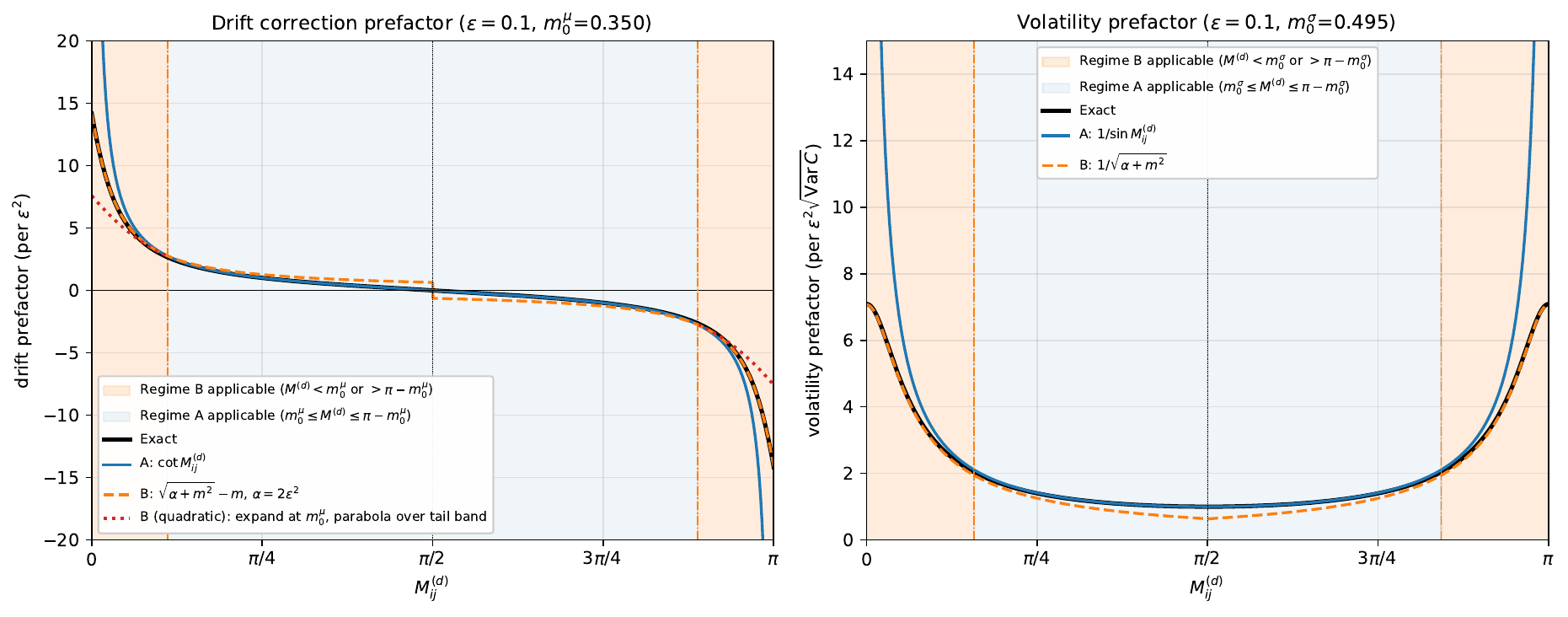}
\caption{RSM forward drift (left) and volatility (right)
correction prefactors versus $M^{(d)}_{ij}$ at
$\varepsilon = 0.1$. Black: exact deterministic prefactor from
\eqref{eq:aag-final-kernel-W}. Blue: bulk branch;
orange dashed: near-boundary branch from
\eqref{eq:aag-mu-spliced}--\eqref{eq:aag-sigma-spliced}; red
dotted (drift only): quadratic Taylor expansion of
$\mu^{\rm B}_{ij}$ about the junction. Shaded bands mark the
applicability regions \eqref{eq:regimeA-window}. Each branch
matches the exact curve to sub-percent accuracy in its region,
and the splices cover $(0,\pi)$ with bounded values.}
\label{fig:aag-regime-comparison}
\end{figure}

Splicing the bulk linear form
\eqref{eq:aag-Mij-expansion} to the near-boundary
form \eqref{eq:aag-Mij-tail-nonlinear} (linearised in $G_{ij}$)
at junction angles $m_{0}^{\mu,\sigma}=O(\sqrt\varepsilon)$, with
$m_{ij}:=M^{(d)}_{ij}$ near the left boundary and
$m_{ij}:=\pi-M^{(d)}_{ij}$ near the right, gives the entry-wise
forward drift and volatility
\begin{equation}
\mu_{ij}\bigl(M^{(d)}_{ij}\bigr)
=
\begin{cases}
M^{(d)}_{ij}+\varepsilon^{2}\cot M^{(d)}_{ij},
& m_{0}^{\mu}\leq M^{(d)}_{ij}\leq\pi-m_{0}^{\mu},\\[4pt]
M^{(d)}_{ij}\pm\bigl[\sqrt{2\varepsilon^{2}+m_{ij}^{2}}-m_{ij}\bigr],
& \text{otherwise},
\end{cases}
\label{eq:aag-mu-spliced}
\end{equation}
\begin{equation}
\sigma_{ij}\bigl(M^{(d)}_{ij}\bigr)
=
\begin{cases}
\varepsilon^{2}/\sin M^{(d)}_{ij},
& m_{0}^{\sigma}\leq M^{(d)}_{ij}\leq\pi-m_{0}^{\sigma},\\[4pt]
\varepsilon^{2}/\sqrt{2\varepsilon^{2}+m_{ij}^{2}},
& \text{otherwise},
\end{cases}
\label{eq:aag-sigma-spliced}
\end{equation}
with the upper/lower sign for the left/right tail.
Figure~\ref{fig:aag-regime-comparison} compares the two spliced
branches to the exact prefactors at $\varepsilon=0.1$, each
matching to sub-percent accuracy in its region and bounded
across $(0,\pi)$.

Two per-pair constraints control
\eqref{eq:aag-Mij-expansion}, without collapsing
$\varepsilon_i,\varepsilon_j$ to a common value. The control is
set by the distance of the expansion point $\cos M^{(d)}_{ij}$
from the branch points $\pm1$ of $\arccos$, since the coefficients
in \eqref{eq:arccos-Taylor-Mij} diverge only as
$\sin M^{(d)}_{ij}\to0$ and $\arccos$ is analytic at the midpoint
$\cos M^{(d)}_{ij}=0$. The linear term dominates the quadratic
when
$|y_{ij}|\ll 2\sin^{2}M^{(d)}_{ij}/|\cos M^{(d)}_{ij}|$, i.e.\
$\varepsilon_i\varepsilon_j\,|G_{ij}|\ll
2\sin^{2}M^{(d)}_{ij}/|\cos M^{(d)}_{ij}|$, which fails as
$M^{(d)}_{ij}\to0,\pi$ and is least restrictive at
$M^{(d)}_{ij}=\pi/2$. Smallness of the symmetric correction
relative to $M^{(d)}_{ij}$ requires
$\varepsilon_i^{2}+\varepsilon_j^{2}\ll
2 M^{(d)}_{ij}|\tan M^{(d)}_{ij}|$, which also fails as
$M^{(d)}_{ij}\to0,\pi$. Twice the first plus the second
collapses them into the single per-pair condition
\begin{equation}
(\varepsilon_i+\varepsilon_j)^{2}
\;\ll\;
\frac{4\,\sin^{2}M^{(d)}_{ij}}{|\cos M^{(d)}_{ij}|\,|G_{ij}|}
\;+\;
2\,M^{(d)}_{ij}\,\bigl|\tan M^{(d)}_{ij}\bigr|,
\label{eq:regimeA-window}
\end{equation}
whose right-hand side depends only on the angles
$M^{(d)}_{ij}$ and $M^{(D-d)}_{ij}(\bar y)$ and vanishes only at
the coincident and antipodal pairs $M^{(d)}_{ij}\to0,\pi$, not at
$\pi/2$. At common amplitude
$\varepsilon_i=\varepsilon_j=\varepsilon$ the left side is
$4\varepsilon^{2}$ and the right recovers the $\min$-form
$\varepsilon^{2}\lesssim\min(M^{(d)}_{ij}|\tan M^{(d)}_{ij}|,
\sin^{2}M^{(d)}_{ij}/|\cos M^{(d)}_{ij}|)$, both factors
controlled by $\sin M^{(d)}_{ij}$ near the endpoints. The BBS
diagnostics of Sec.~\ref{sec:experiments} restrict to the
top-$K$ window where \eqref{eq:regimeA-window} holds.

\subsection{Gegenbauer addition theorem for the geodesic kernel}
\label{sec:appendix-addition}

We derive the double Gegenbauer expansion
\eqref{eq:aag-addition} of the geodesic ambient kernel
\eqref{eq:aag-MD-geodesic},
$M^{(D)}_{ij}=\arccos(\cos^{2}\theta\,t_1+\sin^{2}\theta\,t_2)$
with $t_1=\cos M^{(d)}_{ij}$,
$t_2=\cos M^{(D-d)}_{ij}(\bar y)$ and $\sin\theta=\varepsilon$.
The latent sphere $S^{d-1}$ and residual sphere
$S^{D-d-1}$ carry the ultraspherical indices
$\mu=(d-2)/2$ and $\rho=(D-d-2)/2$; the ambient sphere
$S^{D-1}$ carries $\nu=(D-2)/2=\mu+\rho+1$.

By
\eqref{eq:aag-uv-parametrisation},
$X_i=\cos\theta\,W_0\tilde x_i+\sin\theta\,V_0\bar y_i$, so
every ambient point lies at the common latitude $\theta$ on
the orthogonal join $S^{d-1}\ast S^{D-d-1}\subset S^{D-1}$,
and $X_i\!\cdot\!X_j=\cos^{2}\theta\,t_1+\sin^{2}\theta\,t_2$.
The geodesic distance is a zonal function on $S^{D-1}$, with
the Gegenbauer expansion of BBS-2 \cite{bogomolny2007},
\begin{equation}
\arccos(w)=\frac{\pi}{2}+\sum_{n\ \mathrm{odd}}b_n\,
C_n^{\nu}(w),
\qquad
b_n=-\frac{(D-2)(D-2+2n)}{8\pi}
\left[\frac{\Gamma(\tfrac{D-2}{2})\,\Gamma(\tfrac n2)}
{\Gamma(1+\tfrac{n+D-2}{2})}\right]^{2},
\label{eq:appadd-arccos-amb}
\end{equation}
the even-$n$ coefficients vanishing (BBS-2 Eq.~(70), their
$p=D-2$). Restricted to the common-latitude join, each
ambient zonal $C_n^{\nu}(X_i\!\cdot\!X_j)$ decomposes under
the branching $SO(D)\downarrow SO(d)\times SO(D-d)$, by the
zonal addition theorem applied on each factor (BBS-2
Eq.~(57); Erd\'elyi et al.~\cite{Erdelyi1953v2}, 11.4.2),
\begin{equation}
\begin{aligned}
C_n^{\nu}\!\bigl(\cos^{2}\theta\,t_1+\sin^{2}\theta\,t_2\bigr)
&=\!\!\sum_{\substack{p+q\le n\\ n-p-q\ \mathrm{even}}}\!\!
G^{(n)}_{pq}(\theta)\,C_p^{\mu}(t_1)\,C_q^{\rho}(t_2),
\\[2pt]
G^{(n)}_{pq}(\theta)&=g^{(n)}_{pq}\,(\cos\theta)^{2p}
(\sin\theta)^{2q}\bigl[j^{(n)}_{pq}(\cos2\theta)\bigr]^{2},
\end{aligned}
\label{eq:appadd-branch}
\end{equation}
where $j^{(n)}_{pq}$ is the Jacobi radial polynomial of
degree $(n-p-q)/2$ of the branching and $g^{(n)}_{pq}>0$.
Inserting \eqref{eq:appadd-arccos-amb} and resumming over
$n$ gives \eqref{eq:aag-addition} with
$\Phi_{pq}(\theta)=\sum_{k\ge0}b_{p+q+2k}\,
G^{(p+q+2k)}_{pq}(\theta)$.

Equivalently, the
products $\{C_p^{\mu}(t_1)C_q^{\rho}(t_2)\}$ are a complete
orthogonal basis of $L^{2}([-1,1]^2,\,w_\mu\!\times\!w_\rho)$
with $w_\lambda(t)=(1-t^2)^{\lambda-1/2}$, so the coefficients
are the projections
\begin{equation}
\begin{aligned}
\Phi_{pq}(\theta)=
\frac{1}{\|C_p^{\mu}\|^{2}\,\|C_q^{\rho}\|^{2}}
\int_{-1}^{1}\!\!\int_{-1}^{1}
&\arccos\!\bigl(\cos^{2}\theta\,t_1+\sin^{2}\theta\,t_2\bigr)
\\[2pt]
&\times\,C_p^{\mu}(t_1)\,C_q^{\rho}(t_2)\,
w_\mu(t_1)\,w_\rho(t_2)\,\dd t_1\,\dd t_2,
\end{aligned}
\label{eq:appadd-ortho}
\end{equation}
with $\|C_l^{\lambda}\|^{2}$ from BBS-2 Eq.~(63)
(Erd\'elyi 10.9.7). Eq.~\eqref{eq:appadd-ortho} is the form
evaluated numerically; the two routes agree, and a
truncated \eqref{eq:aag-addition} reconstructs the kernel to
better than $10^{-3}$ relative accuracy at modest order
($p,q\le9$).
Under $(t_1,t_2)\to(-t_1,-t_2)$ the
argument $w\to-w$, while $C_l^{\lambda}(-t)=(-1)^{l}
C_l^{\lambda}(t)$; since $\arccos(w)-\pi/2$ is odd, the
integrand of \eqref{eq:appadd-ortho} is even only when
$p+q$ is odd. Hence $\Phi_{pq}=0$ unless $p+q$ is odd, the
single exception being the mean $\Phi_{00}=\pi/2$.

For numerical work the coefficients drop steeply at low order
($\Phi_{10}\approx0.90$, $\Phi_{30}\approx0.04$,
$\Phi_{50}\approx8\times10^{-3}$ at $\varepsilon=0.45$), so
the first few odd degrees already capture the dominant
quasi-multiplets. The uniform reconstruction error decays
algebraically, $\sim K^{-5/2}$ in the joint truncation
$K=p=q$, the power-law (rather than exponential) rate being
set by the square-root endpoint singularity of $\arccos$ at
$w=\pm1$. A handful of terms therefore suffices for the few
largest-$|\Lambda|$ multiplets used in BBS inference; the
slowly-converging algebraic tail corresponds to the
small-$|\Lambda|$ bulk that the top-$K$ window discards.

For
$\bar y_i$ uniform on $S^{D-d-1}$ the residual cosine kernel
$t_2$ has the marginal density $\propto w_\rho$ by the
Marsaglia--Olkin moments \eqref{eq:appG-moments-exact}, so
$\langle C_q^{\rho}(t_2)\rangle=\delta_{q0}$ and
\eqref{eq:aag-addition} averages to the latent series
\eqref{eq:aag-addition-averaged}, with only odd $p$
(matching the odd-$\ell$ rule of
\eqref{eq:appadd-arccos-amb}). The degree-$p$ term is the
BBS quasi-multiplet of multiplicity $h(p,d)$
\eqref{eq:harmonics-mult}; at $\theta=0$, $M^{(D)}=M^{(d)}$
and $\Phi_{p0}(0)$ is the latent Funk--Hecke coefficient, so
the multiplet is shrunk by $f_p(\theta)=
\Phi_{p0}(\theta)/\Phi_{p0}(0)$, the angular-momentum-level shrinkage
law of Sec.~\ref{sec:attenuation-law}.

How many terms of \eqref{eq:aag-addition-averaged} are needed to reproduce the
realised drift? We answer this on a single fixed random residual realisation
(latent $S^{2}$, residual $S^{3}$, $\varepsilon=0.45$, $N=1000$). In the
angle-remapped Gegenbauer basis $\mu(M^{(d)}_{ij})=\sum_\ell \tilde c_\ell\,
P_\ell^{(d-2)/2}(z)$, $z=(2/\pi)M^{(d)}_{ij}-1$ (the
angle-remapped Gegenbauer basis), the drift is odd about
$M^{(d)}=\pi/2$ because $\arccos$ is odd about its midpoint, so only odd $\ell$
contribute and the even coefficients vanish: $\tilde c_2,\tilde c_4$ are zero to
machine precision, and the empirical values, obtained by binning the simulated
$M^{(D)}_{ij}$ in $z$ and projecting the per-bin conditional mean, are
consistent with zero (see Table~\ref{tab:drift-coeffs}). The odd coefficients fall
off geometrically, $\tilde c_1=1.10$, $\tilde c_3=-0.13$, $\tilde c_5=-0.021$,
$\tilde c_7=-0.0010$, each roughly an order of magnitude below the last, and the
exact (quadrature) and empirical coefficients agree to within the finite-$N$
scatter.

\begin{figure}[!htbp]
\centering
\includegraphics[width=0.82\textwidth]{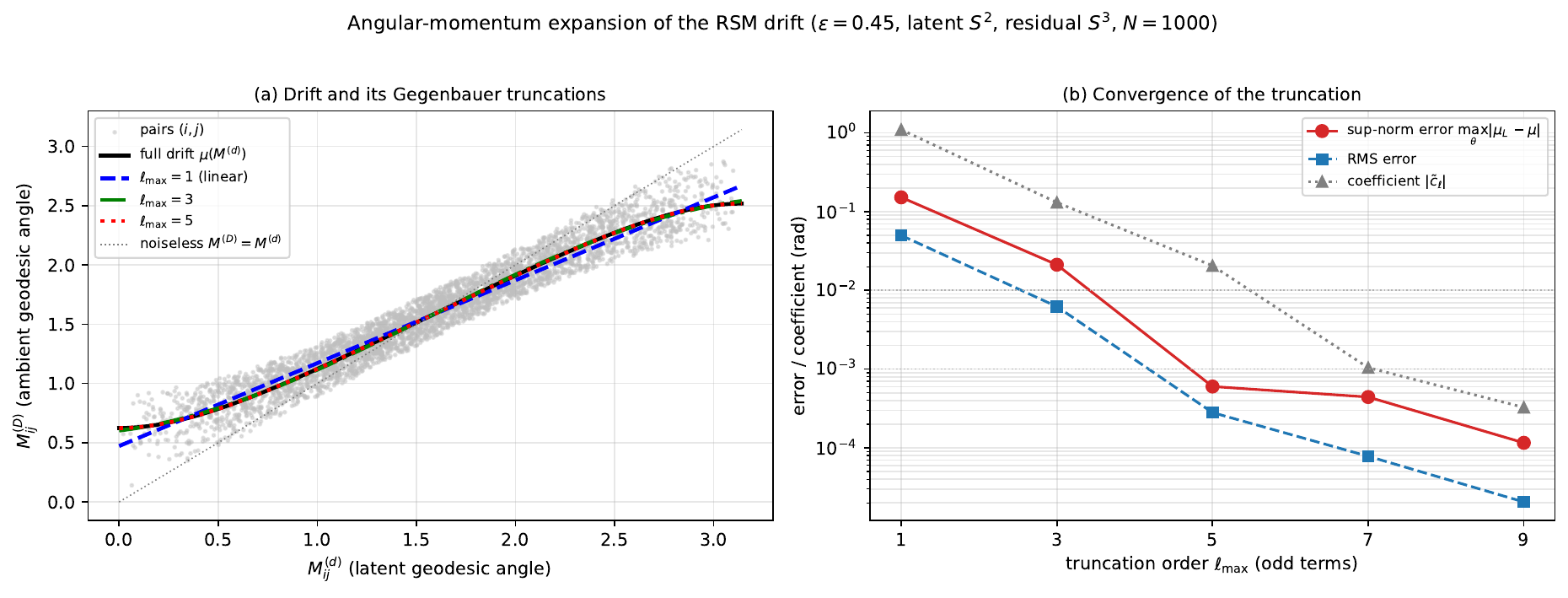}
\caption{Convergence of the angular-momentum Gegenbauer expansion of the RSM
drift (one fixed random residual; latent $S^{2}$, residual $S^{3}$,
$\varepsilon=0.45$, $N=1000$). (a) The realised pairs
$(M^{(d)}_{ij},M^{(D)}_{ij})$ (grey), the full drift $\mu(M^{(d)})$ (black), and
its truncations at $\ell_{\max}=1,3,5$: the linear truncation misses the
endpoint curvature, $\ell_{\max}=3$ tracks the drift to about $1\%$, and
$\ell_{\max}=5$ is indistinguishable from it. (b) Truncation error versus order:
the sup-norm and RMS deviations from the full drift and the coefficient
magnitude $|\tilde c_\ell|$. Only odd $\ell$ contribute, and the error follows
the largest omitted coefficient, falling from $0.15$~rad at $\ell_{\max}=1$ to
$0.02$~rad at $\ell_{\max}=3$ and $6\times10^{-4}$~rad at $\ell_{\max}=5$.}
\label{fig:drift-convergence}
\end{figure}

The truncation error follows the largest omitted coefficient
(see Fig.~\ref{fig:drift-convergence}). The $\ell_{\max}=1$ truncation (the linear
drift) leaves a sup-norm error of $0.15$~rad, concentrated near the endpoints
where the curvature lives. The cubic ($\ell_{\max}=3$) truncation cuts this to
$0.02$~rad, about $1\%$ of the angular range, and the $\ell_{\max}=5$ truncation
reaches $6\times10^{-4}$~rad, indistinguishable from the full drift. Three odd
terms thus give a percent-level description and five an essentially exact one.
At $N=1000$ this is also the practical resolution limit: the realisation scatter
on each coefficient is $\approx0.002$, so $\tilde c_5$ is the last coefficient
resolved above the Monte-Carlo noise while $\tilde c_7$ and beyond lie below it.
The leading correction to the linear drift is the parity-allowed cubic term, and
the expansion has effectively converged by $\ell=5$.

\begin{table}[!htbp]
\centering
\caption{Angle-remapped Gegenbauer coefficients
$\tilde c_\ell$ of the RSM drift (latent $S^{2}$, residual
$S^{3}$, $\varepsilon=0.45$, $N=1000$): exact (quadrature)
against the empirical mean $\pm$ standard deviation over
$40$ fixed random residual realisations. Even-$\ell$
coefficients vanish by parity (empirical values consistent with zero); the odd
coefficients decay geometrically and agree across methods down to the
$\approx0.002$ realisation scatter, which $\tilde c_7$ already reaches.}
\label{tab:drift-coeffs}
\begin{tabular}{c c c}
\hline
$\ell$ & exact & empirical (mean $\pm$ std)\\
\hline
$0$ & $+1.5708$ & $+1.5707 \pm 0.0004$\\
$1$ & $+1.0989$ & $+1.0979 \pm 0.0010$\\
$2$ & $\phantom{+}0.0000$ & $-0.0002 \pm 0.0016$\\
$3$ & $-0.1307$ & $-0.1312 \pm 0.0018$\\
$4$ & $\phantom{+}0.0000$ & $-0.0003 \pm 0.0022$\\
$5$ & $-0.0206$ & $-0.0210 \pm 0.0023$\\
$6$ & $\phantom{+}0.0000$ & $-0.0001 \pm 0.0024$\\
$7$ & $-0.0010$ & $-0.0013 \pm 0.0026$\\
\hline
\end{tabular}
\end{table}

\section{Detailed derivations for the FSM model}
\label{sec:appendix-FSM}

This appendix supplies, in detail, the facts behind the
main-text FSM construction \eqref{eq:fsm-psd-kernel}: that the
product basis is forced by symmetry, that the Schoenberg condition
is necessary as well as sufficient, and the degree-one
impossibility statement that explains why the higher coefficients
must be supplied at the kernel level. It closes with the drift,
the amplitude calibration, and the geodesic forward. The latent
unit vectors are $\vec x_i\in S^{d-1}$ and the residual
$\bar y_i\in S^{D-d-1}$, the zonal kernels $\mathcal Z_p^{(n)}$ are
those of \eqref{eq:fsm-zonal} with $\mathcal Z_0^{(n)}=1$ and
$\mathcal Z_1^{(n)}(t)=t$, and $W$ is the noiseless embedding with
ambient image $\vec h^{(0)}_i:=W\vec x_i$.

An $SO(d)\times SO(D-d)$-invariant kernel of the pair depends only
on the two invariants $u_{ij}=\tilde x_i\!\cdot\!\tilde x_j$ and
$v_{ij}=\bar y_i\!\cdot\!\bar y_j$, and the product Gegenbauer
functions $\{\mathcal Z_p^{(d)}(u)\,\mathcal Z_q^{(D-d)}(v)\}$ are
a complete orthogonal basis for such kernels
(see Appendix~\ref{sec:appendix-addition}), so every invariant cosine
kernel has the form \eqref{eq:fsm-psd-kernel} and the array
$\{\beta_{pq}\}$ is the only model freedom. By Schoenberg's theorem
a single-factor zonal kernel $\sum_p b_p\mathcal Z_p^{(n)}$ is
positive definite, for every configuration and every $N$, if and
only if every $b_p$ is nonnegative \cite{Schoenberg1942}.
Sufficiency on the product is the addition-theorem and Schur
argument of the main text. Necessity carries over to the product:
concentrating the latent and residual points so the quadratic form
localises on the bidegree-$(p,q)$ product harmonic drives that
block negative whenever $\beta_{pq}<0$. The complete positive class
is therefore the nonnegative simplex, the sum-to-one condition
fixing the unit diagonal, and RSM is its $\{(1,0),(0,1)\}$ corner.

The higher coefficients carry genuine cross-particle dependence,
which is why they cannot be reproduced by independent ambient
sampling. Suppose each ambient point $\vec X_k\in S^{D-1}$ were
drawn independently from a per-particle rotation-invariant law
centred at $\vec h^{(0)}_k$. Rotation invariance about
$\vec h^{(0)}_k$ gives $\E[\vec X_k\mid\vec x_k]=a\,\vec h^{(0)}_k$
for a scalar contraction $a\in[0,1]$ independent of $\vec x_k$,
and independence of distinct particles gives the conditional-mean
kernel
\begin{equation}
\E\!\bigl[\cos M^{(D)}_{ij}\mid\vec x_i,\vec x_j\bigr]
=\E[\vec X_i\mid\vec x_i]^{\top}\E[\vec X_j\mid\vec x_j]
=a^{2}\,\vec h^{(0)\top}_i\vec h^{(0)}_j
=a^{2}\,\mathcal Z_1^{(d)}(u_{ij}),
\qquad i\neq j,
\label{eq:fsm-ell1-impossible}
\end{equation}
a pure degree-one latent sector, so $\beta_{p0}=0$ for every
$p\geq 2$. Any FSM with $\beta_{p0}>0$ at some $p\geq 2$ thus
cannot arise from independent per-particle ambient sampling: it
carries cross-particle statistical dependence. The PSD-kernel
construction \eqref{eq:fsm-psd-kernel} supplies that dependence
directly, reaching the higher coefficients that independent
per-particle sampling cannot, while remaining positive
semidefinite by construction.

The conditional-mean drift, the amplitude, and the covariance
follow from the residual average. Averaging over the residual
coordinates $\{\bar y_k\}$, uniform on $S^{D-d-1}$, collapses
every residual zonal to its mean,
$\E_{\bar y}[\mathcal Z_q^{(D-d)}(v_{ij})]=\delta_{q0}$ for
$i\neq j$, leaving the drift kernel
\begin{equation}
\bar\kappa_{ij}
:=\E_{\bar y}\!\bigl[\cos M^{(D)}_{ij}\bigr]
=\sum_{p\geq 0}\beta_{p0}\,\mathcal Z_p^{(d)}(u_{ij}),
\label{eq:fsm-drift-kernel}
\end{equation}
a single zonal expansion in the latent angle whose Funk--Hecke
coefficients place the multiplet positions through the shrinkage
law of Sec.~\ref{sec:attenuation-law}. The degree-one weight sets
the amplitude, $\varepsilon^{2}:=1-\beta_{10}$, matching the RSM
top-$d$ cosine shrinkage $1-\varepsilon^{2}$. The fluctuations of
$\cos M^{(D)}_{ij}$ about $\bar\kappa_{ij}$ are not pair-diagonal:
entries sharing an index are correlated through the shared
residual coordinate, the same coupling that keeps the realised
kernel positive semidefinite. Written in the drift--volatility
form \eqref{eq:layered-generic}, FSM has drift
$\mu_{ij}=\arccos\bar\kappa_{ij}$ at leading order and volatility
$\sigma_{ij}$ the residual standard deviation, and the inverse
analysis of Sec.~\ref{sec:layer2-pert} uses only the
residual-averaged drift, which is well defined for every
nonnegative array.

The geodesic forward follows from the chain rule. Applying
$\dd\arccos(t)/\dd t=-1/\sqrt{1-t^{2}}$ to
\eqref{eq:fsm-psd-kernel} gives
\begin{equation}
M^{(D)}_{ij}
=\arccos\bar\kappa_{ij}
-\frac{\delta\kappa_{ij}}{\sqrt{1-\bar\kappa_{ij}^{2}}}
+O(\delta\kappa_{ij}^{2}),
\qquad
\delta\kappa_{ij}:=\cos M^{(D)}_{ij}-\bar\kappa_{ij},
\label{eq:fsm-geodesic-forward}
\end{equation}
with the centred residual fluctuation $\delta\kappa_{ij}$ entering
through the geometric weight $1/\sqrt{1-\bar\kappa_{ij}^{2}}=
1/\sin\bar M^{(d)}_{ij}$. For the canonical even profile
$\beta_{20}>0$ the drift $\bar\kappa$ carries a degree-two latent
zonal $\mathcal Z_2^{(d)}$, the $\ell=2$ component absent from
isotropic RSM, and the volatility is the heteroscedastic
$|\delta\kappa_{ij}|/\sin\bar M^{(d)}_{ij}$ set by the residual
variance of \eqref{eq:fsm-psd-kernel}.

\section{Spectra of the geodesic distance matrix and its off-manifold residual}
\label{sec:appendix-beta-sphere}
\label{sec:appendix-beta-saddle}

The uniform-sphere residual-Gram moments of
Appendix~\ref{sec:appendix-RSM} assume i.i.d.\ uniform samples; the
single-particle law is in general not uniform, and the geodesic
distance-matrix spectrum depends on it. We derive that spectrum for a
general law by the Euclidean-random-matrix route. Conditional on a
configuration $\{x_i\}$ the matrix is deterministic, so the joint law of
the off-diagonal entries is formally $P(\{M_{ij}\}_{i<j})=\int\prod_{i}p_i(x_i)
\,\dd x_i\prod_{i<j}\delta(M_{ij}-\arccos(x_i\!\cdot\!x_j))$,
with all randomness in the quenched positions. This is only a definition:
the $N(N{-}1)/2$ entries are functions of $N(d{-}1)$ coordinates, so for
the noiseless $M^{(d)}$ the joint entry density is singular (supported on
the rank-$d$ distance-matrix manifold) and is never computed. As is
standard for Euclidean random matrices \cite{Mezard1999, Goetschy2013},
the spectral statistics follow not from $P(M)$ but by averaging over the
positions: one conditions on $\{x_i\}$ and integrates over the
single-particle law at the end, the saddle controlled by the density
$\bar p=\tfrac1N\sum_i p_i$. The spectrum splits into a deterministic,
self-averaging \emph{signal} (the top eigenvalues, carried by the mean
operator of $\bar p$; Sec.~\ref{sec:appendix-mercer-bbs}) and a continuous
\emph{bulk} from the entrywise fluctuations
(see Sec.~\ref{sec:appendix-bulk}). This is the large-$N$ saddle point of the
replica analysis of \cite{Mezard1999}; we obtain each part directly, and
first fix a concrete family of non-uniform laws to test against.

\subsection{The product-Beta ensemble}

Each particle $i$ is placed on $S^{D-1}\subset\R^{D}$ through its $D-1$
hyperspherical angles, drawn independently from Beta distributions,
\begin{equation}
p_i(\theta) = \prod_{k=1}^{D-1}\mathrm{Beta}(\theta_k;\alpha_i^k,\beta_i^k),
\qquad \theta_k\in[0,1],
\label{eq:beta-product}
\end{equation}
with $\phi_k=\pi\theta_k$ ($k<D-1$) and $\phi_{D-1}=2\pi\theta_{D-1}$ mapped to
Cartesian coordinates by the standard hyperspherical map. The symmetric
baseline $\alpha=\beta=2$ gives a smooth concentrated law; varying the
parameters tilts and sharpens it per angle and per particle. We consider
four anisotropy patterns, all varied parameters drawn from
$\mathrm{Uniform}[1,3]$ and redrawn per realisation: common to all
particles and angles (homogeneous), varying across particles only
(per-particle anisotropy), across angles only (per-angle anisotropy), and
across both (per-particle-and-angle anisotropy). The product-Beta law in
hyperspherical angles is coordinate-dependent, and the sphere-coverage
fraction it defines is likewise tied to that chart, so the ensemble serves
as a stress test of the uniform-sampling assumption (in the
misspecification sense of Sec.~\ref{sec:appendix-nonuniform}) and not as
evidence of density-independent geometry inference.

\begin{figure}[!htbp]
\centering
\includegraphics[width=0.85\textwidth]{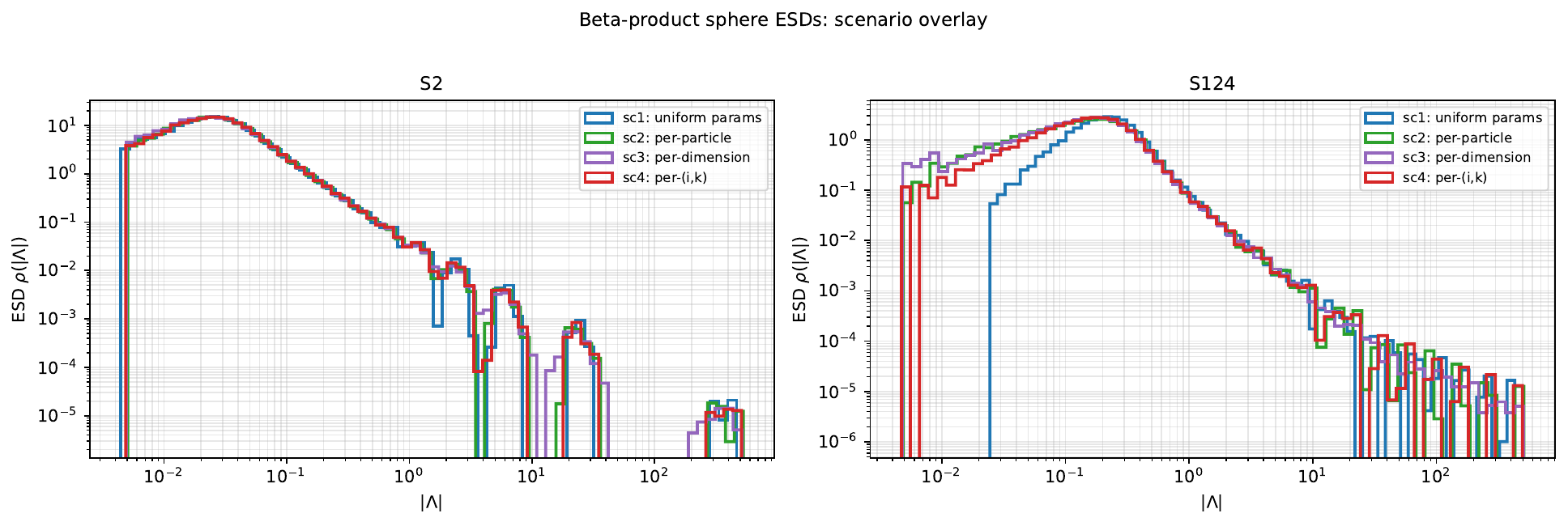}
\caption{Empirical spectral densities of the negative
eigenvalues of the geodesic distance matrix
$M_{ij}=\arccos(x_i\!\cdot\!x_j)$ for the product-Beta
single-particle law \eqref{eq:beta-product}, on $S^{2}$
(left) and $S^{124}$ (right), $N=1000$, 20 realisations
pooled, Perron eigenvalue excluded. The four patterns are
common parameters (sc1), per-particle anisotropy (sc2),
per-angle anisotropy (sc3), and per-particle-and-angle
anisotropy (sc4), with varied parameters drawn from
$\mathrm{Uniform}[1,3]$ about the baseline
$\alpha=\beta=2$. On $S^{2}$ the patterns nearly coincide;
on $S^{124}$ the anisotropic patterns add a
small-$|\Lambda|$ tail that the homogeneous case lacks.}
\label{fig:beta-sphere-esd}
\end{figure}

Figure~\ref{fig:beta-sphere-esd} shows the pooled empirical densities of
the negative eigenvalues of $M$ ($N=1000$, 20 realisations) on $S^2$ and
$S^{124}$. On $S^2$ the four patterns nearly coincide: with two angles the
anisotropy leaves the BBS power-law bulk and the $SO(3)$ quasi-multiplet
spikes unchanged. On $S^{124}$ they separate, the anisotropic patterns
filling a small-$|\Lambda|$ tail absent in the homogeneous case. The
single-particle law thus controls the small-$|\Lambda|$ part of the
spectrum while the large-$|\Lambda|$ multiplet structure is rigid. These
stored spectra are the reference against which the calculation below is
verified.

\subsection{Signal: the operator spectrum and the BBS power law}
\label{sec:appendix-mercer-bbs}

As $N\to\infty$ the eigenvalue problem of the $N\times N$ matrix $M$ is
approximated by that of an integral operator, which is what brings the
Funk--Hecke theorem and the spherical-harmonic expansion into the
analysis. The matrix entries are values of the coupling
$\arccos(x\!\cdot\!y)$, and read as an integral operator weighted by the
population density $\bar p$ this coupling is
\begin{equation}
(\mathcal T_{\bar p}g)(x)=\int \arccos(x\!\cdot\!y)\,g(y)\,\bar p(y)\,\dd y.
\label{eq:appC-operator}
\end{equation}
The matrix $\tfrac1N M$ is the Nystr\"om approximation of
$\mathcal T_{\bar p}$ built from the $N$ sample points $\{x_i\}$ (the
i.i.d.\ draw from $\bar p$), which serve as both the quadrature nodes and
the evaluation points: by the law of large numbers the row average
$\tfrac1N\sum_j\arccos(x_i\!\cdot\!x_j)g(x_j)$ converges to its expectation
$(\mathcal T_{\bar p}g)(x_i)$, so the matrix eigenvalues converge to the
operator's,
$\Lambda_K(M)\to N\,\mathrm{eig}(\mathcal T_{\bar p})$, set by $\bar p$
alone; the finite-$N$ deviations are the corrections of
Appendix~\ref{sec:finite-N}. Figure~\ref{fig:beta-operator} confirms the
limit: the stored top eigenvalues land on an independent large-sample
estimate of $\mathcal T_{\bar p}$ to a few percent on both spheres, the
per-angle pattern showing the larger scatter of a quenched anisotropy. The
multiplet structure that carries the dimension thus reads off the operator
of the actual law. For uniform $\bar p$ this operator is zonal and is
diagonalised exactly by spherical harmonics, with eigenvalues given by
Funk--Hecke; this is the route to the BBS power law, derived next.

\begin{figure}[!htbp]
\centering
\includegraphics[width=0.80\textwidth]{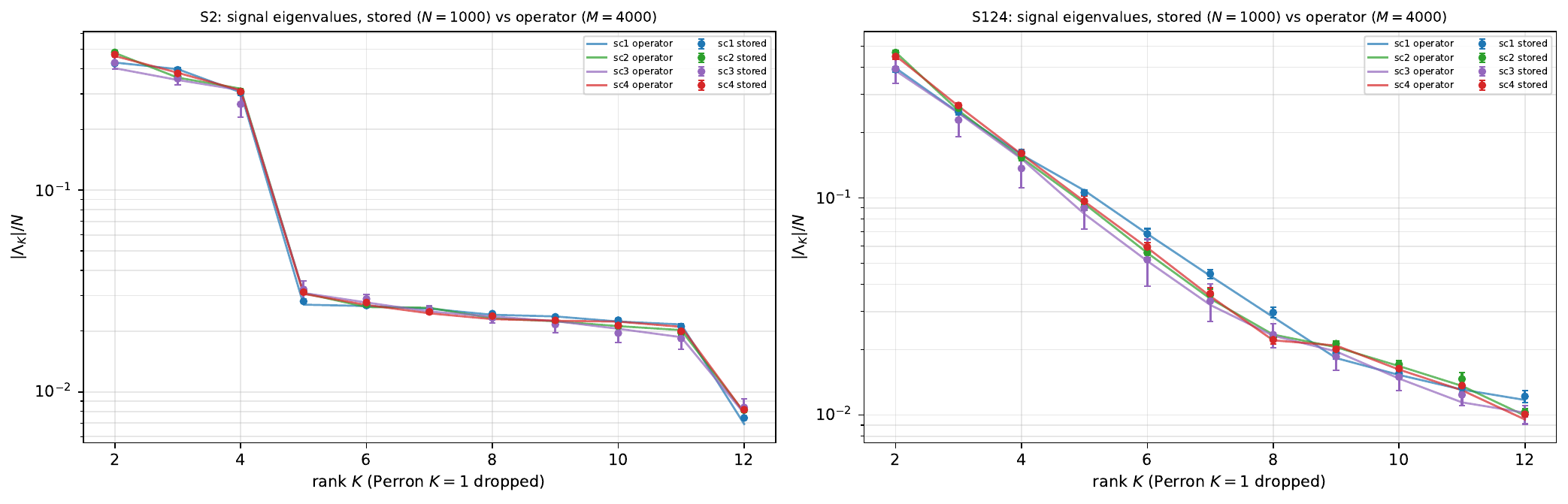}
\caption{Signal limit: the top eigenvalues $|\Lambda_K|/N$ of the stored
matrices ($N=1000$, markers with error bars over 20 realisations) against the
operator spectrum $\mathrm{eig}(\mathcal T_{\bar p})$ from an independent
$M=4000$ sample (lines), on $S^2$ (left) and $S^{124}$ (right). The four
single-particle patterns are reproduced by their respective $\bar p$; the
per-angle pattern (sc3) shows the larger scatter expected of a quenched
anisotropy.}
\label{fig:beta-operator}
\end{figure}

\subsection{Nonuniform sampling as misspecification: splitting of the multiplet}
\label{sec:appendix-nonuniform}

The multiplicity readout assumes uniform sampling, and that assumption is not
cosmetic. The operator $\mathcal T_q$ with a general density $q$ is equivariant
under rotation, $U_R\,\mathcal T_q\,U_R^{-1}=\mathcal T_{q\circ R^{-1}}$, so it
commutes with the full $SO(d)$ only when $q$ is uniform and otherwise commutes
only with the subgroup that fixes $q$ \cite{Helgason1984}. The degeneracy of the
degree-$\ell$ multiplet is a property of an $SO(d)$ irreducible representation,
so once $q$ breaks $SO(d)$ the multiplet is no longer protected: it splits into
the irreducible representations of the residual symmetry group. This splitting
lives in the limiting operator itself, so it is a misspecification of the
sampling model rather than an ambient-noise effect, and it does not vanish as
$N\to\infty$.

The mechanism is explicit on $S^2$. Take the axial density
$q_\alpha(x)=(1+\alpha P_2(x_3))/(4\pi)$ with $P_2(z)=(3z^2-1)/2$ and $|\alpha|$
small, whose residual symmetry is the $SO(2)\times\mathbb Z_2$ that fixes the
$z$-axis. The degree-one triplet splits under that subgroup into an axial
singlet ($z$) and a transverse doublet ($x_1,x_2$). Using
$\mathbb E_0[z^2P_2(z)]=2/15$, the $q$-weighted degree-one second moments are
\begin{equation}
\mathbb E_{q}[z^2]=\tfrac13+\tfrac{2\alpha}{15},
\qquad
\mathbb E_{q}[x_1^2]=\mathbb E_{q}[x_2^2]=\tfrac13-\tfrac{\alpha}{15},
\label{eq:axial-moments}
\end{equation}
so the singlet and the doublet separate by $\alpha/5$ at first order. The
degree-one block of $\mathcal T_q$ inherits the same singlet--doublet structure:
its three eigenvalues, degenerate at $a_1=-\pi/8$ for $\alpha=0$, split with the
axial level moving as $a_1-c\,\alpha$ and the two transverse levels staying
degenerate, with $c\approx0.24$ by quadrature, so the matrix eigenvalues split
by $O(N\alpha)$. This is an operator-level shift linear in the anisotropy, not
an $O(\sqrt N)$ sampling fluctuation.

The readout therefore has a quantifiable tolerance radius. The $h(1,d)$-fold
degeneracy and the cluster count survive as long as the density-induced
splitting stays below the separation $g_*$ of Sec.~\ref{sec:product-spectrum},
which by the Weyl and Davis--Kahan bounds holds while
$\|\mathcal T_q-\mathcal T_{\rm unif}\|_{\rm op}<g_*/2$. For the axial family
$\|\mathcal T_{q_\alpha}-\mathcal T_{\rm unif}\|_{\rm op}=c\,|\alpha|+O(\alpha^2)$,
giving a tolerance $|\alpha|<\alpha_*\approx g_*/(2c)$. Beyond it the triplet
resolves into singlet plus doublet and the single-multiplet hyper-sphere
inversion misreports the dimension, though the full multiplet sequence still
identifies the geometry class through the table of
Sec.~\ref{sec:multiplet-table}. The ellipsoid experiment of
Sec.~\ref{sec:discussion} realises exactly this axial density: drawing uniform
points, stretching along $z$, and renormalising to unit directions leaves the
spherical metric unchanged and instead reweights the sampling, so its
observed singlet--doublet split at stretch $r\gtrsim1.25$ is the
$\mathcal T_q$ symmetry breaking, read at finite anisotropy past
$\alpha_*$.

We first recover the BBS power law from the Mercer--Funk--Hecke spectrum.
For uniform $\bar p$ the harmonics diagonalise $\mathcal T_{\bar p}$, and
its spectrum gives the BBS delocalised exponent $\beta_{\rm deloc}=d/(d-1)$
(see Sec.~\ref{sec:bbs-power-laws}) directly, without the counting argument of
\cite{bogomolny2003}, as a ratio of two growth rates. Here $\nu$ is the \emph{latent}
ultraspherical index $(d-2)/2$ of $S^{d-1}$, not the ambient
$(D-2)/2$ of the observed sphere $S^{D-1}$: the ambient kernel is reduced
to this latent expansion by the branching and residual average of
Appendix~\ref{sec:appendix-addition} (with no residual noise the data lie
on $S^{d-1}$ and the latent expansion is immediate). The geodesic kernel
on $S^{d-1}$ (ultraspherical index $\nu=(d-2)/2$) has the Gegenbauer
expansion \eqref{eq:appadd-arccos-amb}, $\arccos(x\!\cdot\!y)=\tfrac\pi2
+\sum_{\ell\ \mathrm{odd}}b_\ell C_\ell^{\nu}(x\!\cdot\!y)$, and the
addition theorem $C_\ell^{\nu}(x\!\cdot\!y)\propto\sum_{m=1}^{h(\ell,d)}
Y_{\ell m}(x)Y_{\ell m}(y)$ turns it into the Mercer form
\begin{equation}
\arccos(x\!\cdot\!y)=\sum_{\ell}a_\ell\sum_{m=1}^{h(\ell,d)}
Y_{\ell m}(x)\,Y_{\ell m}(y),
\label{eq:appC-mercer-kernel}
\end{equation}
with $a_\ell$ the Funk--Hecke eigenvalue on degree $\ell$
(proportional to $b_\ell$), carrying the degeneracy
$h(\ell,d)\sim\ell^{d-2}$; only odd $\ell$ contribute. The decay follows
from the square-root coincidence singularity
$\arccos(1-x)=\sqrt{2x}\,[1+O(x)]$: reading \eqref{eq:appadd-arccos-amb} on
$S^{d-1}$ (its dimension parameter set to $d-2$), the exact $b_\ell$ contain
$\Gamma(\ell/2)/\Gamma(\tfrac{\ell+d}{2})\sim(\ell/2)^{-d/2}$, so
\begin{equation}
b_\ell\sim\ell^{-(d-1)},\qquad a_\ell\sim\ell^{-d},
\label{eq:appC-aell-decay}
\end{equation}
the second after the zonal normalisation
$\|C_\ell^\nu\|^2/C_\ell^\nu(1)\sim\ell^{-1}$ (numerically $a_\ell$ decays
as $2.00, 2.98, 3.93$ on $S^{1,2,3}$; on $S^1$ it is the Fourier value
$a_\ell=-4/\pi\ell^2$). This is the generic polynomial decay of a
non-analytic zonal kernel, the same Gegenbauer asymptotics governing the
singular kernels $|t|^\alpha$ and $(1-t)^{-\alpha}$ \cite{Stepanov2017}.
Ranking the nonzero (odd-$\ell$) eigenvalues by cumulative degeneracy
$K(\ell)=\sum_{\ell'\le\ell,\,\mathrm{odd}}h(\ell',d)\sim\ell^{d-1}$,
\begin{equation}
\frac{|\Lambda_K|}N=|a_{\ell(K)}|\sim K^{-d/(d-1)},
\qquad \beta_{\rm deloc}=\frac{d}{d-1},
\label{eq:appC-bbs-power}
\end{equation}
the coefficient-decay rate $d$ over the degeneracy-growth rate $d-1$,
matching the BBS counting function \eqref{eq:BBS64}; a different
singularity order would give a different $\beta_{\rm deloc}$ through the same ratio.
Here $K$ is the rank in descending $|\Lambda|$ order, with $K=1$ the discarded
Perron eigenvalue and $K=2$ the most negative, so the delocalised branch occupies
the top of the spectrum, $K\in[2,\sqrt N]$, with the localised tail beyond. Within this
branch \eqref{eq:appC-bbs-power} is the $K\to\infty$ (large-$\ell$)
asymptote, reached only as $N\to\infty$ (when the window edge $\sqrt N$
exposes large $\ell$); at finite $N$ the small-$K$ end is steeper and the
windowed slope approaches $\beta_{\rm deloc}$ slowly from above, as discussed in
Appendix~\ref{sec:finite-N}. The same operator spectrum,
$\{a_\ell\}$ (uniform) or $\{\tau_j\}=\mathrm{eig}(\mathcal T_{\bar p})$
(general), is the population input to the bulk: signal and bulk are the
discrete and continuous parts of one self-averaging resolvent, the top
eigenvalues being its Perron and quasi-multiplet $\delta$'s
(see Appendix~\ref{sec:finite-N}).

\begin{figure}[!htbp]
\centering
\includegraphics[width=0.48\textwidth]{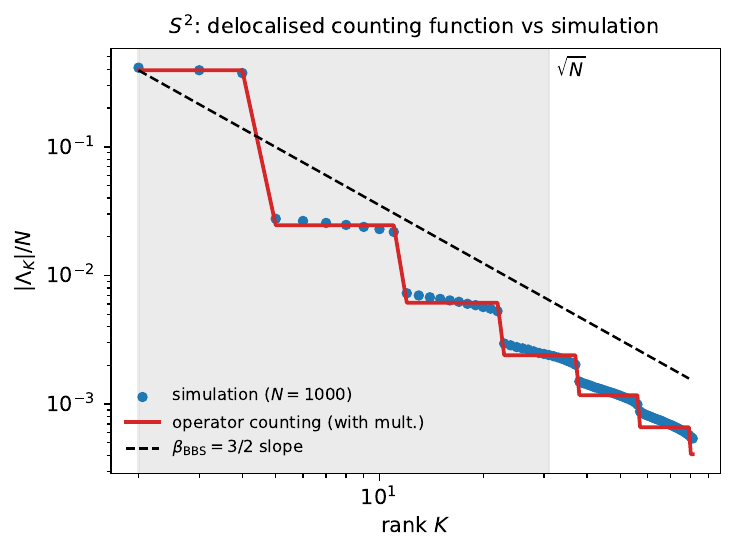}
\caption{Delocalised rank plot on $S^2$ ($N=1000$): $|\Lambda_K|/N$ versus
rank $K$ in descending $|\Lambda|$, with the Perron mode at $K=1$ omitted so
the first non-Perron multiplet is at $K=2$. Markers: eigenvalues of the simulated
matrix $M$ (mean over 16 realisations). Red line: the operator counting
function, the ranked Funk--Hecke spectrum $\{|a_\ell|\}$ with
multiplicities $2\ell+1$. Dashed: the $\beta_{\rm BBS}=3/2$ reference slope.
Shaded: the fit window $[2,\sqrt N]$. The counting function tracks the
simulation and bends from a steeper small-$K$ slope toward
$\beta_{\rm BBS}$ at larger $K$.}
\label{fig:counting-verify}
\end{figure}

We check the prediction against simulation.
The counting function makes the power law directly checkable. Ranking the
operator point masses $\{a_\ell\}$ with their multiplicities $h(\ell,d)$ in
descending $|a_\ell|$ gives the counting function and hence
$|\Lambda_K|/N=|a_{\ell(K)}|$; its slope over $K\in[2,\sqrt N]$ is the
operator prediction for $\beta_{\rm deloc}$. Table~\ref{tab:counting-verify} compares
it with the slope read the same way from the eigenvalues of the actual
random matrix $M$, and with the asymptote $\beta_{\rm BBS}=d/(d-1)$. The
operator counting and the simulation agree to about one percent, both
sitting above $\beta_{\rm BBS}$ by the finite-$N$ offset and drifting
toward it as $N$ grows ($S^2$: $1.79\to1.67$ over $N=10^3\to4\times10^3$,
$\Delta\beta\sim N^{-0.3}$; Appendix~\ref{sec:finite-N}).
Figure~\ref{fig:counting-verify} shows the $S^2$ rank plot: the operator
counting tracks the simulated $|\Lambda_K|$ and bends from a steeper
small-$K$ slope toward the $\beta_{\rm BBS}$ reference at larger $K$. The
counting route (C.6) and the resolvent (see Sec.~\ref{sec:appendix-bulk}) are
not independent of each other, both built on the same operator spectrum;
the independent check is this operator prediction against the direct
diagonalisation, which it matches.

\begin{table}[!htbp]
\centering
\caption{Delocalised slope $\beta_{\rm deloc}$ from the operator counting function
(with multiplicities) versus direct simulation of $M$, against the BBS
asymptote $\beta_{\rm BBS}=d/(d-1)$, all fit over $K\in[2,\sqrt N]$. The two
agree to $\sim1\%$ and approach $\beta_{\rm BBS}$ as $N\to\infty$.}
\label{tab:counting-verify}
\begin{tabular}{c c c c c}
\hline
sphere & $N$ & $\beta_{\rm BBS}$ & counting (with mult.) & simulation \\
\hline
$S^1$ & $1000$ & $2.000$ & $2.230$ & $2.228$ \\
$S^2$ & $1000$ & $1.500$ & $1.786$ & $1.771$ \\
$S^2$ & $2000$ & $1.500$ & $1.727$ & $1.713$ \\
$S^2$ & $4000$ & $1.500$ & $1.672$ & $1.664$ \\
$S^3$ & $1000$ & $1.333$ & $1.765$ & $1.698$ \\
\hline
\end{tabular}
\end{table}

\subsection{Bulk: a deformed Marchenko--Pastur law}
\label{sec:appendix-bulk}

Throughout this subsection $M=M^{(d)}$ is the noiseless geodesic matrix
$M_{ij}=\arccos(x_i\!\cdot\!x_j)$ on the sphere $S^{d-1}$ (not the ambient
$M^{(D)}$): the same matrix whose signal Sec.~\ref{sec:appendix-mercer-bbs}
analysed, studied here for both a low-dimensional sphere ($S^2$) and a
high-dimensional one ($S^{124}$). Its entrywise fluctuations about the
signal fill a continuous bulk, obtained from the same Mercer expansion read
as a sample-covariance model. Evaluating
\eqref{eq:appC-mercer-kernel} at the sample points, truncated to its $P$
retained modes, writes the matrix exactly as the standard ERM split
$M=HTH^\top$ into a position-random feature matrix\footnote{The feature
matrix $\Phi$ in this relation is not a Haar rotation: it is rectangular,
and its rows are the harmonics at the (sphere-image) sample points. It is
asymptotically isotropic, $\tfrac1N\Phi^\top\Phi\to C$ ($=I$ for uniform
$\bar p$), so $\Phi/\sqrt N$ behaves like a truncated-Haar (Stiefel) matrix.
Replacing it by an i.i.d.-Gaussian (or Haar--Stiefel) $H$ makes
$M=HTH^\top$ a generalised sample covariance whose bulk is the free
\emph{multiplicative} convolution of the Marchenko--Pastur law with the
population spectrum $\{\tau_j\}$ (the $S$-transforms multiply,
\cite{MingoSpeicher2017}). Computing the spectrum of an $M=\Phi A\Phi^\top$
with $\Phi$ approximated as Gaussian or Stiefel this way is standard:
M\'ezard--Parisi--Zee and Goetschy treat the Euclidean random matrix as a
generalised Wishart $HTH^\top$ \cite{Mezard1999, Goetschy2013} (Goetschy,
Eq.~65), and Couillet--Liao give the Silverstein/deformed-MP equation for
such kernel matrices \cite{CouilletLiao2022}. It is the multiplicative
counterpart of the free \emph{additive} convolution of the cosine-kernel
noise (see Sec.~\ref{sec:pt-no-single-level}), the matrix analogue of Zee's law
of addition where a square Haar rotation randomises the relative eigenbasis
\cite{Zee1996}.} and a deterministic mode matrix \cite{Mezard1999, Goetschy2013},
\begin{equation}
M=\Phi A\Phi^\top,\qquad
\Phi_{i,(\ell m)}=Y_{\ell m}(x_i)\in\R^{N\times P},\qquad
A=\mathrm{diag}(a_\ell)\in\R^{P\times P}.
\label{eq:beta-mercer}
\end{equation}
Since $\mathrm{eig}(\Phi A\Phi^\top)\setminus\{0\}
=\mathrm{eig}(A\,\Phi^\top\Phi)\setminus\{0\}$, the nonzero spectrum of
$M/N$ is that of $S^{1/2}AS^{1/2}$ with
\begin{equation}
S=\tfrac1N\Phi^\top\Phi
\;\xrightarrow{N\to\infty}\;C:=\E_{\bar p}[YY^\top]\succeq0,
\label{eq:appC-S}
\end{equation}
the sample covariance of the features, with population $C$. Read this way,
as $S^{1/2}AS^{1/2}$, $M/N$ is a \emph{signed} sample covariance: a
conventional sample covariance has a positive semidefinite population,
whereas here the weight $A$ is indefinite (the geodesic kernel is
conditionally negative definite, with one positive Perron mode $a_0>0$ and
the delocalised modes $a_\ell<0$), so the population spectrum
\begin{equation}
\{\tau_j\}=\mathrm{eig}(C^{1/2}AC^{1/2})=\mathrm{eig}(CA)
=\mathrm{eig}(\mathcal T_{\bar p})
\label{eq:appC-tau}
\end{equation}
carries both signs, negative on the delocalised branch. The last equality
holds because $AC$ is the matrix of $\mathcal T_{\bar p}$ in the harmonic
basis ($\mathcal T_{\bar p}Y_{\ell'm'}$ has coefficient
$(AC)_{(\ell m)(\ell'm')}$ on $Y_{\ell m}$, and $\mathrm{eig}(AC)
=\mathrm{eig}(CA)$): the population spectrum that deforms the bulk is the
same operator spectrum that fixes the signal
(see Sec.~\ref{sec:appendix-mercer-bbs}).

We now average over the latent positions.
The single-particle law enters only through $C$, the ``integrate last''
step. Expanding $\bar p=\sum_{LM}\mu_{LM}Y_{LM}$ with moments
$\mu_{LM}=\E_{\bar p}[Y_{LM}]$ and using the Gaunt coefficients
$\mathcal G^{LM}_{(\ell m)(\ell'm')}=\int Y_{\ell m}Y_{\ell'm'}Y_{LM}
\,\dd\sigma$,
\begin{equation}
C_{(\ell m)(\ell'm')}
=\sum_{L\geq0,M}\mu_{LM}\,\mathcal G^{LM}_{(\ell m)(\ell'm')}
=\delta_{\ell\ell'}\delta_{mm'}
+\sum_{L\geq1,M}\mu_{LM}\,\mathcal G^{LM}_{(\ell m)(\ell'm')}.
\label{eq:appC-Cgaunt}
\end{equation}
The $L=0$ term is the identity ($\mu_{00}=1$, $Y_{00}=1$), so $C-I$ is the
Gaunt contraction of the anisotropy moments $\{\mu_{LM}\}_{L\geq1}$, with
selection rules $|\ell-\ell'|\le L\le\ell+\ell'$, $M=m+m'$. For uniform
$\bar p$ all $\mu_{L\ge1}=0$, so $C=I$ and $\{\tau_j\}=\{a_\ell\}$. For the
Beta-product the moments factorise over the angles into one-dimensional
Beta integrals (azimuthal: a Beta characteristic function; polar: moments
of $P_L^M(\cos\pi\theta)$), so $C$ is analytic in the Beta parameters. On
$S^2$ it gives $\|C-I\|_F/\sqrt P\approx0.4$, growing with anisotropy.

Signal and bulk come from one resolvent.
The full spectrum is read from the self-averaging resolvent of $M/N$,
$\underline m(z)=\lim_{N\to\infty}\tfrac1N\mathrm{Tr}(M/N-z)^{-1}$ (density
$\rho(\lambda)=\tfrac1\pi\mathrm{Im}\,\underline m(\lambda+i0)$), which
separates into discrete signal poles and a continuous bulk,
\begin{equation}
\underline m(z)=\frac1N\,\frac{1}{a_0-z}
+\frac1N\sum_{\ell}\frac{h(\ell,d)}{a_\ell-z}
+\underline m_{\rm bulk}(z).
\label{eq:appC-resolvent}
\end{equation}
The Perron pole at $a_0$ and the resolved low-$\ell$ multiplet poles at
$a_\ell$ (weights $h(\ell,d)/N$) are the signal of
Sec.~\ref{sec:appendix-mercer-bbs}; their finite-$N$ shifts and widths are
computed in Appendix~\ref{sec:finite-N}. The continuous part
$\underline m_{\rm bulk}$, into which the high-$\ell$ tower merges, remains
to be determined, which is the rest of this subsection.

As a sample covariance with population $\{\tau_j\}$, the bulk is the free
multiplicative convolution of the Marchenko--Pastur law (ratio $c=P/N$)
with $H=\tfrac1P\sum_j\delta_{\tau_j}$, the multiplicative counterpart of
the additive cosine-kernel case (see Sec.~\ref{sec:pt-no-single-level}).
Marchenko--Pastur theory in Silverstein's form
\cite{BaiSilverstein2010, CouilletLiao2022}
gives $\underline m$ as the unique $\mathbb C^+$ solution of
\begin{equation}
z=-\frac1{\underline m}
+\frac1N\sum_j\frac{\tau_j}{1+\tau_j\,\underline m}
=-\frac1{\underline m}+c\!\int\frac{\tau}{1+\tau\,\underline m}\,\dd H(\tau),
\label{eq:beta-silverstein}
\end{equation}
valid for signed $\tau_j$, so the negative population places the bulk on
the negative axis; the only inputs are $\{\tau_j\}$ and $c$.

\begin{figure}[!htbp]
\centering
\includegraphics[width=0.82\textwidth]{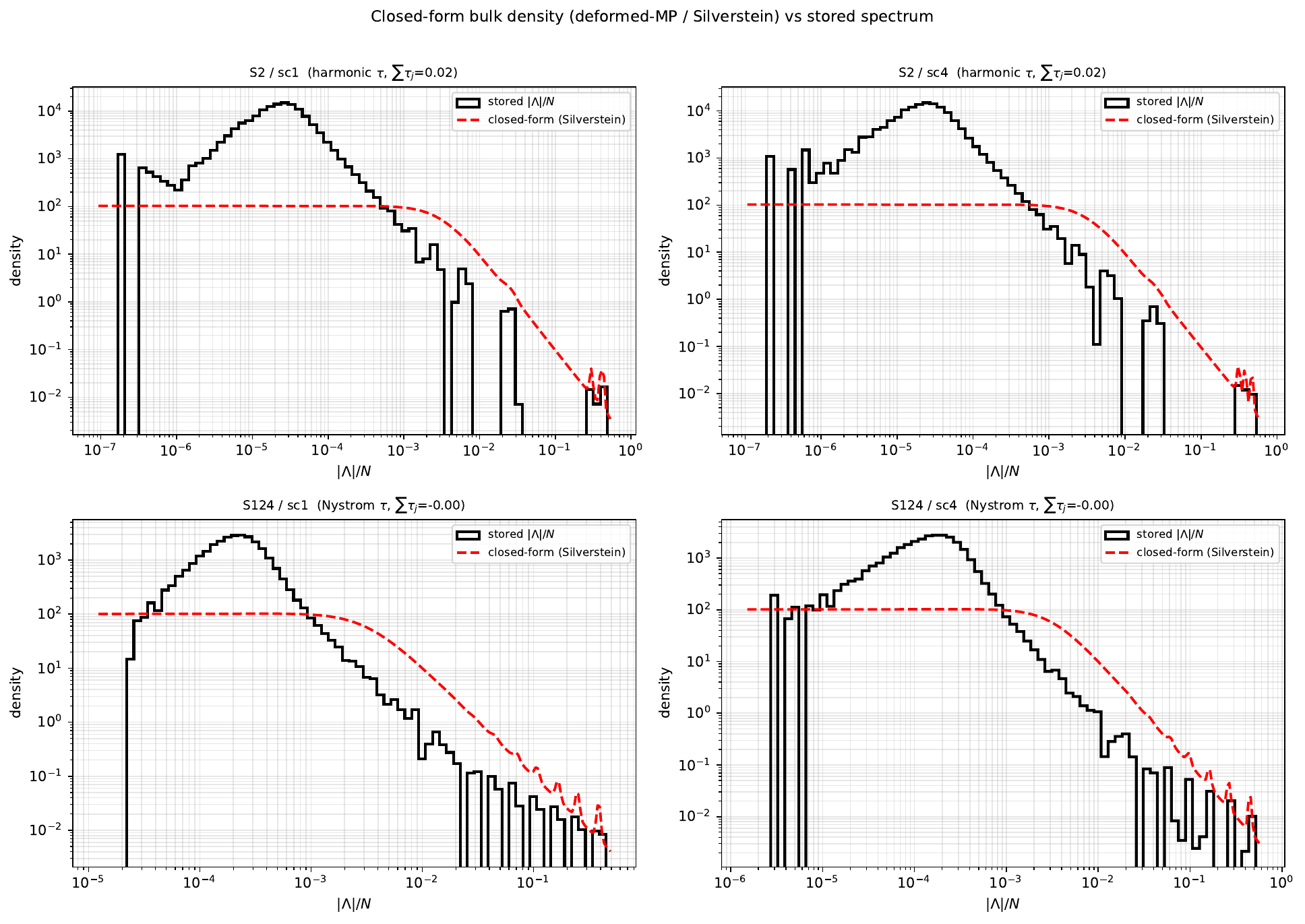}
\caption{Closed-form bulk density from the deformed-MP self-consistent equation
\eqref{eq:beta-silverstein} (dashed) versus the stored eigenvalue density
(solid), for the homogeneous (sc1) and full-anisotropy (sc4) patterns on $S^2$
(operator spectrum from harmonics) and $S^{124}$ (from a large Nystr\"om
sample). The closed form matches the signal and the bulk support; the
small-$|\Lambda|$ localized peak is a kernel-tail feature beyond the leading
deformed-MP law.}
\label{fig:beta-closed-form}
\end{figure}

Following El-Karoui \cite{ElKaroui2008}, we solve \eqref{eq:beta-silverstein}
by discretising the population into point masses, which the operator
spectrum already is: $\{\tau_k\}_{k=1}^{K}$ with multiplicities $g_k$
($\sum_k g_k=P$), on $S^2$ the harmonic eigenvalues $a_\ell$ with
degeneracies $2\ell+1$, on $S^{124}$ a large Nystr\"om sample of
$\mathcal T_{\bar p}$ coarse-grained to $K$ masses. Equation
\eqref{eq:beta-silverstein} is then the finite sum
\begin{equation}
z=-\frac1{\underline m}
+\frac1N\sum_{k=1}^{K}\frac{g_k\,\tau_k}{1+\tau_k\,\underline m},
\label{eq:appC-mp-discrete}
\end{equation}
and multiplying through by $\underline m\prod_k(1+\tau_k\underline m)$
gives the polynomial actually solved at each probe $z=\lambda+i0^+$,
\begin{equation}
(z\,\underline m+1)\prod_{k=1}^{K}(1+\tau_k\underline m)
\;-\;\frac{\underline m}{N}\sum_{k=1}^{K}g_k\tau_k\!\!
\prod_{l\neq k}(1+\tau_l\underline m)\;=\;0,
\label{eq:appC-mp-poly}
\end{equation}
of degree $K+1$ in $\underline m$. Since the probe $z=\lambda+i\eta$
($\eta\to0^+$) is complex, \eqref{eq:appC-mp-poly} is a polynomial with
complex coefficients and has $K+1$ complex roots; at each grid point
$\lambda$ (small finite $\eta$) we find all of them with a standard root
finder and keep the physical one, the unique root with
$\mathrm{Im}\,\underline m>0$ continued from $\underline m\sim-1/z$ at
large $|z|$. The single complex equation determines both parts of
$\underline m=\mathrm{Re}\,\underline m+i\,\mathrm{Im}\,\underline m$ at
once; we impose neither, using $\mathrm{Im}\,\underline m>0$ only to select
the root among the $K+1$. Their meaning follows from
$\underline m(z)=\int\rho(t)/(t-z)\,\dd t$: at $z=\lambda+i0^+$ the
imaginary part is the density we want,
$\mathrm{Im}\,\underline m=\pi\rho(\lambda)$, while the real part is its
Hilbert transform, $\mathrm{Re}\,\underline m=\mathrm P\!\int
\rho(t)/(t-\lambda)\,\dd t$ (a principal value), part of the
self-consistent solution but not itself needed for the density. We check
the solver two ways: against direct fixed-point iteration of
\eqref{eq:appC-mp-discrete}, which it reproduces to machine precision
($\sim10^{-13}$), and against the data.
Figure~\ref{fig:beta-closed-form} overlays $\rho$ (dashed), computed from
\eqref{eq:appC-mp-poly} with $\{\tau_k,g_k\}$ read off the operator, on the
stored empirical eigenvalue density of $M/N$ (solid): the two agree on the
signal outliers, the multiplet positions, and the negative-bulk support.
The sharp small-$|\Lambda|$ peak is not captured by this leading
deformed-MP density; it is a kernel-tail feature (the many near-degenerate
small operator eigenvalues), the noise-dominated localised branch of
Sec.~\ref{sec:discussion} (see Fig.~\ref{fig:localized-branch}).

Whether this deformed-MP density is itself sharp is dimension-dependent
(see Sec.~\ref{sec:bbs-power-laws}). At high latent dimension ($S^{124}$,
$N<d^2$) the feature rows are nearly isotropic and the matrix is in the
genuine Marchenko--Pastur limit, where the deformed-MP law tracks the bulk
end-to-end. At low dimension ($S^2$, the BBS regime $N>d^2$) the matrix is
\emph{not} in the MP limit: the spectrum is dominated by the discrete BBS
multiplets and the near-zero localised peak, and the deformed-MP density
reproduces the signal outliers and the bulk support but not that peak. This
does not affect the dimension inference, which rests on the signal
multiplets, the $N\to\infty$ operator limit of
Sec.~\ref{sec:appendix-mercer-bbs} that is exact at any $d$, rather than on
the bulk being MP-exact. The deformed-MP law is used here as a description
of the continuous part, sharp at high $d$ and a leading approximation at
low $d$.

The inverse problem is a regularised multiplicative deconvolution on a fixed grid.
The same self-consistent equation runs backwards. The forward solve takes the
population $\{\tau_k,g_k\}$ as known and finds $m$ from the polynomial
\eqref{eq:appC-mp-poly} at each probe, a consistency check that a given operator
spectrum reproduces the data. The inverse reverses which side is known:
the observed spectrum determines $m$, and the population becomes the unknown.
In $S$-transform language the multiplicative convolution is a product,
$S_{M/N}=S_{\mathrm{MP}_c}\,S_{H}$, which suggests dividing out the
Marchenko--Pastur factor, $S_H=S_{M/N}/S_{\mathrm{MP}_c}$
\cite{MingoSpeicher2017}. We use this only as a heuristic motivation: the
standard free multiplicative deconvolution is established for \emph{positive}
populations, whereas the operator here can be signed
(see Sec.~\ref{sec:appendix-bulk}), so the $S$-transform division is not justified as
a theorem at the indefinite atoms, and we rely instead on the linear inversion
below, validated numerically. The observed eigenvalues
$\mu_i=\mathrm{eig}(M)/N$ fix the companion transform
$\underline m(z)=\tfrac1N\sum_i(\mu_i-z)^{-1}$ directly, as a finite sum at any
probe $z$ off the real axis, and \eqref{eq:beta-silverstein} rearranges to
\begin{equation}
z+\frac1{\underline m(z)}=\sum_k w_k\,\frac{t_k}{1+t_k\,\underline m(z)},
\label{eq:appC-deconv}
\end{equation}
which is \emph{linear} in the weights $w_k\ge0$ \emph{once the atom locations
$t_k$ are fixed on a prescribed grid} $\{t_k\}$, in contrast to the polynomial
\eqref{eq:appC-mp-poly} solved forward. Stacking several probes and solving by
non-negative least squares then returns the weights $w_k$ on that grid; the
recovered support, the grid points carrying nonzero weight, localises the
operator eigenvalues and their multiplicities $w_k=h_k/N$. This is a regularised
linear inversion on a candidate grid, related numerically to the inversion of
El~Karoui \cite{ElKaroui2008}, not a blind recovery of unknown atom positions:
the resolution is set by the grid spacing and the conditioning of the probe
matrix, not by a closed-form deconvolution.

\begin{figure}[!htbp]
\centering
\includegraphics[width=0.82\textwidth]{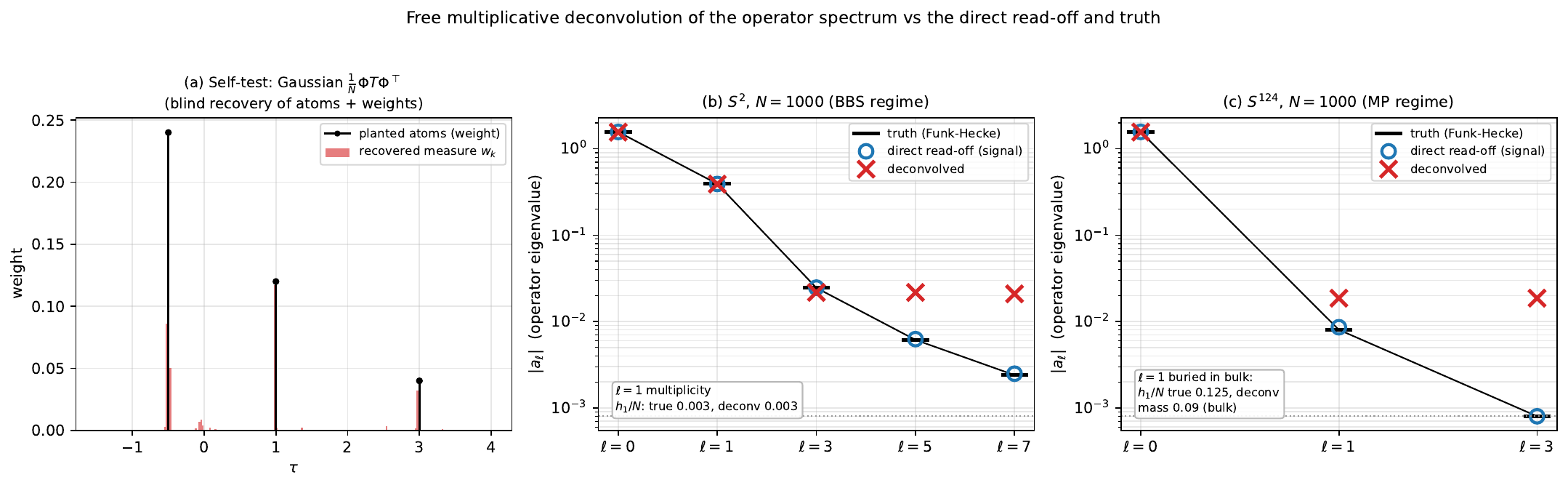}
\caption{Free multiplicative deconvolution of the operator spectrum (the
inverse of the Silverstein map \eqref{eq:beta-silverstein},
Eq.~\eqref{eq:appC-deconv}) against the direct top-eigenvalue read-off and the
truth, $N=1000$. (a) A planted Gaussian sample covariance
$\tfrac1N\Phi T\Phi^\top$ with known signed population is recovered to its atoms
and weights. (b) On $S^2$
(BBS regime) the inverse recovers the Perron value $a_0$ and the $\ell=1$
eigenvalue with multiplicity $h_1/N$ (hence $d=h_1$), matching the direct
read-off and Funk--Hecke; the deep small levels fall on the deconvolution's
resolution floor (dotted), where the direct read-off still resolves them.
(c) On $S^{124}$ (MP regime) the $\ell=1$ level is buried in the bulk and
neither route separates it.}
\label{fig:inverse-deconvolution}
\end{figure}

Figure~\ref{fig:inverse-deconvolution} reports three experiments. Panel~(a) is a
controlled self-test: a planted Gaussian sample covariance
$\tfrac1N\Phi T\Phi^\top$ with a known signed population (atoms
$\tau\in\{-0.5,1,3\}$) is inverted on a grid spanning those values, and the
recovered support and weights $w_k=\#\{\tau\}/N$ reproduce the planted ones. Panels~(b,c) apply the
same inversion to the geodesic matrix on $S^2$ and $S^{124}$, where the atoms are
the operator Funk--Hecke eigenvalues $a_\ell$ and the weights their
multiplicities $h(\ell,d)/N$. On $S^2$ the inverse recovers the Perron value
$a_0$ and the $\ell=1$ eigenvalue with weight $h_1/N$, hence $d=h_1$, agreeing
with the direct top-eigenvalue read-off and the Funk--Hecke truth. The two routes are
complementary rather than independent: both invert the same forward map, the
direct read-off taking the resolved top eigenvalues at face value, the
deconvolution recovering the weights and hence multiplicities on a prescribed
grid. Two limits are
visible. The deep small-$|a_\ell|$ levels sit inside the near-zero localised
pile and fall below the deconvolution's resolution, the generic
ill-conditioning of spectral deconvolution, whereas the sharp BBS staircase
still lets the direct read-off resolve them. And in the MP regime ($S^{124}$)
the $\ell=1$ level is buried in the bulk and neither route separates it,
consistent with the BBS window $[2,\sqrt N]$ not reaching past the first
multiplet there ($\sqrt N<d$). For an ambient matrix $M^{(D)}$ the
operator recovery is preceded by the additive stage of
Sec.~\ref{sec:pt-no-single-level}, which strips the residual cosine-kernel
contribution; the multiplicative deconvolution here acts on the latent matrix
that stage returns.

\subsection{Residual-spectrum RMT consistency}
\label{sec:exp-rmt}

\begin{figure}[!htbp]
\centering
\includegraphics[width=0.82\textwidth]{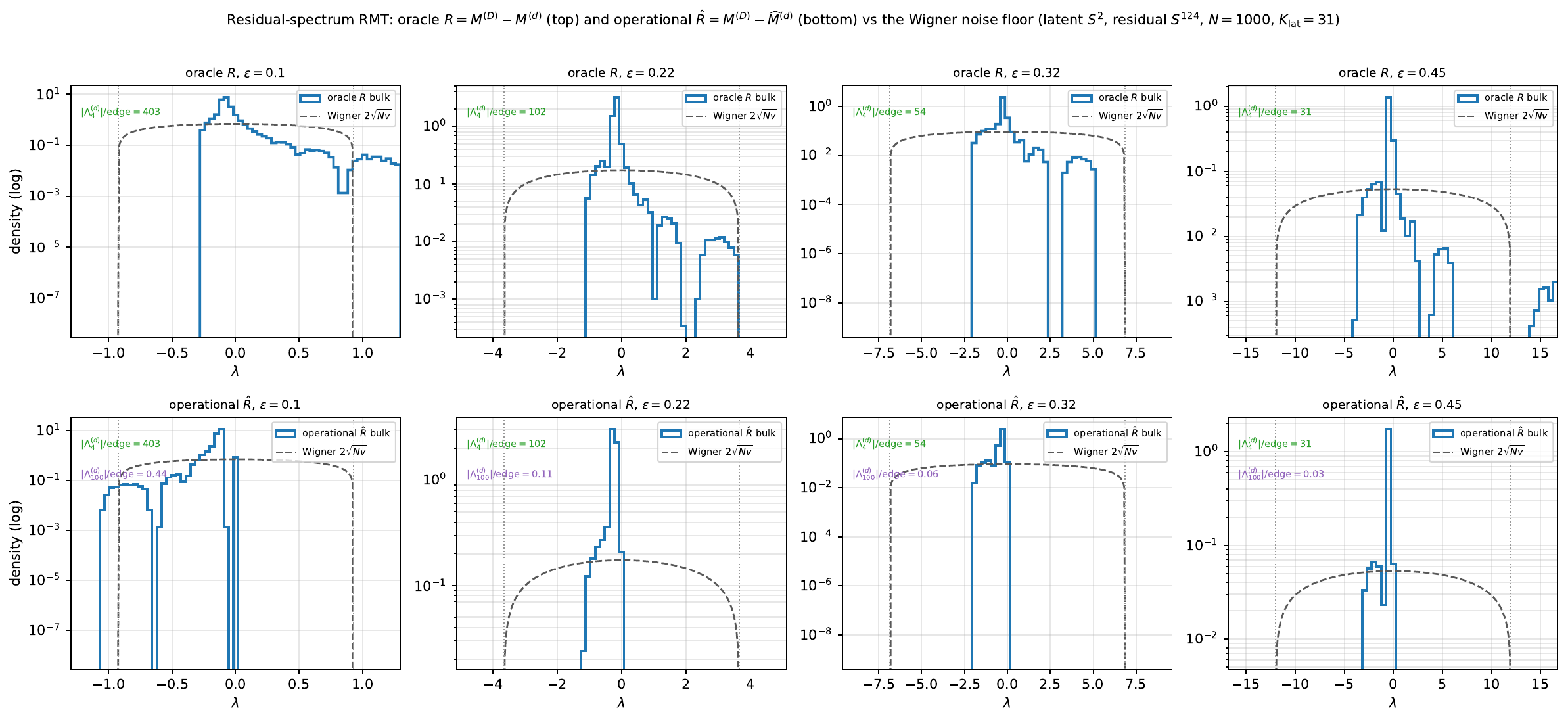}
\caption{Residual-spectrum RMT consistency on $S^2$
($N = 1000$, $D - d = 125$). Top row: oracle residual
$R = M^{(D)} - M^{(d)}$. Bottom row: operational residual
$\hat R = M^{(D)} - \hat M^{(d)}$ from
\eqref{eq:operational-residual} with
$K_{\rm lat} = \lfloor\sqrt N\rfloor$. Columns are
$\varepsilon \in \{0.10, 0.22, 0.32, 0.45\}$. The dashed curve is
the exploratory semicircle reference \eqref{eq:wigner-prediction} at the
residual's own entry variance; the annotations give the ratio of
the latent eigenvalue $|\Lambda^{(d)}|$ to the bulk edge,
which is large at every $\varepsilon$ and quantifies the gap
protection. The operational residual keeps the bulk shape but
retains the localised band of $M^{(d)}$ removed by the oracle
subtraction.}
\label{fig:residual-rmt-merged}
\end{figure}

Once the manifold and the noise model have been identified, a
final consistency check probes whether the off-manifold residual
behaves as structureless random-matrix noise. Subtracting the
recovered latent geometry from the ambient matrix leaves a
residual whose bulk we compare to a semicircle reference set by
the residual's own entry variance. This residual is not a Wigner
matrix: at small noise the oracle residual is
$R_{ij}\approx\varepsilon^{2}\cot M^{(d)}_{ij}
-\varepsilon^{2}\,G_{ij}/\sin M^{(d)}_{ij}$, a deterministic
curvature kernel plus a heteroscedastic Hadamard transform of the
residual Gram matrix $G=\bar Y\bar Y^{\top}$, with entries that
share indices and obey rank and polynomial constraints. Its
limiting bulk is therefore of Marchenko--Pastur type, a
Gram-matrix law, not the semicircle of an i.i.d.\ Wigner ensemble
\cite{AndersonGuionnetZeitouni2010},
and in the proportional regime $D-d\propto N$ it is the deformed-MP
law of Sec.~\ref{sec:appendix-bulk}. We use the semicircle
\eqref{eq:wigner-prediction} only as an exploratory reference for
the bulk shape and edge scale, not as a proven limiting law; a
quantitative residual theory would identify the variance profile
and the limiting spectral distribution under a stated joint limit,
which we do not claim here. The
oracle residual $R = M^{(D)} - M^{(d)}$ uses the
true latent matrix, available only in synthetic tests. The
operational residual replaces it by a rank-$K_{\rm lat}$
truncation of the ambient matrix, with
$K_{\rm lat} = \lfloor\sqrt N\rfloor$,
\begin{equation}
\hat M^{(d)} := \sum_{K=1}^{K_{\rm lat}} \Lambda^{(D)}_K
u^{(D)}_K (u^{(D)}_K)^\top,
\qquad
\hat R := M^{(D)} - \hat M^{(d)},
\label{eq:operational-residual}
\end{equation}
which needs no access to $M^{(d)}$. The semicircle reference for the
residual bulk is
\begin{equation}
\rho_{\rm sc}(\lambda) =
\frac{1}{2\pi N v}\sqrt{4 N v - \lambda^2},
\qquad |\lambda| \leq 2\sqrt{N v},
\label{eq:wigner-prediction}
\end{equation}
with $v$ the residual off-diagonal entry variance, used as a
shape-and-scale guide rather than a derived law.
Figure~\ref{fig:residual-rmt-merged} overlays this reference on
the oracle and operational residual spectra from the
convex-combination forward at
$\varepsilon \in \{0.10, 0.22, 0.32, 0.45\}$ on $S^2$. The oracle
residual shows a semicircle-shaped bulk with the structured outliers of
the deterministic curvature kernel, while the latent
eigenvalues $|\Lambda^{(d)}|$ sit one to three orders of
magnitude beyond the bulk edge: the spectral statement of the
gap protection of Sec.~\ref{sec:exp-multiplets}, that the
dimension-carrying multiplet lives where the residual noise
cannot reach. The operational residual keeps the bulk shape but
retains the small-$|\Lambda|$ localised band that the rank
truncation discards, so it supports a bulk-shape check rather
than a quantitative edge match. Quantitative validation on
representation-space data is in the companion papers
\cite{halperin2026OMD, halperin2026grokking}.

\section{Finite-$N$ corrections to the BBS theory on $S^{d-1}$}
\label{sec:finite-N}

This appendix gives the finite-$N$ corrections to the operator
limit of Appendix~\ref{sec:appendix-beta-saddle}, where the
leading spectrum is $\Lambda_K\to N a_\ell$ with multiplicity
$h(\ell,d)$. The
fitted delocalised slope $\hat\beta_{\rm deloc}$ drifts above the
asymptote $d/(d-1)$ by a deterministic, noise-free amount, the
finite-window sampling of the operator spectrum
(see Sec.~\ref{sec:finite-N-window}); this is the dominant
finite-$N$ effect, and the one the dimension estimator corrects
for. The individual eigenvalue values receive two subleading
corrections from the ambient sampling
(see Sec.~\ref{sec:CLT-correction}): a mean-zero $O(N^{-1/2})$
broadening, and a deterministic $O(1)$ self-energy shift of the
multiplet centre. The broadening cancels in the slope by
self-averaging and the centre shift undershoots it, so neither
controls the slope, but the centre shift is the leading
correction to the value $N a_\ell$ itself.

The localised-branch power law $\beta_{\rm loc} = 1/(d-1)$
\eqref{eq:BBS100} is not treated here: the $v^2 \sim N$
obstruction of Sec.~\ref{sec:bbs-power-laws} makes the BBS
$1/\sqrt N$ machinery inapplicable on that branch, and any
reference to the small-$|\Lambda|$ exponent below is to the
leading-order BBS expression. Throughout this appendix the bare
$\beta$ (and its finite-window fit $\hat\beta$) abbreviates the
delocalised slope $\beta_{\rm deloc}$.

\subsection{Finite-window sampling of the operator spectrum}
\label{sec:finite-N-window}

The BBS power law $|\Lambda_K|\propto K^{-\beta_{\rm deloc}}$,
$\beta_{\rm deloc}=d/(d-1)$, is the $\ell\to\infty$ asymptote of the ranked
operator spectrum derived in Sec.~\ref{sec:appendix-mercer-bbs}
($a_\ell\sim\ell^{-d}$ ranked by $K(\ell)\sim\ell^{d-1}$,
multiplet $\ell$ at $\Lambda\simeq N a_\ell$). The delocalised
diagnostic fits the slope over the window $K\in[2,\sqrt N]$.
Since the cumulative degeneracy is $K(\ell)\sim\ell^{d-1}$, this
window reaches only up to
$\ell_{\max}\sim(\sqrt N)^{1/(d-1)}=N^{1/(2(d-1))}$, a handful of
multiplets at moderate $N$ (e.g.\ $\ell_{\max}\approx5.6$ on
$S^2$ at $N=1000$); the fit therefore samples \emph{low}
$\ell$, where the ranked slope is steeper than $\beta$. As $N$
grows the window edge $\sqrt N$ reaches higher $\ell$ and
$\hat\beta$ drifts down toward $\beta$.

This drift is deterministic, following from the operator
spectrum $\{a_\ell, h(\ell,d)\}$ alone with no reference to the
noise or the random matrix: ranking $\{N|a_\ell|\}$ and fitting
over $[2,\sqrt N]$ gives $\hat\beta(N)-\beta\sim N^{-\nu}$ with a
shallow effective exponent $\nu\approx0.2$--$0.3$ on
$S^1, S^2, S^3$ over the experimental range, well below the
noise-only $\nu=1/2$. These are effective exponents: the ranked
spectrum is a staircase of discrete multiplets whose envelope
reaches the asymptotic slope only as $\ell\to\infty$, so the
drift is a discreteness/pre-asymptotic effect.
It is self-averaging: the per-realisation fitted slope equals
the slope of the realisation-averaged spectrum to numerical
precision, so the $O(N^{-1/2})$ multiplet splitting
(see Sec.~\ref{sec:CLT-correction}) cancels in $\hat\beta$. The
dominant finite-$N$ bias is therefore a property of the
continuum operator sampled on a finite, $N$-dependent window,
not a random-matrix effect.

\subsection{Noise corrections to the multiplet eigenvalues}
\label{sec:CLT-correction}

Beyond the window effect of Sec.~\ref{sec:finite-N-window}, the
ambient sampling corrects the eigenvalue values. In the
factorisation $M=\Phi A\Phi^\top$ of
Sec.~\ref{sec:appendix-bulk} (see Eq.~\eqref{eq:beta-mercer}), the
sampling fluctuation about the population covariance is
$E=\tfrac1N\Phi^\top\Phi-C$, with $\E[E]=0$ and
$E_{ab}=O(N^{-1/2})$ (here $C=I$, uniform). Expanding the
eigenvalues of the degree-$\ell$ block in $E$ gives, in absolute
units, $\Lambda_K = N a_\ell + \sqrt N\,\xi_K + \Delta_\ell + \cdots$: the
leading BBS term $N a_\ell$ ($O(N)$), a mean-zero random broadening
$\sqrt N\,\xi_K$ ($O(\sqrt N)$), and a deterministic self-energy shift
$\Delta_\ell$ ($O(1)$), the two corrections acting at different orders in $E$.

The first correction is a mean-zero broadening of each multiplet.
The first-order term splits the $h(\ell,d)$-fold degenerate
value $N a_\ell$ into $h(\ell,d)$ nearby eigenvalues of width
$O(\sqrt N)$. Equivalently, through the central-limit
decomposition of the row sums
$(Mu)_i=N\mu(\vecx_i)+\sqrt N\,\sigma(\vecx_i)\,\xi_i+O(1)$ and
first-order perturbation theory on the BBS plane-wave ansatz
$u_0=e^{i\vecq\cdot\vecz}$, the rms fluctuation of a delocalised
eigenvalue is
\begin{equation}
\bigl(\Var\Lambda(q)\bigr)^{1/2}=\sqrt N\,\mathcal S(q),
\qquad \mathcal S(q)=\mathcal O(1),
\label{eq:rms_fluct}
\end{equation}
so the relative fluctuation is $O(N^{-1/2})$ and the correction
is mean-zero ($\E\xi_i=0$). The covariance kernel between rows
and the variance integral give the counting-function
correction, a steeper term
$\langle\mathbf N(\lambda)\rangle=C_d(-\lambda)^{-(d-1)/d}
+\tfrac{(d-1)(2d-1)}{2d^2}C_d N^{-1/d}\mathcal S^2
(-\lambda)^{-(3d-1)/d}+O(N^{-2/d})$, of order $N^{-1/d}$ in
scaled units. This broadening
cancels in the slope: the per-realisation fitted
$\hat\beta_{\rm deloc}$ equals the slope of the
realisation-averaged spectrum $\langle|\Lambda_K|\rangle$ to
numerical precision (see Sec.~\ref{sec:finite-N-window}), so it does
not enter the dimension estimate.

A deterministic second-order self-energy shifts the multiplet centre.
The leading correction to the multiplet \emph{value} itself is
the second-order, self-averaging shift of the centre,
\begin{equation}
\Lambda_\ell^{\rm centre}=N a_\ell+\Delta_\ell+O(1/N),
\qquad
\Delta_\ell=\sum_{\ell'\neq\ell}
\frac{h(\ell',d)\,a_\ell\,a_{\ell'}}{a_\ell-a_{\ell'}},
\label{eq:self-energy}
\end{equation}
the pole shift of the resolvent \eqref{eq:appC-resolvent} (the multiplet
pole moves from $a_\ell$ to $a_\ell+\Delta_\ell/N$). It follows from
degenerate second-order perturbation theory in $E$ under the
simplifying assumption that the inter-block matrix elements are
uncorrelated with common variance, $\E E_{ab}^2\approx1/N$:
the off-diagonal coupling between blocks $\ell$ and
$\ell'$ is $a_\ell^{1/2}a_{\ell'}^{1/2}E_{ab}$, and averaging
over the $h(\ell',d)$ states of each
other block gives \eqref{eq:self-energy}. For spherical harmonics
the inter-block second moments are not uniform: the Gaunt and
Wigner coefficients give a nontrivial covariance tensor, so
\eqref{eq:self-energy} is the leading isotropic estimate of the
shift, not its exact value, and we use it as a heuristic rather
than a derived closed form. This shift is $O(1)$
absolute ($O(1/N)$ relative), hence \emph{smaller} than the
$O(\sqrt N)$ broadening: each multiplet broadens by
$\pm O(\sqrt N)$ about a centre displaced from $N a_\ell$ by only
$O(1)$. It is the genuine leading correction to the BBS
eigenvalue value, and summing it over the resolved tower
reproduces the mean multiplet positions of the stored spectra.
The ranked slope it alone implies is comparable to the full
window drift on $S^1, S^2, S^3$, so $\Delta_\ell$ is a leading
contributor to the deterministic slope drift.

The resolvent structure is simple.
These corrections act on the discrete poles of the
resolvent \eqref{eq:appC-resolvent}, the continuous bulk
$\underline m_{\rm bulk}$ unchanged at this order. Averaged over the
configuration the resolvent has an integer-power expansion
$\underline m=\underline m_0+N^{-1}\underline m_1+O(N^{-2})$ with
\emph{no} $1/\sqrt N$ term \cite{Mezard1999}: the $O(N^{-1/2})$
within-multiplet splitting is mean-zero and cancels in the average,
leaving a deterministic object. Only the bulk integrates to a density.

Finally, we estimate the magnitude on $S^2$ ($d=3$).
The combined finite-$N$ departure on $S^2$ at $N=1000$ is of
order $N^{-1/3}\approx0.1$ relative to the asymptote, dominated
by the window drift; this is the right magnitude for the
observed departures from a pure power law (a $T^2$ Fourier
embedding at $N=1000$ gives $\hat\beta\approx1.89$ against the
BBS asymptote $1.5$). The noise estimate \eqref{eq:rms_fluct}
alone would give $N^{-1/2}$ on $d=2$, against the observed
shallower $\approx N^{-0.3}$ (see Sec.~\ref{sec:finite-N-numerics}),
confirming the window effect as the controlling one.

\subsection{Numerical verification}
\label{sec:finite-N-numerics}

\begin{figure}[!htbp]
\centering
\includegraphics[width=0.82\textwidth]{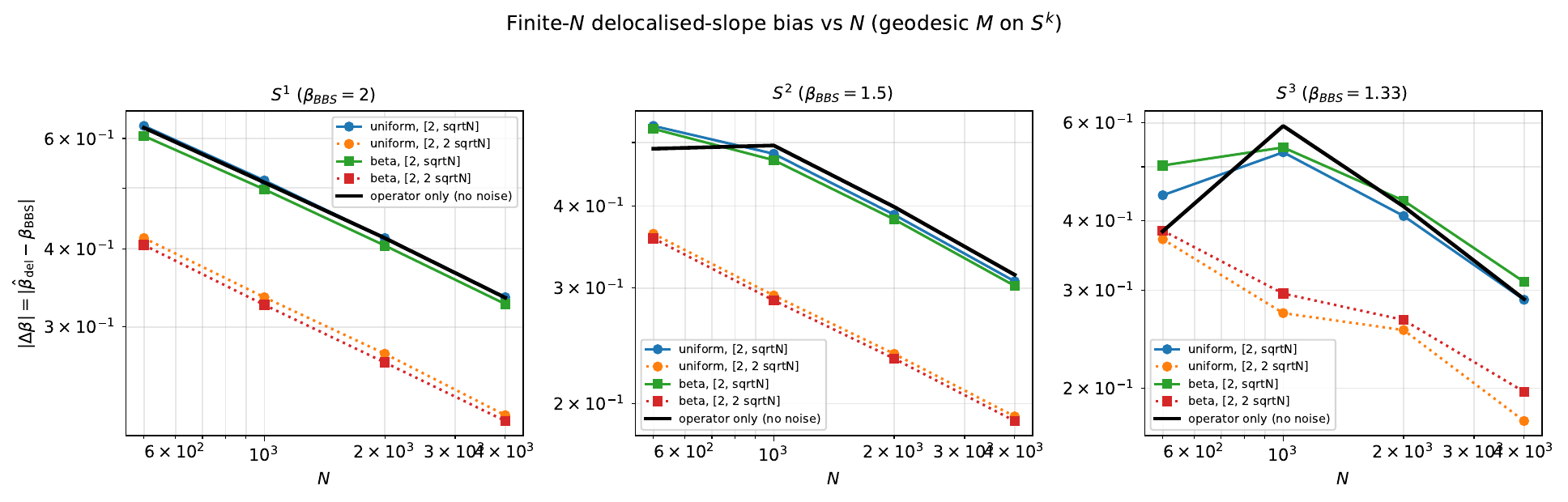}
\caption{Finite-$N$ delocalised-slope bias
$|\hat\beta_{\rm deloc}-\beta_{\rm BBS}|$ versus $N$ for the
geodesic matrix on $S^1, S^2, S^3$ (uniform and product-Beta
sampling, two rank windows). The black line is the deterministic
operator-only bias from the ranked $\{N|a_\ell|\}$ (no noise):
it accounts for most of the drift, the full random-matrix curves
lying close to it. The curves are nearly parallel across the uniform
and product-Beta sampling laws,
and the effective rate ($\nu\approx0.2$--$0.3$) is far
shallower than the $N^{-1/2}$ a noise-only estimate would
give.}
\label{fig:finite-n-scaling}
\end{figure}

We test this on the geodesic matrix of i.i.d.\ particles on
$S^1, S^2, S^3$, reading $\hat\beta_{\rm deloc}$ from the
most-negative eigenvalues over $K\in[2,c\sqrt N]$ and tracking
$\hat\beta_{\rm deloc}-\beta_{\rm BBS}$ against $N\in[500,4000]$
for uniform and product-Beta laws and two windows ($c=1,2$).
Figure~\ref{fig:finite-n-scaling} shows the bias with the
deterministic operator-only curve overlaid: the bare operator
$\{N|a_\ell|\}$, with no noise, already reproduces most of it,
with shallow effective exponents $\nu\approx0.2$--$0.3$ on
$S^1, S^2, S^3$, the full random matrix lying close by, the
ambient noise of Sec.~\ref{sec:CLT-correction} only slightly
steepening the decay. The exponent is the same for the uniform
and product-Beta laws, the curves differing only by a prefactor,
so the drift is set by the operator and the local kernel
geometry, not the single-particle density, and the slope
converges to $\beta_{\rm BBS}$ in every case. The rate is far
shallower than the $N^{-1/2}$ of a noise-only counting argument.
We therefore treat $\Delta\beta(N,d)$ below as an empirically
measured offset.

\subsection{Operational slope correction and its use as a consistency check}
\label{sec:exp-baseline}

The delocalised rank-decay slope enters Algorithm~1 only
as a low-noise consistency check on the dimension already
fixed by the multiplet multiplicity
(see Sec.~\ref{sec:mult-invariant}); the angular-momentum-level
shrinkage law of Sec.~\ref{sec:attenuation-law} is the
primary positional tool.

\begin{figure}[!htbp]
\centering
\includegraphics[width=0.55\textwidth]{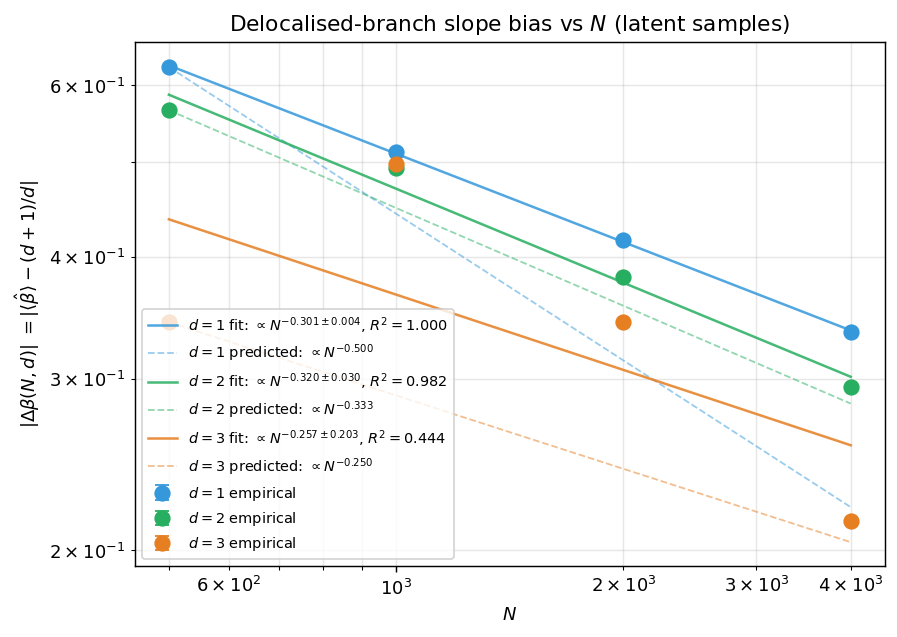}
\caption{Delocalised-branch slope bias $|\Delta\beta(N, d)|$
vs $N$ on latent samples of $S^{d-1}$. Solid lines: power-law
fits with a shallow exponent $\approx0.2$--$0.3$ for $d=2,3,4$,
tracking the deterministic operator-window drift of
Sec.~\ref{sec:finite-N-window}; the rate is shallower than
the $N^{-1/2}$ of a noise-only estimate.}
\label{fig:beta-del-N}
\end{figure}

At finite $N$ the fitted latent slope sits above the BBS
asymptote $d/(d-1)$ by an $O(1)$ offset, dominated by the
finite-window curvature of Sec.~\ref{sec:finite-N-window}: at
$N = 1000$,
$\hat\beta_{\rm deloc}^{(d=2)} = 2.512 \pm 0.003$
(BBS $2$) and $\hat\beta_{\rm deloc}^{(d=3)} =
1.992 \pm 0.004$ (BBS $1.5$), reproducible to four digits
across $20$ uniform-sample configurations. We define the
empirical correction
\begin{equation}
\Delta\beta(N, d)
\;:=\;
\bigl\langle \hat\beta^{(d, N), s}_{\rm lat}\bigr\rangle_{s}
\;-\;
\frac{d}{d-1}
\label{eq:delta-beta}
\end{equation}
from $50$ uniform latent samples (at $N = 1000$,
$\Delta\beta = 0.513, 0.491, 0.497$ for $d = 2, 3, 4$),
the corrected estimator $\hat\beta^{\rm corr}_{\rm deloc}(d_{\rm guess}) :=
\hat\beta_{\rm deloc}(K \in [d_{\rm guess}, \lfloor\sqrt N\rfloor]) -
\Delta\beta(N, d_{\rm guess})$, whose per-candidate fit window
coincides with the delocalised window $[2,\sqrt N]$ at
$d_{\rm guess}=2$ and starts just past the leading multiplet
otherwise, and the inference rule
\begin{equation}
\hat d_{\beta}
\;:=\;
\mathrm{argmin}_{d_{\rm guess}}
\Bigl| \hat\beta^{\rm corr}_{\rm deloc}(d_{\rm guess})
       - \frac{d_{\rm guess}}{d_{\rm guess}-1}
\Bigr|.
\label{eq:d-beta-rule}
\end{equation}
The $N$-dependence of $\Delta\beta$ is the operator-window
drift of Sec.~\ref{sec:finite-N-window};
Fig.~\ref{fig:beta-del-N} shows it measured across
$N \in \{500, 1000, 2000, 4000\}$ and $d \in \{2, 3, 4\}$, with a
shallow fitted exponent $\approx0.2$--$0.3$, consistent with the
operator window-drift of Sec.~\ref{sec:finite-N-numerics}. So
$\Delta\beta$ is a deterministic, $N$-dependent offset, not a
rate-$1/2$ noise correction.

Applied to ambient data, the corrected slope matches the
BBS asymptote at $\varepsilon = 0$ but drifts monotonically
downward under noise: on $S^2$ under FSM,
$\hat\beta^{\rm corr}_{\rm deloc}(2) = 1.50, 1.31, 1.12,
0.93, 0.65$ at $\varepsilon = 0, 0.02, 0.05, 0.10, 0.20$.
The $\mathrm{argmin}$ rule \eqref{eq:d-beta-rule} therefore
recovers the true $d$ reliably only near $\varepsilon = 0$ and
flips for $\varepsilon > 0$. The slope is a low-noise
consistency check, not a stand-alone dimension estimator,
consistent with Sec.~\ref{sec:pt-no-single-level}: the
noise reorganises the spectrum collectively, and the
smaller-$|\Lambda|$ (higher-$\ell$) components shrink
fastest (see Sec.~\ref{sec:attenuation-law}).

\section{Quasi-degenerate perturbation theory cross-check}
\label{sec:appendix-qdpt}

The shrinkage law of Sec.~\ref{sec:attenuation-law} was derived
as the noise-averaged Funk--Hecke coefficient, and the argument
of Sec.~\ref{sec:pt-no-single-level} interprets it as the
resummation of degenerate perturbation theory on the
gap-isolated multiplet. We test this numerically with
the multi-block L\"owdin--Schrieffer--Wolff--van Vleck
transformation, in the algorithmic form of \cite{ArayaDay2025};
the agreement below is a numerical cross-check, not a proof of
resummation, which would identify the block-diagonal continuum
operator and show that its eigenvalue equals the coefficient
ratio.
The construction places $M^{(d)}$ on $N=300$ points of $S^{2}$
as the unperturbed operator and the noise-averaged forward drift
$\langle M^{(D)}\rangle$ as the perturbation, with the residual
drawn on $S^{P-1}$, $P=127$. We use the bounded spliced
form~\eqref{eq:aag-mu-spliced} of that drift.\footnote{Extending the naive
$\arccos$ expansion of the kernel into the region $M^{(d)}_{ij}\to0,\pi$ where
it no longer applies produces an apparent singularity (the unbounded
$\cot,\csc$ prefactors); splicing to the bounded near-boundary form fixes it.}
The $\ell=1$ triplet (the three most negative non-Perron levels of
$M^{(d)}$, separated from the rest by a gap of order $N$) is the
decoupled block, and the transformation returns its effective
$3\times3$ Hamiltonian. The multiplet position at noise level
$\varepsilon$ is the mean of the block eigenvalues, normalised by
its latent value, and gives the shrinkage
$f_1^{\mathrm{PT}}(\varepsilon)$.

\begin{figure}[!htbp]
\centering
\includegraphics[width=0.82\textwidth]{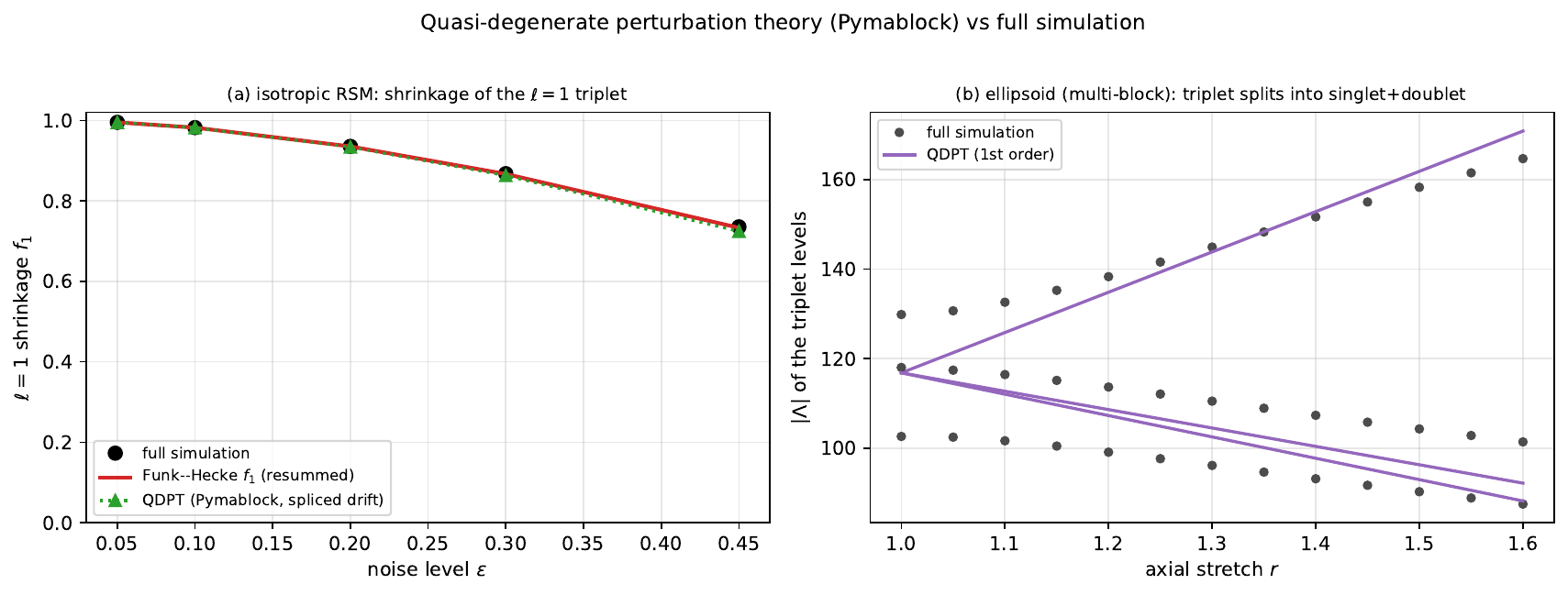}
\caption{Quasi-degenerate perturbation theory
(Pymablock \cite{ArayaDay2025}) against the full simulation.
(a) Isotropic case: the $\ell=1$ shrinkage $f_1$ from the block
reduction of the spliced drift, the full simulation, and the
closed Funk--Hecke coefficient, which coincide across the noise
range. (b) Anisotropic latent: under an axial stretch $r$
the triplet splits into a singlet and a doublet, and the
first-order block prediction (lines) follows the full
diagonalisation (markers).}
\label{fig:qdpt}
\end{figure}

For isotropic noise, Table~\ref{tab:qdpt} compares the
block reduction against the full simulation and the closed
shrinkage $f_1$ of Eq.~\eqref{eq:attenuation-law}. The three
agree across the noise range, and the reduction converges
already at first Schrieffer--Wolff order because the $O(N)$ gap
suppresses the coupling to the bulk; the same reduction
reproduces the shrinkage on $S^1, S^2, S^3$ in
Fig.~\ref{fig:q2}. The closed Funk--Hecke coefficient therefore
agrees with the block reduction across the range, consistent with
its being the resummation of that reduction rather than an
independent fit, the role assigned to it in
Sec.~\ref{sec:attenuation-law}; the agreement is a numerical
cross-check, not a proof of resummation. Fig.~\ref{fig:qdpt}(a) shows the
three curves together.

\begin{table}[t]
\centering
\begin{tabular}{cccc}
\hline
$\varepsilon$ & block reduction (spliced) &
Funk--Hecke $f_1$ & full simulation \\
\hline
$0.05$ & $0.9952$ & $0.9953$ & $0.9953$ \\
$0.10$ & $0.9819$ & $0.9821$ & $0.9821$ \\
$0.20$ & $0.9339$ & $0.9353$ & $0.9356$ \\
$0.30$ & $0.8631$ & $0.8669$ & $0.8677$ \\
$0.45$ & $0.7244$ & $0.7332$ & $0.7355$ \\
\hline
\end{tabular}
\caption{Shrinkage of the $\ell=1$ triplet on $S^{2}$
($N=300$, residual on $S^{124}$). The quasi-degenerate block
reduction (Pymablock \cite{ArayaDay2025}) of the bounded spliced
drift~\eqref{eq:aag-mu-spliced} reproduces the full-simulation
shrinkage across the noise range; the closed Funk--Hecke
coefficient $f_1$ of Eq.~\eqref{eq:attenuation-law} is the
resummation of the same reduction.}
\label{tab:qdpt}
\end{table}

We also test an anisotropic latent. A single $3$-dimensional block
suffices only while the latent is isotropic and the triplet
stays degenerate. Stretching the latent $S^{2}$ along its axis by
a factor $r$ breaks the rotational symmetry to $SO(2)\times
\mathbb{Z}_2$ and splits the triplet into a singlet (the axial
mode) and a doublet. The block treatment captures this through
the secular problem inside the multiplet: the first-order
effective Hamiltonian has one isolated eigenvalue and a
near-degenerate pair, and the resulting level positions
$\Lambda_1+(r-1)\,\delta\Lambda$ follow the full diagonalisation
through the onset of the split, as shown in
Fig.~\ref{fig:qdpt}(b). Resolving co-existing sub-levels of this
kind is the capability that distinguishes the multi-block
algorithm from a two-subspace Schrieffer--Wolff reduction.

\end{document}